\documentclass{article} %

\usepackage{hyperref}
\usepackage{url}

\usepackage{iclr2023_conference,times}

\usepackage[utf8]{inputenc} %
\usepackage[T1]{fontenc}    %
\usepackage{hyperref}       %
\usepackage{url}            %
\usepackage{booktabs}       %
\usepackage{amsfonts}       %
\usepackage{nicefrac}       %
\usepackage{microtype}      %
\usepackage{xcolor}         %
\usepackage{floatrow}
\usepackage{float}
\usepackage{amsmath}
\usepackage{graphicx}
\usepackage{cleveref}
\Crefname{algocf}{Algorithm}{Algorithms}

\usepackage[skip=2pt]{caption}
\usepackage{wrapfig}
\usepackage[ruled,vlined,linesnumbered, commentsnumbered]{algorithm2e}
\SetKwInput{KwData}{Input}
\SetKwInput{KwResult}{Output}

\usepackage{sidecap}

\usepackage{etoolbox}

\usepackage{pdfpages}

\newcommand{\Fc}{\mathcal{F}}

\newcommand{\Ic}{\mathcal{I}}

\newcommand{\Mc}{\mathcal{M}}
\newcommand{\Nc}{\mathcal{N}}

\newcommand{\Qc}{\mathcal{Q}}

\newcommand{\Tc}{\mathcal{T}}

\newcommand{\Fcb}{{\boldsymbol \Fc}}

\newcommand{\0}{{\boldsymbol 0}}

\newcommand{\Bb}{{\boldsymbol B}}

\newcommand{\Db}{{\boldsymbol D}}

\newcommand{\Fb}{{\boldsymbol F}}

\newcommand{\Ib}{{\boldsymbol I}}

\newcommand{\Jb}{{\boldsymbol J}}

\newcommand{\Pb}{{\boldsymbol P}}

\newcommand{\Rb}{{\boldsymbol R}}
\newcommand{\Sb}{{\boldsymbol S}}

\newcommand{\Ub}{{\boldsymbol U}}

\newcommand{\Xb}{{\boldsymbol X}}

\newcommand{\ab}{{\boldsymbol a}}
\newcommand{\bb}{{\boldsymbol b}}

\newcommand{\eb}{{\boldsymbol e}}
\newcommand{\fb}{{\boldsymbol f}}
\newcommand{\gb}{{\boldsymbol g}}

\newcommand{\hb}{{\boldsymbol h}}

\newcommand{\qb}{{\boldsymbol q}}

\newcommand{\ub}{{\boldsymbol u}}

\newcommand{\xb}{{\boldsymbol x}}
\newcommand{\yb}{{\boldsymbol y}}

\newcommand{\pd}[2]{ \dfrac{\partial #1}{\partial #2}}
\newcommand{\pdflat}[2]{\tfrac{\partial #1}{\partial #2}}
\newcommand{\pdtwo}[2]{ \dfrac{\partial^2 #1}{\partial #2^2}}
\newcommand{\pdflattwo}[2]{ \tfrac{\partial^2 #1}{\partial #2^2}}
\newcommand{\grad}{{\nabla}}
\newcommand{\gradb}{{\boldsymbol \nabla}}

\DeclareMathOperator{\trace}{\sf tr}

\definecolor{Blue}{RGB}{166,206,227}
\definecolor{Green}{RGB}{178,223,138}
\definecolor{Pink}{RGB}{251,154,153}
\definecolor{Orange}{RGB}{253,191,111}
\definecolor{Red}{RGB}{255,0,0}

\newcommand{\rev}[1]{{#1}}

\newcommand{\RR}[1]{\mathbb{R}^{#1}}

\newcommand{\nSpatial}{{P}}

\newcommand{\nTemporal}{{T}}

\newcommand{\params}{\boldsymbol \mu}
\newcommand{\paramDomain}{\mathcal D}
\newcommand{\paramDomainTrain}{\mathcal D_\mathrm{train}}
\newcommand{\paramDomainTest}{\mathcal D_\mathrm{test}}

\newcommand{\lowDimensionalManifold}{{\gb}}
\newcommand{\lowDimensionalManifoldArgs}[2]{\lowDimensionalManifold(#1,#2)}

\newcommand{\weights}{\theta_g}
\newcommand{\lowDimensionalManifoldNN}{{\gb_{\weights}}}
\newcommand{\lowDimensionalManifoldNNArgs}[2]{\lowDimensionalManifoldNN(#1,#2)}

\newcommand{\lowDimensionalManifoldPrior}{{\gb_\nSpatial}}

\newcommand{\latentSpaceVec}{\qb}
\newcommand{\latentSpaceVecArg}[1]{\latentSpaceVec(#1)}
\newcommand{\latentSpaceVecArgParams}[1]{\latentSpaceVec(#1;\params)}
\newcommand{\latentSpaceVecDot}{\dot{\latentSpaceVec}}
\newcommand{\discreteLatentSpaceVec}{\latentSpaceVec_{n}}
\newcommand{\discreteLatentSpaceVecDot}{\dot{\latentSpaceVec}_{n}}
\newcommand{\discreteLatentSpaceVecPlus}{\latentSpaceVec_{n+1}}
\newcommand{\discreteLatentSpaceVecMinus}{\latentSpaceVec_{n-1}}
\newcommand{\discreteLatentSpaceVecDotZero}{\dot{\latentSpaceVec}_{0}}

\newcommand{\fullOrderModel}{{\fb}}
\newcommand{\fullOrderModelArgs}[2]{\fullOrderModel(#1,#2)}
\newcommand{\fullOrderModelArgsParam}[2]{\fullOrderModel(#1,#2;\params)}

\newcommand{\fullOrderModelDiscretized}{{\fullOrderModel_\nSpatial}}
\newcommand{\disretizationCoeff}{{\ab}}
\newcommand{\disretizationCoeffDiscrete}{{\disretizationCoeff^i}}
\newcommand{\disretizationCoeffDiscreteArg}[1]{\disretizationCoeffDiscrete(#1)}

\newcommand{\disretizationShape}{{N}}
\newcommand{\disretizationShapeDiscrete}{{\disretizationShape^i}}
\newcommand{\disretizationShapeDiscreteArg}[1]{\disretizationShapeDiscrete(#1)}

\newcommand{\discretePos}{\xb^{i}}
\newcommand{\discretePosSet}{\{\discretePos\}_{i=1}^{\nSpatial}}
\newcommand{\discretePosFirst}{\xb^{1}}
\newcommand{\discretePosLast}{\xb^{\nSpatial}}
\newcommand{\continuousTime}{t}
\newcommand{\discreteTime}{t_{n}}
\newcommand{\discreteTimePlus}{t_{n+1}}

\newcommand{\discreteEnd}{t_{\nTemporal}}
\newcommand{\pos}{\xb}
\newcommand{\posSample}{{\yb}}
\newcommand{\discretePosSample}{{\yb^{j}}}

\newcommand{\spatialDomain}{\Omega}
\newcommand{\temporalDomain}{\Tc}

\newcommand{\spatialIndexSet}{{i=1, \ldots, \nSpatial}}

\newcommand{\weightsEncoder}{\theta_e}
\newcommand{\encoder}{{\eb_{\weightsEncoder}}}
\newcommand{\encInput}{{\overrightarrow{\fullOrderModel}}}
\newcommand{\encInputArg}[1]{\encInput(#1)}
\newcommand{\encInputDetailArg}[1]{\begin{pmatrix} \fullOrderModelArgs{\discretePosFirst}{#1}, \hdots \fullOrderModelArgs{\discretePos}{#1}, \hdots \fullOrderModelArgs{\discretePosLast}{#1} \end{pmatrix}^{T}}
\newcommand{\encInputDetailArgShort}[1]{\begin{pmatrix} \fullOrderModelArgs{\discretePosFirst}{#1}, \hdots \fullOrderModelArgs{\discretePosLast}{#1} \end{pmatrix}^{T}}

\newcommand{\sampleSet}{\Mc}
\newcommand{\sampleSetCardinality}{|\sampleSet|}

\newcommand{\sampleSetNeighbor}{\Nc}
\newcommand{\sampleSetNeighborCardinality}{|\sampleSetNeighbor|}

\newcommand{\gradbspatial}{\gradb_{\xb}}
\newcommand{\gradbspatialSample}{\gradb_{\posSample}}

\newcommand{\dt}{\Delta t}

\newcommand{\TimeIntegrate}{\Ic_{\pdelhs}}

\newcommand{\nred}{r}

\newcommand{\temperatureVec}{\ub}
\newcommand{\fluidVelo}{\ub}
\newcommand{\pressure}{p}

\newcommand{\externalForce}{\fb_{ext}}
\newcommand{\deformationMap}{{\boldsymbol\phi}}
\newcommand{\PKStress}{\Pb}
\newcommand{\bodyForce}{\Bb}
\newcommand{\deformationGrad}{\Fb}

\newcommand{\deltaFullOrder}{\Delta\fb}
\newcommand{\deltaFullOrderArgs}[2]{\deltaFullOrder(#1,#2)}

\newcommand{\lowDimensionalManifoldNNwrtLatent}{\pd{\lowDimensionalManifoldNN}{\latentSpaceVec}}
\newcommand{\lowDimensionalManifoldNNwrtLatentArgs}[2]{\lowDimensionalManifoldNNwrtLatent(#1,#2)}
\newcommand{\lowDimensionalManifoldNNwrtLatentFlat}{\pdflat{\lowDimensionalManifoldNN}{\latentSpaceVec}}
\newcommand{\lowDimensionalManifoldNNwrtLatentFlatArgs}[2]{\lowDimensionalManifoldNNwrtLatentFlat(#1,#2)}

\newcommand{\deltaDiscreteLatentSpaceVecPlus}{\Delta\discreteLatentSpaceVecPlus}

\newcommand{\dimensionIn}{m}
\newcommand{\dimensionOut}{d}
\newcommand{\scaleMlp}{\beta}

\newcommand{\identity}{\Ib}

\newcommand{\gridSpacing}{{\Delta x}}
\newcommand{\gridSpacingY}{{\Delta y}}

\newcommand{\podbasis}{\Ub}
\newcommand{\decoder}{\hb}
\newcommand{\decoderArg}[1]{\hb(#1)}

\newcommand{\remeshHiVertexNumber}{{2,065}}
\newcommand{\remeshHiTetNumber}{{9,346}}

\newcommand{\remeshLoVertexNumber}{{8}}
\newcommand{\remeshLoTetNumber}{{5}}

\newcommand{\piggyVertexNumber}{{66,608}}

\newcommand{\piggyIntegrationSamples}{{40}}

\newcommand{\imagePixelNumber}{{65,536}}

\newcommand{\cromPiggyError}{{1.46\%}}
\newcommand{\podPiggyError}{{5.76\%}}

\newcommand{\defeq}{:=}

\newcommand{\pdelhs}{\Fcb}

\newcommand{\firstlame}{\lambda}
\newcommand{\secondlame}{\mu}

\newcommand{\zerob}{{\boldsymbol 0}}

\newcommand{\viscMax}{0.02}
\newcommand{\viscMaxExtrapolate}{0.1}

\newcommand{\numtry}{\Qc}

\newcommand{\advectQuantity}{u}
\newcommand{\advectSpeed}{a}

\newcommand{\burgerQuantity}{w}

\newcommand{\burgerD}{\mu_{D}}
\newcommand{\burgerDtrain}{\mu_{D_{train}}}
\newcommand{\burgerDtest}{\mu_{D_{test}}}
\newcommand{\inDimension}{p^{*}}

\title{CROM: Continuous Reduced-Order Modeling of PDEs Using Implicit Neural Representations}
\author{%
  Peter Yichen Chen$^{\textbf{1,3}}$\quad Jinxu Xiang$^{\textbf{1}}$\quad
  Dong Heon Cho$^{\textbf{1}}$\quad
  Yue Chang$^{\textbf{4}}$\quad
  G A Pershing$^{\textbf{1}}$
  \And
  Henrique Teles Maia$^{\textbf{1}}$\quad Maurizio M. Chiaramonte$^{\textbf{2}}$\quad Kevin Carlberg$^{\textbf{2}}$\quad
  Eitan Grinspun$^{\textbf{4,1}}$ \\
  \And
  {\normalfont $^{1}$Columbia University\quad}
  {\normalfont $^{2}$Meta Reality Labs Research\quad}
  {\normalfont $^{3}$MIT CSAIL\quad}
  {\normalfont $^{4}$University of Toronto}
}

\iclrfinalcopy %
\begin{document}

\maketitle

\begin{abstract}
    The long runtime of high-fidelity partial differential equation (PDE) solvers makes them unsuitable for time-critical applications. We propose to accelerate PDE solvers using reduced-order modeling (ROM). Whereas prior ROM approaches reduce the dimensionality of \emph{discretized} vector fields, our \emph{continuous} reduced-order modeling (CROM) approach builds a \rev{low-dimensional embedding} of the \emph{continuous} vector fields themselves, not their discretization. We represent this reduced manifold using continuously differentiable neural fields, which may train on any and all available numerical solutions of the continuous system, even when they are obtained using diverse methods or discretizations. We validate our approach on an extensive range of PDEs with training data from voxel grids, meshes, and point clouds. Compared to prior discretization-dependent ROM methods, such as linear subspace proper orthogonal decomposition (POD) and nonlinear manifold neural-network-based autoencoders, CROM features higher accuracy, lower memory consumption, dynamically adaptive resolutions, and applicability to any discretization. For equal latent space dimension, CROM exhibits 79$\times$ and 49$\times$ better accuracy, and 39$\times$ and 132$\times$ smaller memory footprint, than POD and autoencoder methods, respectively. Experiments demonstrate 109$\times$ and 89$\times$ wall-clock speedups over unreduced models on CPUs and GPUs, respectively. Videos and codes are available on the project page: \url{https://crom-pde.github.io}.
\end{abstract}

\vspace{-6mm} %
\section{Introduction}
Many scientific and engineering models are posed as partial differential equations (PDEs) of the form
\begin{align}
    \pdelhs(\fullOrderModel, \gradb\fullOrderModel, \gradb^2\fullOrderModel, \ldots, \dot{\fullOrderModel}, \ddot{\fullOrderModel}, \ldots) = \0, \quad \fullOrderModelArgs{\pos}{t}: \spatialDomain \times \temporalDomain \to \RR{\dimensionOut} \ ,    \label{eqn:full-order-continuous}
\end{align}
subject to initial and boundary conditions. Here $\fullOrderModel$ is a spatiotemporal dependent, multidimensional continuous vector field, such as temperature, velocity, or displacement; $\gradb$ and $\dot{(\cdot)}$ are the spatial and temporal gradients; $\spatialDomain\subset\RR{\dimensionIn}$ and $\temporalDomain\subset\RR{}$ are the spatial and temporal domains, respectively.

We may solve for $\fullOrderModel$ by discretizing in space, $\fullOrderModel(\pos,t) \approx \fullOrderModelDiscretized(\pos,t)= \sum_{i=1}^{\nSpatial}\disretizationCoeffDiscreteArg{t}\disretizationShapeDiscreteArg{\pos}$, transforming the continuous spatial representation to a ($\nSpatial \cdot {\dimensionOut}$)-dimensional vector whose coefficients $\disretizationCoeffDiscreteArg{t}: \temporalDomain \to \RR{\dimensionOut}$ and the corresponding basis functions $\disretizationShapeDiscreteArg{\pos}: \spatialDomain \to \RR{}$ (e.g., polynomial basis, fourier basis) approximate the continuous solution. For instance, if $\disretizationShapeDiscrete$ is the linear finite element basis, the coefficients $\disretizationCoeffDiscreteArg{t}=\fullOrderModelArgs{\discretePos}{t}$ are field values at spatial samples $\discretePos$ \citep{hughes2012finite}. 

\begin{figure}
  \vspace{-12mm} %
    \centering
    \includegraphics[width=\textwidth]{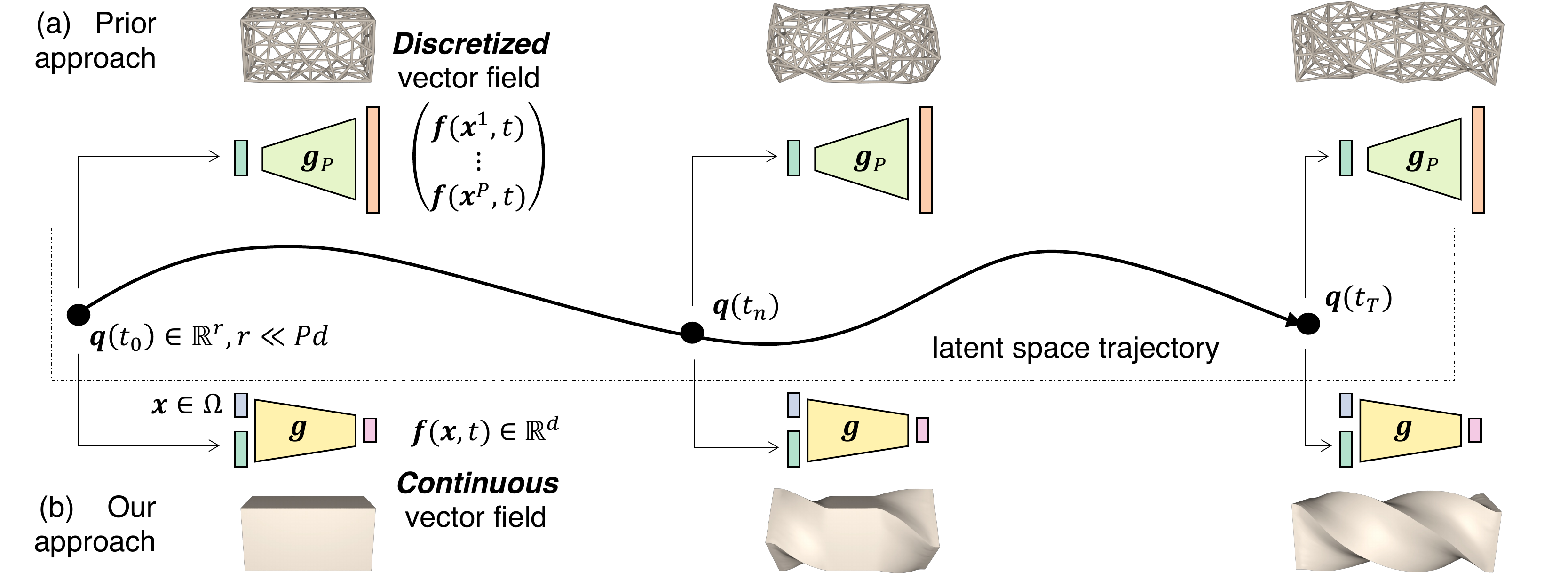}
    \vspace{-6mm} %
    \caption{Model reduction solves PDEs via temporal evolution of the low-dimensional latent space vector $\latentSpaceVecArg{t}$. (a) Prior work assumes that the low-dimensional representation $\lowDimensionalManifoldPrior$ is built for the \emph{already-discretized} vector field; (b) our approach constructs \rev{the manifold-parameterization function} $\lowDimensionalManifold$ directly for the \emph{continuous} vector field itself. In this case, the vector field $\fullOrderModel$ represents the twisting material governed by the elastodynamics equation.\vspace{-0mm}} %
    \label{img:vs_classic}
  \end{figure}

After introducing temporal samples $\{\discreteTime\}_{n=0}^{\nTemporal}$, we temporally evolve the solution by solving for $\nSpatial$ unknowns  $\{\disretizationCoeffDiscreteArg{\discreteTimePlus}\}$ given the previous state $\{\disretizationCoeffDiscreteArg{\discreteTime}\}$. Unfortunately, when $\nSpatial$ is large, processing and memory costs of these \emph{full-order} solves become intractable. To alleviate this computational burden, prior model reduction techniques \citep{berkooz1993proper,willcox2002balanced,benner2015survey} construct a manifold-parameterization function $\lowDimensionalManifoldPrior: \mathbb{R}^{\nred} \mapsto \mathbb{R}^{\nSpatial\dimensionOut}$, with $\nred \ll \nSpatial\dimensionOut$, such that every low-dimensional latent space vector $\latentSpaceVecArg{t}\in\RR{\nred}$ maps to a discrete field $
\lowDimensionalManifoldPrior(\latentSpaceVec) \mapsto
(\disretizationCoeff^1,\ldots,\disretizationCoeff^{\nSpatial})^{T}$. For instance, for linear finite elements \citep{barbivc2005real}, $
\lowDimensionalManifoldPrior(\latentSpaceVec) \mapsto
\encInputDetailArgShort{t}$, as depicted in \Cref{img:vs_classic}a. ROM saves computation because it requires evolving only $\nred \ll \nSpatial\dimensionOut$ latent space variables.\footnote{The latent space vector is also known as the feature, subspace, or state vector; or the generalized coordinates.}

Since existing ROM approaches apply to \emph{already-discretized} fields, model training and PDE solving are tied to the dimension and discretization type of the training data, causing key limitations: 

\noindent\textbf{Discretization dependence.}
  If we alter the training simulation resolution ($\nSpatial$) or the discretization types (e.g., meshes to point clouds), we must also alter the architecture and numbers of parameters.
  
\noindent\textbf{Memory scaling.} 
  Memory footprint grows with discretization resolution $\nSpatial$.
  
\noindent\textbf{Fixed discretization.}
We cannot dynamically \emph{adapt} spatial resolution $\nSpatial$,  discretization type, or  basis function $\disretizationShapeDiscrete$ during latent-space-PDE solves, e.g., dynamic remeshing \citep{PERAIRE1987449}.

Altogether these problems arise because the architecture of $\lowDimensionalManifoldPrior(\latentSpaceVec)$ is tied to the discretization $(\disretizationCoeff^1,\ldots,\disretizationCoeff^{\nSpatial})^{T}$.

\textbf{Introducing a discretization-independent architecture}  In an alternative point of departure, we train a manifold-parameterization function 
$\lowDimensionalManifoldArgs{\pos}{\latentSpaceVec} \approx \fullOrderModel(\pos, t)$ to approximate the \emph{continuous} field itself, \emph{not} its discretization (see \Cref{img:vs_classic}b). Note that the domain and co-domain of $\lowDimensionalManifold$ are continuous domains: they do not depend on the choice of discretization(s) used at any stage of the process, i.e., during preparation of training data, nor during latent-space-PDE solving. In this sense, the manifold-parameterization architecture is \emph{discretization independent.}  In our implementation, $\lowDimensionalManifold$ is embodied as an implicit neural representation \citep{park2019deepsdf, chen2019learning,mescheder2019occupancy}, also known as a neural field, yielding a smooth and analytically-differentiable manifold-parameterization. This representation's memory footprint depends on the complexity of fields produced by the PDE, \emph{not} the discretization \emph{resolution}.

After training, we evolve the latent variables, as governed by the PDE, for previously-unexplored parameters.
Unlike approaches that discard the PDE after training, we evaluate the original PDE at a small number of domain points at every time integration step. We validate our approach on classic PDEs with discretized data from voxels, meshes, and point clouds. In comparison to the full-order model, our approach reduces the number of spatial degrees of freedom, memory, and computational cost. In comparison to prior linear and nonlinear discretization-dependent model reduction methods, our method exhibits higher accuracy and consumes less memory. To highlight another benefit of being discretization-agnostic, we demonstrate an elasticity simulation that readily adapts mesh resolution.

\section{Related Work}

\textbf{Reduced-Order Modeling for PDEs.}  Early works on identifying a low-dimensional latent space focused on linear methods \citep{berkooz1993proper,holmes2012turbulence}, e.g., proper orthogonal decomposition (POD) or principal component analysis (PCA). Recent nonlinear manifolds \citep{fulton2019latent,lee2020model}, often constructed via autoencoder neural networks, have been shown to significantly outperform their linear counterparts on slowly decaying Kolmogorov n-width problems \citep{peherstorfer2022breaking}. Most of these prior works exclusively focus on building a latent space for the \emph{already-dicretized} vector fields. \citet{chen2023model,pan2022neural} and our work attempt to construct the latent space for the continuous vectors themselves. However, \citet{chen2023model}'s treatment specializes in the material point method (discretization) and elasticity (PDE); our general treatment is both discretization and PDE agnostic. Likewise, \citet{pan2022neural} \emph{train} a discretization-agnostic latent space for PDE data; we further \emph{solve} the PDEs in the reduced space via rapid latent space traversal. Specifically, we evolve the latent space in a PDE-constraint manner while \citet{yin2023continuous} take a data-driven approach. Additional references to ROM are listed in \Cref{sec:lite_rev}.

\textbf{Implicit neural representations} use (fully-connected) neural networks to represent arbitrary vector fields. While the Euclidean space spatial coordinates always form part of the input to the network, it is also common to have a latent space vector to complement the rest of the input. Different latent space vectors correspond to different states of the continuous vector field (see \Cref{img:vs_classic}b), e.g. different geometries \citep{park2019deepsdf, chen2019learning,mescheder2019occupancy} or different radiance fields \citep{mildenhall2020nerf}. A key contribution of our work is nonlinearly traversing the latent space of neural representations under an explicit PDE constraint.

\textbf{Machine learning (ML) for PDEs.} Physics-informed neural networks (PINNs) \citep{raissi2019physics,sitzmann2020implicit}
demonstrate that PDEs can be accurately solved via neural representations. Notably, PINN enables prediction and discovery from incomplete models and incomplete data \citep{karniadakis2021physics}. However, the degrees of freedom involved in their approaches are still $\nSpatial\dimensionOut$, and the underlying gradient-descent-based solver is often computationally more \emph{expensive} than traditional solvers (see Table 1 by \citet{zehnder2021ntopo}). By contrast, our goal is building a computationally more \emph{efficient} solution that solves for only $\nred$ degrees of freedom ($\nred\ll\nSpatial\dimensionOut$). In fact, our approach can be viewed as an extension of PINN for model reduction. Setting the latent space vector $\latentSpaceVec(\continuousTime)$ in our formulation (see \Cref{img:vs_classic}b) to the time variable $\continuousTime$ recovers the exact formulation of PINN. In addition to PINN, \citet{sanchez2020learning} show graph neural network (GNN) architectures are also capable of learning PDEs. Yet, like PINNs, GNNs also do not offer dimension reduction. 
\section{Method: Overview and Manifold Construction}
\textbf{Overview} Our goal is to efficiently obtain the solution of \Cref{eqn:full-order-continuous}. We begin by constructing a low-dimensional embedding and the corresponding manifold-parameterization function (see below), after which we solve PDEs by time-integrating the dynamics of the manifold's latent space vector (see \Cref{sec:latent-space-dynamics}). As we demonstrate in examples from various scientific disciplines (see \Cref{sec:results}), this general method is applicable regardless of the discretization of the training data (e.g., voxel grids, meshes, point clouds) or the discretization deemed most useful for evaluating the gradients (e.g., physical forces) when solving PDEs on the constructed manifold.

\textbf{Manifold Construction}
\label{sec:manifold-construction}
As depicted in \Cref{img:vs_classic}b, we seek a manifold-parameterization function $\lowDimensionalManifoldArgs{\pos}{\latentSpaceVec}$,
\begin{align}
    \lowDimensionalManifoldArgs{\pos}{\latentSpaceVecArgParams{t}} \approx  \fullOrderModelArgsParam{\pos}{t}\ ,\quad \forall\pos\in\spatialDomain\ ,\quad \forall t\in\temporalDomain\ ,\quad \forall\params\in \paramDomain \ ,
\end{align}
that well approximates the continuous field $\fullOrderModelArgsParam{\pos}{t}$ throughout the spatiotemporal domain $\spatialDomain \times \temporalDomain$, and for a workable range of problem parameters $\params\in \paramDomain$. For ease of exposition, we omit the dependencies of $\latentSpaceVec$ and $\fullOrderModel$ on the problem parameters $\params$. Here $\paramDomain$ is an arbitrary parameter space (e.g.,  material properties, external forces, user settings). For scenarios that do not feature trivial parameterizations, e.g., external force via crowd-sourcing \citep{barbivc2005real}, $\paramDomain$ can also be implicitly defined. 

What distinguishes our approach from typical model reduction is that $\lowDimensionalManifold$ takes the position $\pos\in\spatialDomain$ as an input. Thus, unlike prior approaches that infer only discrete coefficients, our approach infers field values at arbitrary domain positions $\pos$. 

We parameterize $\lowDimensionalManifold$ with a neural network $\lowDimensionalManifoldNN$ whose weights $\weights$ satisfy the minimization problem
\begin{align}
    \min_{\weights} \sum_{i=1}^P \sum_{n=0}^T  \sum_{\params\in \paramDomainTrain} \|\lowDimensionalManifoldNNArgs{\discretePos}{\latentSpaceVecArg{\discreteTime}} - \fullOrderModelArgs{\discretePos}{\discreteTime}\|_2^2 \ ,
    \label{eqn:loss_decoder_only}
\end{align}
where $\paramDomainTrain\subset\paramDomain$ is the training set, and $\latentSpaceVecArg{\discreteTime}$ is the latent space vector shared among all spatial samples. This objective aims to reproduce all the field values present in the training data, generated via full-order PDE solutions. Notably, our approach imposes no limit on the discretization strategy of the PDE solver. For instance, this framework is applicable to training data from both finite difference methods and finite element methods as well as both voxel grids and meshes.

\begin{figure}
    \vspace{-14mm} %
    \centering
    \includegraphics[width=0.7126\textwidth]{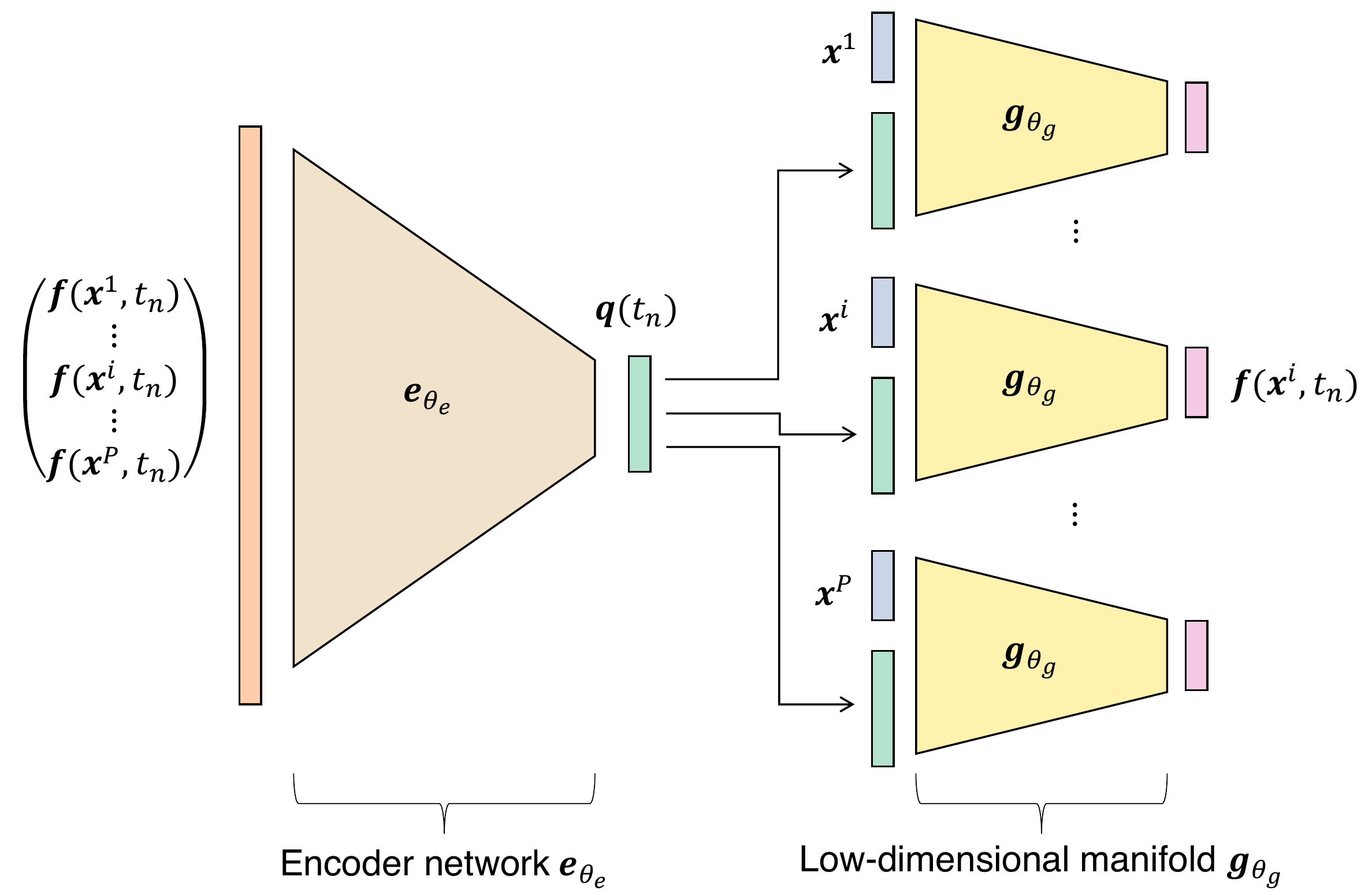}
    \caption{Constructing the manifold-parameterization function as a neural network trained via supervised learning. We pass each snapshot (time step) from the training dataset into an encoder to obtain a latent space vector $\latentSpaceVec$. We then concatenate $\latentSpaceVec$ with the spatial coordinates and pass that into the manifold-parameterization function with the goal of reconstructing $\fullOrderModel$ for each individual spatial sample. The same $\latentSpaceVec$ is shared among all spatial samples in this time step.}
    \label{img:training}
  \end{figure}

There are commonly two approaches to define the latent space vector $\latentSpaceVec$: the auto-decoder approach \citep{park2019deepsdf} that trains $\latentSpaceVec$ along with $\lowDimensionalManifoldNN$ and the encoder approach \citep{chen2019learning,mescheder2019occupancy} that trains a separate network to output $\latentSpaceVec$. While both approaches work for our application, we adopt the latter.

The encoder network $\encoder$ with weights $\weightsEncoder$ takes an input vector constructed by concatenating all the discrete degrees of freedom from the training data and outputs a latent space vector (see \Cref{img:training}): \[
    \encoder(\encInputArg{t}) = \latentSpaceVecArg{t}, \quad \text{where}\,\encInputArg{t} = \encInputDetailArg{t}.
    \]
We emphasize that this discretization-dependent encoder \citep{xie2021neural} is a tool for training the smoothly varying latent space \rev{and for determining the initial latent space vector from the discretized training data}. Otherwise, the implicit neural representation $\lowDimensionalManifoldNN$ remains a discretization-agnostic architecture.

Adding the encoder, \Cref{eqn:loss_decoder_only} now becomes
\begin{align}
    \min_{\weights,\weightsEncoder} 
    \sum_{i=1}^P \sum_{n=0}^T  \sum_{\params\in \paramDomainTrain} \|\lowDimensionalManifoldNNArgs{\discretePos}{\encoder(\encInputArg{\discreteTime})} - \fullOrderModelArgs{\discretePos}{\discreteTime}\|_2^2 \ .
    \label{loss}
\end{align}
\Cref{img:training} illustrates the training pipeline. Please refer to \Cref{sec:network_details} for network and training details and \Cref{sec:hyper-study} for hyperparameter selection.
\section{Method: Latent Space Dynamics}
\label{sec:latent-space-dynamics}

\begin{figure}
    \vspace{-12mm} %
    \centering
    \includegraphics[width=\textwidth]{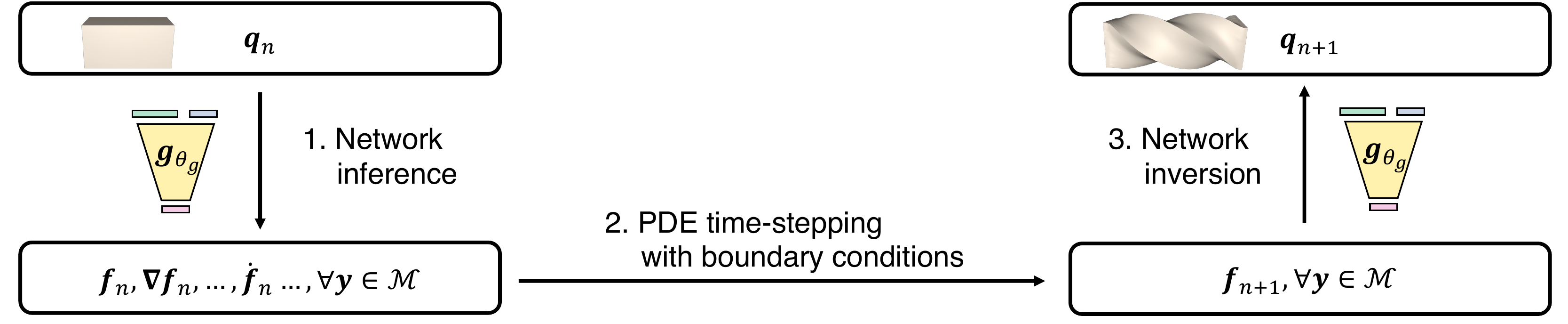}
    \caption{Latent space dynamics: temporally evolve from one latent space vector to another, governed by the PDE. The entire pipeline only involves degrees of freedom from a small spatial subset $\sampleSet$, where $\sampleSetCardinality\ll \nSpatial$.}
    \label{img:dynamics}
\end{figure}

After the manifold is constructed, we compute the low-dimensional latent space dynamics ($\discreteLatentSpaceVec\mapsto\discreteLatentSpaceVecPlus$) in three steps (see \Cref{img:dynamics,sec:pseudocode}): (1) network inference, (2) PDE time-stepping, and (3) network inversion. The neural network strictly serves as a \emph{kinematic spatial} representation in the maps from and to the latent space (steps 1 and 3, respectively). The time integration (step 2) itself uses the original PDE, not a neural network approximation thereof. The ROM community has demonstrated that these three steps can yield strong long-time stability even on stiff and chaotic dynamical systems \citep{carlberg2013gnat,carlberg2017galerkin}. \Cref{sec:latent_space_traj,sec:stability} demonstrate smooth latent space trajectories and empirical stability analysis using this approach.

Commonly shared among all three steps are ``integration samples'', a finite set of spatial domain points 
$\sampleSet \defeq \{\discretePosSample \in \spatialDomain \ |\  1\leq j \leq \sampleSetCardinality\}$ chosen at the user's discretion (see \Cref{sec:sample_choice}). These samples need not coincide with the previously mentioned full-order finite element discretization samples $\discretePosSet$.

\subsection{Step 1: Network Inference}
\label{sec:net-inf}
We first aim to gather all the full-space spatiotemporal information ($\forall\posSample\in\sampleSet$) necessary for PDE time integration. The function value $\fullOrderModel$ itself can be evaluated via inferencing of the neural network $\fullOrderModel(\posSample,\discreteTime)=\lowDimensionalManifoldNNArgs{\posSample}{\discreteLatentSpaceVec}$. The spatial and temporal gradients are computed either by differentiating the network,  $\gradb\fullOrderModel(\posSample,\discreteTime)=\gradbspatialSample\lowDimensionalManifoldNN$ and $\dot{\fullOrderModel}(\posSample,\discreteTime)=\lowDimensionalManifoldNNwrtLatentFlat\discreteLatentSpaceVecDot$,  respectively, or by numerical approximation. Higher-order gradients may be generalized in a similar manner. Further details on gradient computation are listed in \Cref{sec:net_grad}.

\subsection{Step 2: PDE Time-stepping}
\label{sec:PDE-time-stepping}
We now evolve time from $\discreteTime$ to $\discreteTimePlus$. Unlike end-to-end learning-based latent space dynamics methods \citep{lusch2018deep}, we evaluate time derivatives using the exact PDE \eqref{eqn:full-order-continuous}, \emph{not} a learned surrogate. 
At each integration point $\posSample$, we evaluate the temporal derivative by solving the PDE \eqref{eqn:full-order-continuous} for $\dot{\fullOrderModel}_{n+1}(\posSample)$:
\begin{align}
   \label{eqn:discrete-time}
   \pdelhs(\fullOrderModel_{n}, \gradb\fullOrderModel_{n}, \ldots, \dot{\fullOrderModel}_{n+1}, \ldots) = \0 \ .
\end{align}
We evolve the configuration to time $\discreteTimePlus = \discreteTime + \dt$ using 
the chosen explicit time integration method $\TimeIntegrate$ 
subject to given boundary conditions, e.g., Runge-Kutta methods \citep{dormand1980family}:
\begin{align}
    \label{eqn:time-integration}
    {\fullOrderModel}_{n+1} = \TimeIntegrate(\dt, \fullOrderModel_{n}, \dot{\fullOrderModel}_{n+1}, \ldots) \quad \ \forall\posSample\in\sampleSet \ .
\end{align}
While this work focuses on explicit time integration, we can also extend the framework for implicit time integration \citep{carlberg2017galerkin}.

\subsection{Step 3: Network Inversion}
\label{sec:net-inv}
We project back onto the reduced manifold by finding the corresponding input $\discreteLatentSpaceVecPlus$ that best matches the evolved configuration ${\fullOrderModel}_{n+1}$ in a least-squares sense: \citep{quarteroni2014reduced}
\begin{align}
    \label{eqn:invert}
    \min_{\discreteLatentSpaceVecPlus\in\RR{\nred}}\sum_{\posSample\in\sampleSet} \|\lowDimensionalManifoldNNArgs{\posSample}{\discreteLatentSpaceVecPlus} - \fullOrderModelArgs{\posSample}{\discreteTimePlus}\|_2^2 \ .
\end{align}
The objective is similar to the training loss found in \Cref{eqn:loss_decoder_only}, but with two dimensions significantly reduced: the dimension of the unknown $\discreteLatentSpaceVecPlus$, and the summation bound $\sampleSetCardinality$. Consequently, instead of using a stochastic gradient descent type of method, such as the auto-decoder scheme by \citet{park2019deepsdf}, we achieve rapid inversion using the Gauss-Newton algorithm \citep{nocedal2006numerical} with conditionally quadratic convergence. Further details are listed in \Cref{sec:net-inversion-details}.

\subsection{Spatial Sample Reduction}
\label{sec:sample_choice}
\rev{A necessary condition for the well-posedness} of the least squares formulation from \Cref{eqn:invert} is $\nred\leq\dimensionOut\sampleSetCardinality$. Since the manifold-parameterization function construction guarantees that $\nred\ll \nSpatial\dimensionOut$, we choose $\tfrac{\nred}{\dimensionOut}\leq\sampleSetCardinality\ll\nSpatial$. To obtain the next-time step $\fullOrderModelArgs{\posSample}{\discreteTimePlus},\forall\posSample\in\sampleSet$ necessary for the least squares solves, we only require PDE updates (\Cref{sec:PDE-time-stepping}) and spatiotemporal data (\Cref{sec:net-inf}) at these $\sampleSetCardinality$ samples. As such, the entire latent space dynamics framework (\Cref{img:dynamics}) requires only $\sampleSetCardinality$ samples, compared to the full-order solver's $\nSpatial$ samples. Hyper-reduction approaches like this have captured a wide range of real-world scenarios, including massive elasticity deformations \citep{fulton2019latent} and large turbulent flows \citep{grimberg2021mesh}. Unlike our discretization-independent approach, prior methods only support hyper-reduction samples that coincide with the full-order discretization.

\begin{figure}
    \vspace{-12mm}
    \centering
    \includegraphics[width=\textwidth]{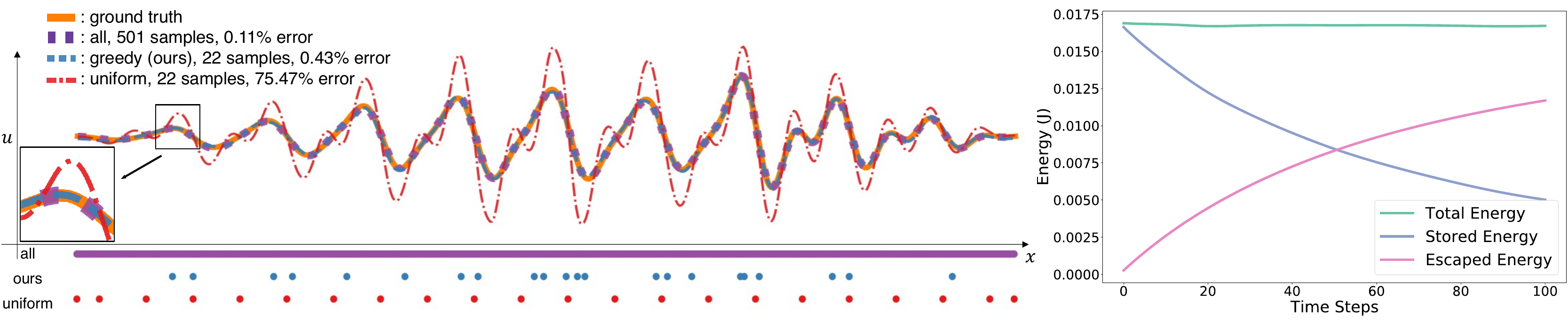}
    \vspace{-8mm}
    \caption{Thermodynamics. Left: integration samples ($\sampleSet$). Greedily selecting the samples (blue) allows us to use significantly fewer degrees of freedom than the full-order simulation (purple) while getting a much higher accuracy than naive uniform sampling (red). The depicted field is temperature governed by the heat equation after $100$ time steps. Right: as dictated by the PDE, the reduced system is conservative: total energy (stored thermal energy plus cumulative flux at boundary) is conserved.}
    \label{img:heat_all}
\end{figure}  

A naive selection of the integration samples can lead to inaccurate latent space dynamics, even if $\sampleSetCardinality\geq\tfrac{\nred}{\dimensionOut}$; refer to \Cref{img:heat_all} for the failure case of uniform sampling. As noted in the hyper-reduction literature, stochastic sampling can eliminate such errors \citep{carlberg2011model}. To better \emph{control} hyper-reduction error, we draw inspirations from the cubature approach by \citet{an2008optimizing} and propose a greedy algorithm that augments the sample set to meet a target residual (see \Cref{sec:optimal-sampling}). \rev{This strategy iteratively adds spatial samples to the sampling set until the target residual is met. In this case, the user decides the target accuracy, not the sample set size $\sampleSetCardinality$. Alternatively, the user can terminate the algorithm early once it reaches a chosen sample set size $\sampleSetCardinality$.} Results of our sampling approach are shown in \Cref{img:heat_all,img:roten,img:image-smoothing}.

\section{Experiments}
\label{sec:results}
We analyze the proposed framework on classic PDEs, with training data produced using a variety of discretizations (voxel grids, meshes, and point clouds). Unless otherwise noted, for each PDE, we delineate a testing set where $\paramDomainTest\subset\paramDomain$ with
$\paramDomainTrain\cap\paramDomainTest=\emptyset$. We construct the manifold (\Cref{sec:manifold-construction}) with data from $\paramDomainTrain$, and then validate the latent space dynamics (\Cref{sec:latent-space-dynamics}) on $\paramDomainTest$. We compare our approach with prior discretization-dependent ROM methods, including POD \citep{berkooz1993proper,holmes2012turbulence} and neural-network-based autoencoder approaches \citep{fulton2019latent,lee2020model,shen2021high}\rev{, under the same latent space dimensions}. Additional implementation and reproducibility details are listed from \Cref{sec:thermo_details} to \Cref{sec:solid_details}. Experiment statistics are summarized in \Cref{tab:dimension-reduction}. The temporal evolutions of the PDEs are best illustrated via the \textbf{supplementary video}.

\noindent
\textbf{Thermodynamics}, $\pdflat{u}{t}-\nu(x)\pdflattwo{u}{x} = 0\ .$ Temperature $u$ is governed by a one-dimensional heat equation. $\nu$ describes the spatially-varying diffusion speed. \Cref{img:heat_all} displays our approach's ability to use very few integration samples and to capture conservation of energy.

\begin{figure}
\centering
\vspace{-12mm}
\includegraphics[width=0.8\textwidth]{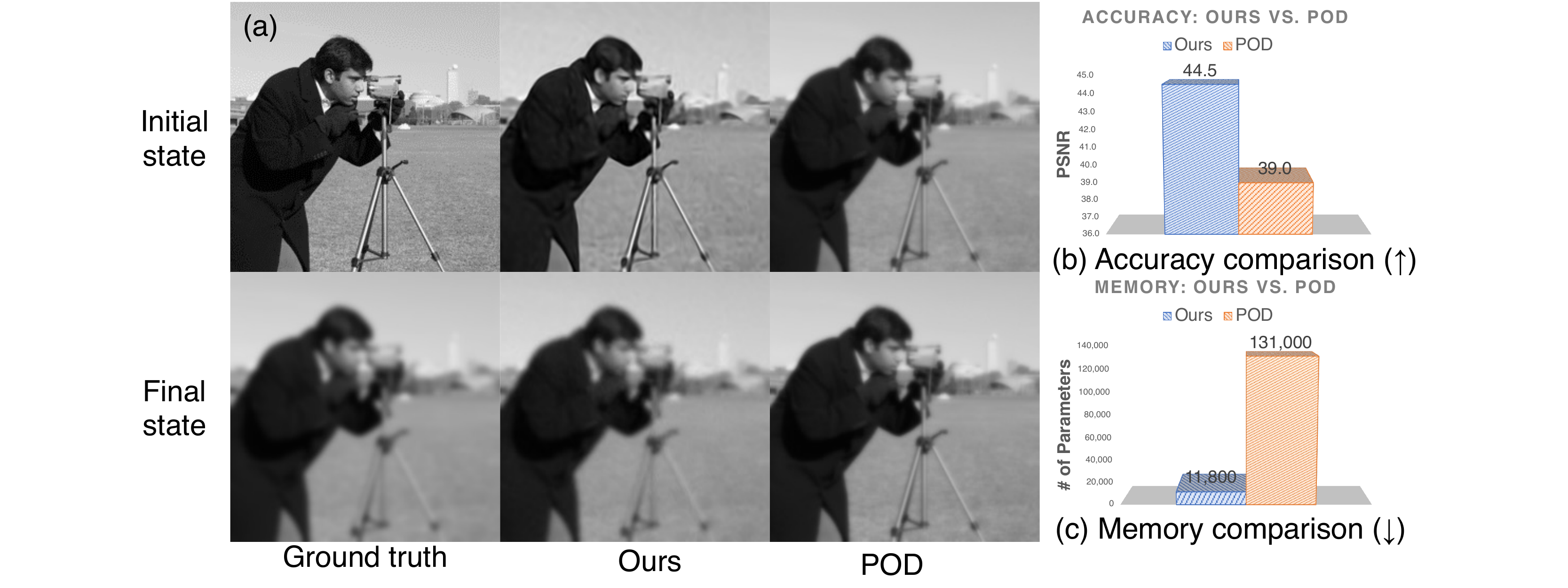}
\vspace{-4mm}
\caption{Image processing: comparison with POD. Ground truth solution uses $P = 65, 536$ pixels. (a) Visually, our approach better captures the sharp initial state and the smoothed finial state than POD. (b) Quantitatively, CROM obtains a higher PSNR than POD ($\uparrow$ the higher the better). (c) Our approach also uses an order-of-magnitude less memory than POD ($\downarrow$ the lower the better). Both POD and our approach use the same latent space dimension ($\nred=3$).}
\label{img:vs_prior:imgprocess}
\end{figure}

\noindent
\textbf{Image Processing}, $\quad\pdflat{u}{t}-\nu(\xb)\gradb^2u = 0 \ .$ We model image blurring with the 2D diffusion equation \citep{perona1990scale}. \Cref{img:vs_prior:imgprocess} shows that under the same latent space dimension ($\nred=3$), our method is more accurate than POD \citep{berkooz1993proper,holmes2012turbulence}, both visually and quantitatively. With its architecture independent of pixel count, CROM uses an order of magnitude less memory than POD (see \Cref{img:vs_prior:imgprocess}c).

\begin{figure}
    \centering
    \vspace{-4mm}
    \includegraphics[width=0.8\textwidth]{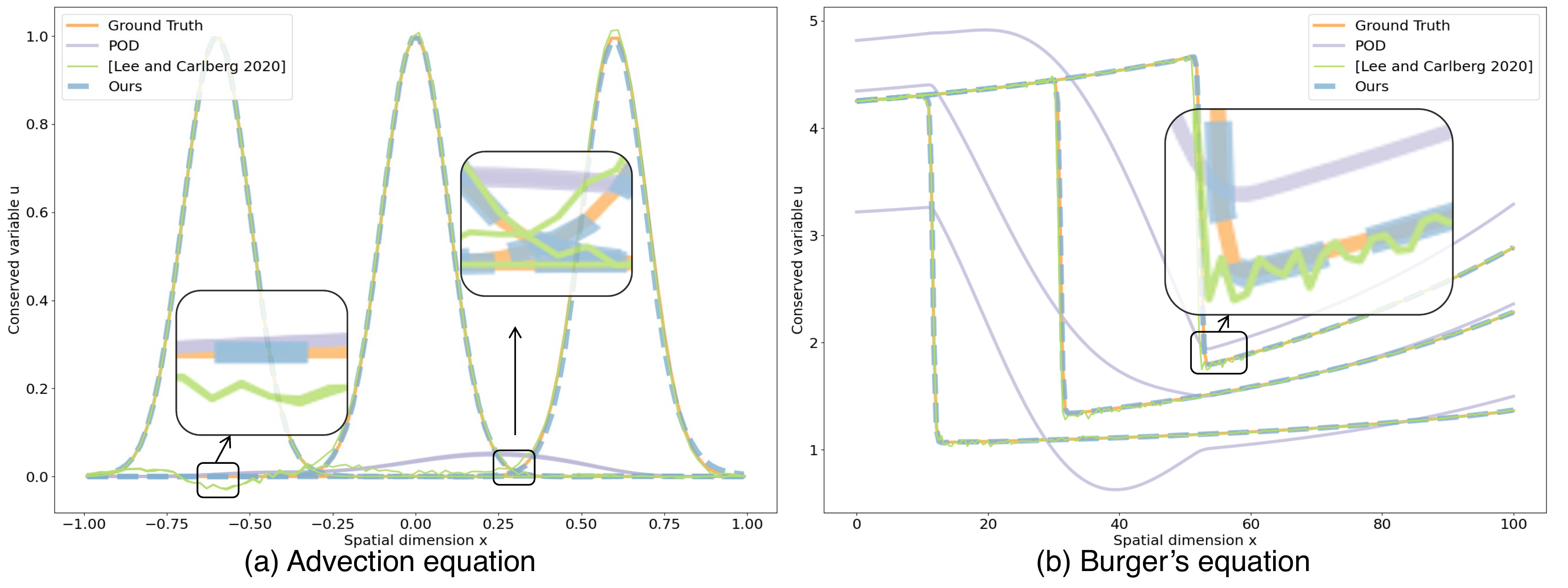}
    \vspace{-4mm}
    \caption{Breaking the Kolmogorov barrier: CROM outperforms POD and convolutional autoencoders \citep{lee2020model} in tracking both the Advection and Burgers' trajectories. For both cases, we use the intrinsic solution-manifold dimension \citep{lee2020model}, the lower bound of $\nred$, as the latent space dimension $\nred$. To isolate the source of error, no hyper-reduction method is applied.\vspace{-4mm}}
    \label{img:advection_burgers}
\end{figure}

\noindent
\textbf{Transport dominated systems}.
Next, we examine two transport-dominated slowly decaying Kolmogorov n-widths problems, the Advection Equation and Burgers' Equation, where classic model reduction techniques often struggle \citep{peherstorfer2022breaking}:

\noindent
\textbf{Advection Equation}, $\pdflat{\advectQuantity}{t} + (\advectSpeed\cdot\grad)\advectQuantity=0 \ .$
Here $\advectQuantity$ is the advected quantity and $\advectSpeed$ is the advection velocity. \Cref{img:advection_burgers}a depicts CROM's favorable trajectory tracking relative to both POD and the convolutional autoencoder (CAE) of \citet{lee2020model}, for equal latent space dimension (also see \Cref{img:advection}).

\noindent
\textbf{Burgers' Equation}, $\pdflat{\burgerQuantity}{t} + \pdflat{0.5\burgerQuantity^2}{x}=0.02e^{\burgerD x} \ .$ \Cref{img:advection_burgers}b shows that CROM more accurately captures the nonlinear dynamics than both POD and CAE (also see \Cref{img:relative_error}). Additionally, CROM uses $12\times$ less memory than CAE (see \Cref{img:vs_prior_cnn:memory}). These accuracy and memory advantages are consistent with the implicit neural representation literature \citep{chen2019learning}. 

\begin{figure}
    \centering
    \vspace{-4mm}
    \includegraphics[width=\textwidth]{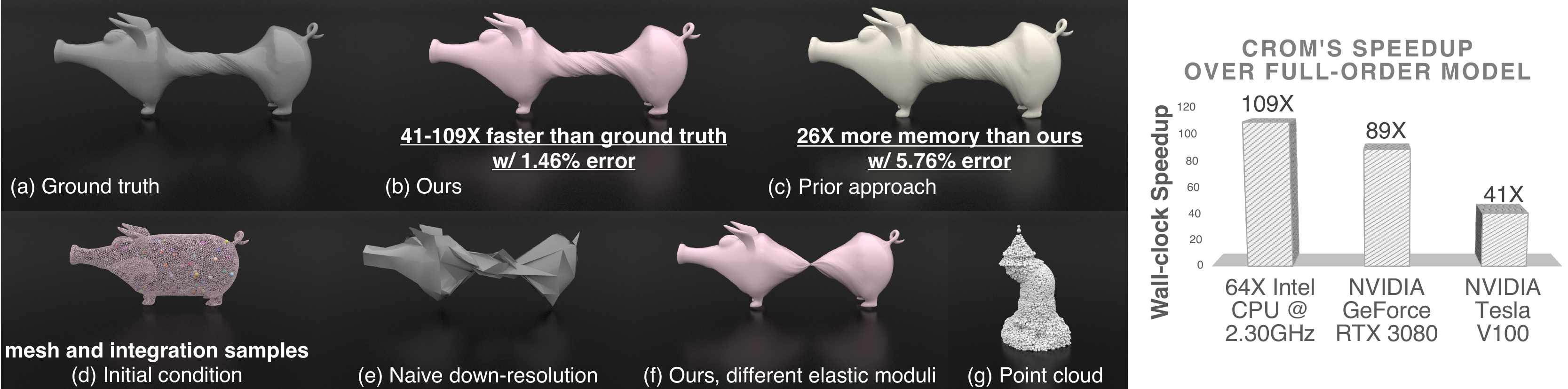}
    \vspace{-6mm} %
    \caption{Solid mechanics (a) The ground truth is generated via the full-order PDE solver. (b) Our approach is 41-109$\times$ faster than the ground truth while capturing detailed shearing and volume-preserving behaviors ($\cromPiggyError$ error). (c) Prior approach \citep{barbivc2012fem} of the same latent space dimension consumes 26$\times$ more memory while suffering from volume-gain artifacts ($\podPiggyError$ error). (d) These simulations adopt a tetrahedral discretization (d, pink mesh, $\nSpatial=\piggyVertexNumber$). Instead of using the expensive high-resolution mesh, our approach computes dynamics using very few integration samples (d, colorful spheres, $\sampleSetCardinality=\piggyIntegrationSamples$). (e) Naive down-resolution of the ground truth simulation yields a similar runtime but leads to significantly worse quality. (f) After training, our model can capture a wide range of material properties. (g) The same manifold-parameterization function architecture (different network weights) can also be used for model-reducing point-cloud based simulation (reproduced from the work by \citep{chen2023model}). The speedup plot on the right demonstrates the effectiveness of our approach on diverse computing platforms. Disclaimer: the authors do not support animal cruelty.
    }
    \label{img:roten}
  \end{figure}

Whereas CAE is applicable only to voxel grid data, CROM is applicable to data from \emph{any discretization}, as highlighted by the tetrahedral mesh and point cloud data examples (see \Cref{img:roten}).

\begin{figure}
  \centering
  \vspace{-4mm}
  \includegraphics[width=0.9\textwidth]{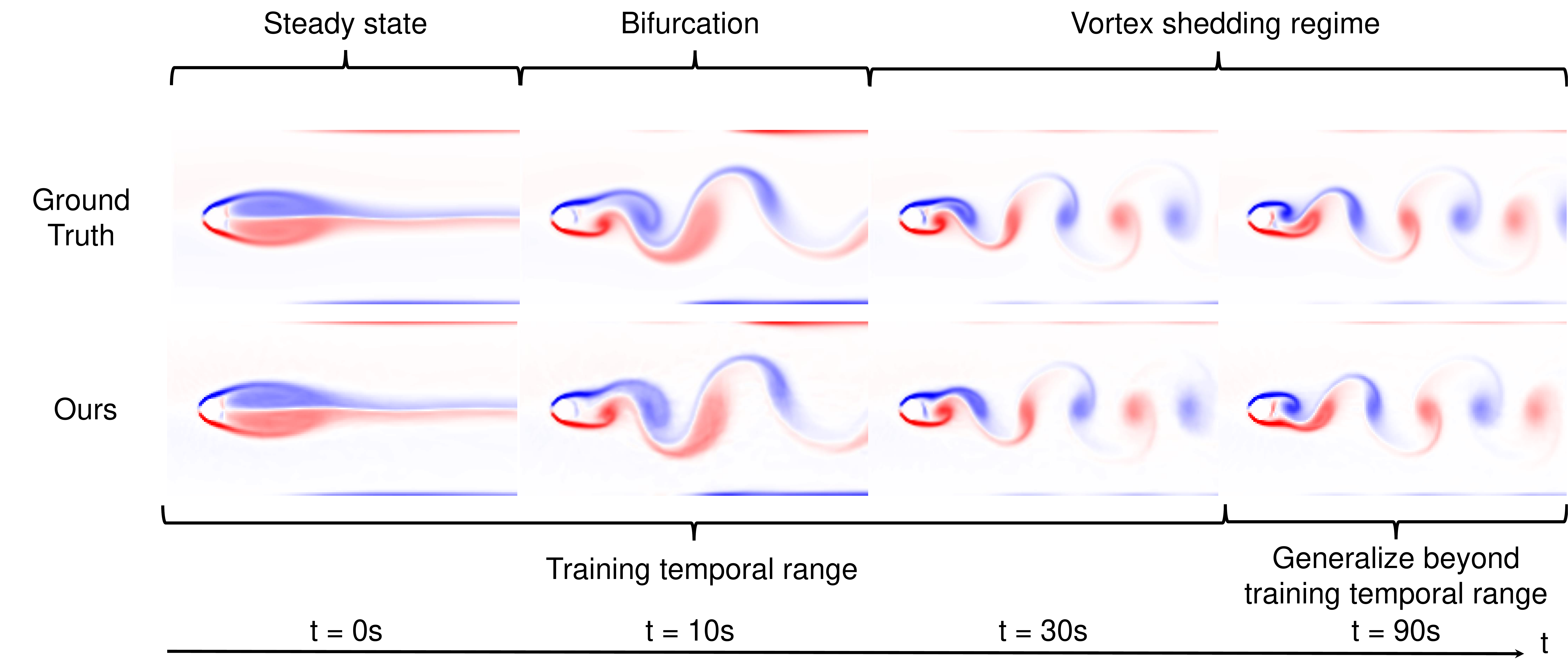}
  \vspace{-2mm}
  \caption{\rev{The Karman vortex street. Our latent space dynamics solver accurately captures the flow both before and after the bifurcation: the steady state and the periodic vortex shedding regime. The visualized quantity is the the curl of the velocity field. Our approach even generalize well beyond the training temporal range, thanks to the kinematics-approximation-only nature of CROM.}}
  \label{img:fluid:karman}
\end{figure}

\noindent
\textbf{Fluid Dynamics}, $\pdflat{\fluidVelo}{t} + (\fluidVelo\cdot\gradb)\fluidVelo = -\gradb \pressure + \nu\gradb^2\temperatureVec + \externalForce, \, \gradb \cdot \fluidVelo = 0.$ CROM captures the challenging Karman vortex street (see \Cref{img:fluid:karman}), governed by the incompressible Navier-Stokes equations. It even generalizes well beyond the training temporal range.

\begin{figure}
    \centering
    \vspace{-8mm}
    \includegraphics[width=\textwidth]{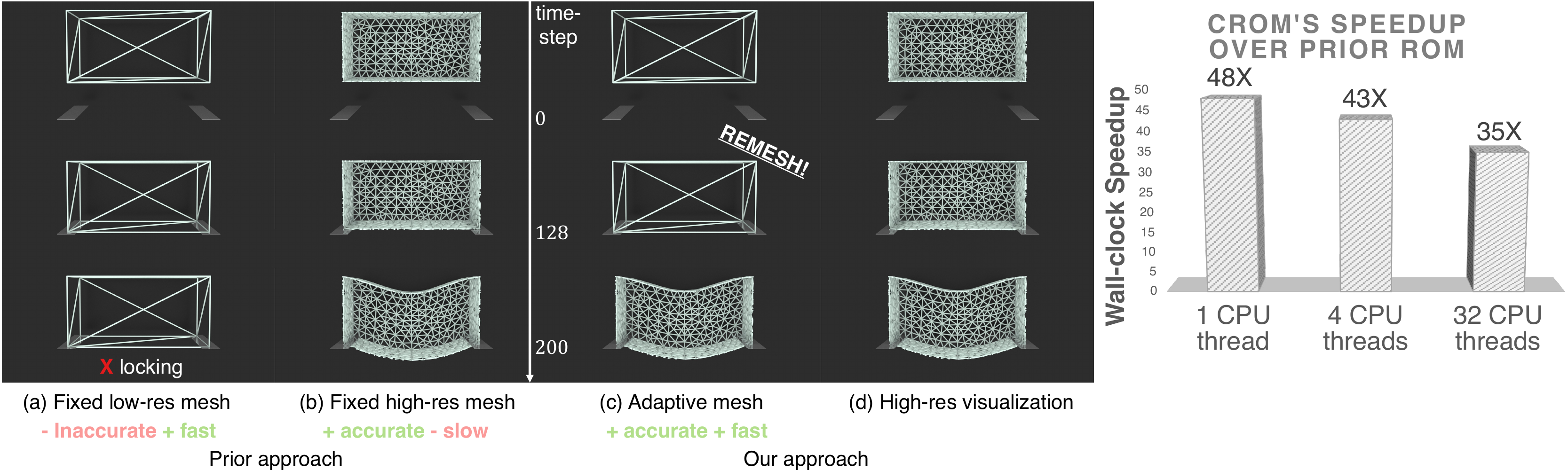}
    \vspace{-6mm} %
    \caption{Adaptive discretization in solid mechanics. A falling deformable body impacts two static objects. (a and b) Prior approaches \citep{barbivc2005real,fulton2019latent,shen2021high} only support reduced-order dynamics with fixed discretizations, forcing a tough choice between accuracy and computational cost; (c) our approach allows dynamic mesh adaptivity to balance accuracy and cost as problem difficulty varies over time. (d) Our resolution-independent encoding of the deformation map allows us to visualize the field using any method (e.g., a high resolution mesh) without regard to the computational mesh. CROM's speedup over prior discretization-dependent ROM methods is observed for 1–32 threads, and emphasized in the case of limited computational resources. To ensure representativeness of prior ROM architectures, we report timing strictly for the PDE-time-stepping stage (Step 2), which is shared among all prior ROM architectures.}
    \label{img:falling_remeshing}
  \end{figure}

\begin{figure}
    \centering
    \vspace{-14mm}
    \includegraphics[width=\textwidth]{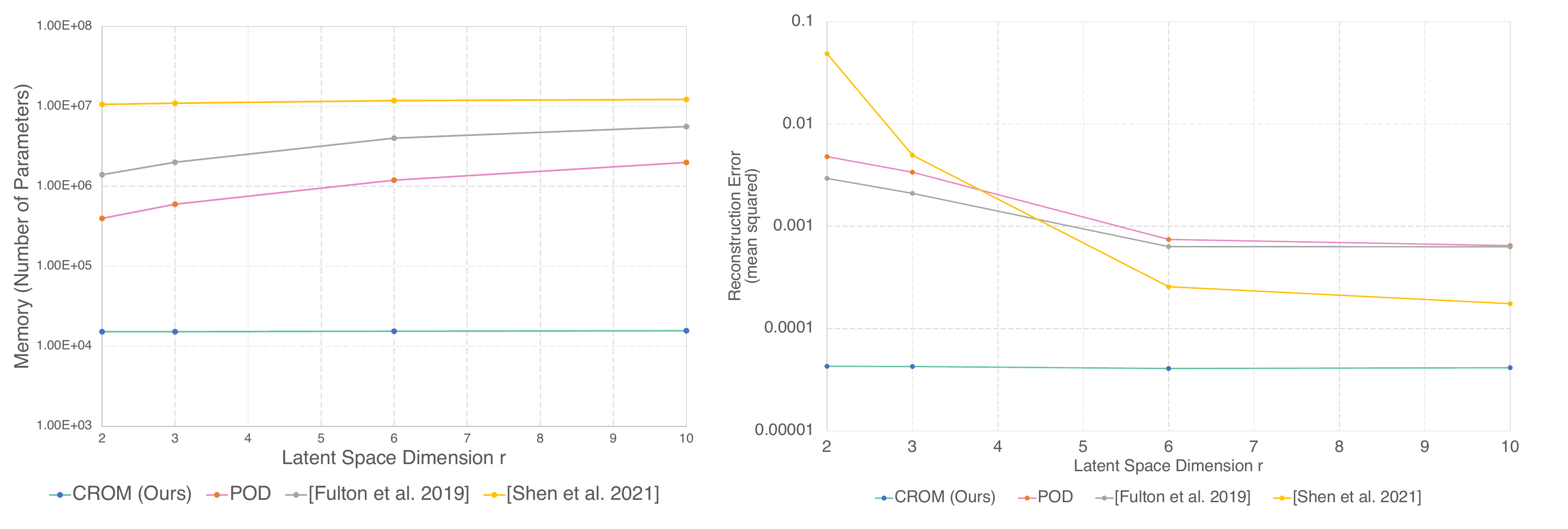}
    \vspace{-8mm} %
    \caption{Our approach uses significantly less memory and simultaneously yields higher accuracy than POD and the autoencoder-based ROM methods of \citet{fulton2019latent} and \citet{shen2021high} for the solid mechanics experiment depicted in \Cref{img:roten}. For example, with $\nred=3$, our approach uses 39$\times$ and 132$\times$ less memory, while simultaneously offering 79$\times$ and 49$\times$ more accuracy, compared to POD and the autoencoder approach of \citet{fulton2019latent}, respectively.\vspace{-2mm}}
    \label{img:mem_and_accuracy}
\end{figure}

\noindent
\textbf{Solid Mechanics}, $\rho_0 \ddot{\deformationMap} = \gradb \cdot \PKStress(\gradb \deformationMap )  + \rho_0\bodyForce \ .$
We solve the second-order elastodynamics equation for the deformation map $\deformationMap$ of soft bodies, where $\rho_0$ is the initial density, $\PKStress$ is the first Piola–Kirchhoff stress, and $\bodyForce$ is the body force density (see \Cref{img:falling_remeshing,img:roten,img:mem_and_accuracy,img:dino_and_dragon}).

\noindent
\textbf{Adaptive resolution.}
Prior ROM methods (see \Cref{img:vs_classic}a) fix the discretization at onset, precluding the dynamic adaptivity approaches that benefit problems with evolving complexity. When modeling a falling block (see \Cref{img:falling_remeshing}), prior ROM methods must begin with a high-resolution mesh, despite that the high-resolution is required only during contact (\Cref{img:falling_remeshing}a and b). Our discretization-independent representation allows the time integrator to freely adapt the spatial discretization throughout the dynamic evolution. For instance, we can employ a coarse mesh ($\nSpatial=\remeshLoVertexNumber$ vertices) to economize computation during the rigid falling phase (\Cref{img:falling_remeshing}c), and a high-resolution mesh ($\nSpatial=\remeshHiVertexNumber$) to capture details during contact (\Cref{img:falling_remeshing}c), yielding 35$\times$–48$\times$ wall-clock speedups over fixed discretization ROM methods. To ensure generalizability of the comparison, the timing is done strictly for the PDE time-stepping stage (Step 2). Prior ROM architectures (i.e, both POD and autoencoder approaches) share the same PDE time-stepping routine.

\noindent
\textbf{Memory footprint.}
Most prior ROM approaches are discretization-dependent. As discretization resolution ($\nSpatial$) increases, so does memory consumption for $\lowDimensionalManifold$. By contrast, our discretization-independent architecture does not ``see'' the parameter $\nSpatial$ and in practice memory footprint does not scale with $\nSpatial$. Rather, as with all neural fields \citep{xie2021neural}, memory grows with the intrinsic complexity of the training data.

\Cref{img:mem_and_accuracy} compares memory footprint and reconstruction error for CROM, POD \citep{barbivc2012fem}, and neural-network-based autoencoder approaches \citep{fulton2019latent,shen2021high}, as a function of latent space dimension; \Cref{img:roten} depicts the corresponding large deformation simulations. \Cref{img:dino_and_dragon} compares CROM's memory and accuracy on another large deformation example. For these examples, CROM offers simultaneously lower memory consumption \emph{and} reconstruction error. Furthermore, due to hyper-reduction (from $\nSpatial=\piggyVertexNumber$ to $\sampleSetCardinality=\piggyIntegrationSamples$), CROM also obtains significant wall-clock speedups on CPUs, consumer GPUs, and data center GPUs over the unreduced model (see \Cref{img:roten}).

\noindent
\textbf{Different discretizations, same architecture.}
Because CROM is not constructed for a fixed discretization, we can adopt an identical manifold-parameterization function architecture to model-reduce simulations of both tetrahedral meshes and point clouds (see \Cref{img:roten}g). Moreover, even though these examples (\Cref{img:falling_remeshing,img:roten}) leverage completely different meshes, they use the same network architecture. This opens the door for future work on transfer learning \citep{weiss2016survey} and weight sharing among various discretizations and PDEs.

Additional comparisons between CROM, previous ROM, and the full-order models are discussed in \Cref{sec:vs-prior,sec:performance_eval}.

\section{Discussion and Conclusion}
CROM is a model reduction framework for PDEs featuring a discretization-independent architecture. CROM disentangles the low-dimensional embedding from the discretization by reformulating the reduced manifold as a map accepting not only $\latentSpaceVec$ but also $\pos$ as an input. CROM outperforms  discretization-dependent ROM approaches, such as POD and autoencoders, in terms of accuracy and memory consumption for equal latent space dimension, and in the ability to dynamically adapt the discretization during time integration.

While offering key advantages over prior ROM methods, CROM also inherits a limitation of ROM. It can only treat PDE solutions in the space spanned by the manifold-parameterization function (determined by the training data) and does not generalize to arbitrary unseen scenarios. Such a limitation is also commonly found among other implicit neural representation works \citep{xie2021neural}. Future research may consider improving generalizability and data efficiency via meta-learning and integration of stronger priors \citep{sitzmann2020metasdf}.

Compared to end-to-end ML solutions to PDEs \citep{sanchez2020learning}, CROM employs the neural network strictly as a spatial representation (see \Cref{sec:net-inf,sec:net-inv}) for the kinematics and solves the PDE using classical PDE-integration numerical methods (see \Cref{sec:PDE-time-stepping}). As such, we believe CROM will open doors for more forthcoming hybrid ML-PDE solutions. As shown in our work, these solutions can retain the PDE's physical invariants (see \Cref{img:heat_all}), allow for easy integration with existing PDE solvers (see \Cref{sec:PDE-time-stepping}), and obtain practical computational savings that can be directly employed in production (see \Cref{img:roten}).
\subsubsection*{Acknowledgments}
This work was supported in part by the National Science Foundation (Grant CBET-17-06689), Meta, Natural Sciences and Engineering Research Council of Canada (Discovery Grant), and SideFX. We thank Honglin Chen and Rundi Wu for their insightful discussions. We thank Rundi Wu for sharing his implementation of Stable Fluids. We thank Raymond Yun Fei for rendering advice. We thank Keenan Crane for the "Origins of the Pig" mesh.

\clearpage %

\bibliography{iclr2023_conference}
\bibliographystyle{iclr2023_conference}

\newpage
\appendix
\part{Appendix}

We highly encourage viewing the \textbf{supplementary video} where our results are best illustrated.

\section{Additional Literature Review}
\label{sec:lite_rev}
\subsection{Reduced-Order Modeling}
Accelerating PDE solutions via reduced-order modeling (ROM) has a rich history. Two common building blocks of ROM are (1) identifying a low-dimensional latent space of the original complex system and (2) solving PDEs via latent space dynamics, i.e., evolving the latent space vector over time.

Early works on latent space construction started with linear methods where the the latent space and the full space has a linear relationship. Notable works include proper orthogonal decomposition \citep{berkooz1993proper,holmes2012turbulence,barbivc2005real,barbivc2012fem}, the reduced-basis technique \citep{Prud'homme200270,Rozza20071}, balanced truncation \citep{Moore198117}, rational interpolation \citep{baur2011interpolatory, gugercin2008h_2}, and Craig–Bampton model reduction \citep{craig1968coupling}. Recently, nonlinear methods \citep{8062736,7799153,gu2011model,erichson2019physics,maulik2020time,regazzoni2019machine,fulton2019latent,maulik2021reduced,romero2021learning,shen2021high}, often constructed via neural networks, have gained significant attentions due to their ability to more accurately construct the manifold-parameterization function. In particular, nonlinear manifolds significantly outperform their linear counterparts on slowly decaying Kolmogorov n-width problems (e.g., advection) \citep{peherstorfer2022breaking,ohlberger2013nonlinear,peherstorfer2015online,taddei2015reduced,peherstorfer2020model,ehrlacher2020nonlinear,lee2019deep}.

Most of these prior works exclusively focus on building a latent space for the \emph{already-dicretized} vector fields of interest. \citet{chen2023model,pan2022neural} and our work are the first attempts at constructing the latent space for the continuous vectors themselves, thereby allowing the training data to come from any discretizations.

After the latent space is identified, solving PDEs with parameters unseen during training requires evolving the latent space vector over time. Two families of approaches have been proposed for latent space dynamics. The first family of approaches \citep{kim2019deep} computes latent space dynamics via a data-driven operator that is learned from the training data. A key benefit from learning the operator from data is that the user requires no knowledge of the system's governing equations and does not need access to the training solver's source code. However, the downside of this approach is that additional assumptions (e.g., the availability of a Koopman invariant subspace as in \citet{lusch2018deep} and the existence of a finite-dimensional quadratic representation as it is in the lift and learn approach by \citet{qian2020lift}) are made. Without these assumptions, the learned operator is prone to violation of physical laws. The second family of approaches \citep{lee2020model,fulton2019latent}, to which our method belongs, computes latent space dynamics according to the PDE used for training data generation. The advantage of this approach is that it does not require learning an additional time-evolution operator. It works with arbitrary dynamical systems with known PDE governing equations and conserves physical laws by construction since explicit PDE model is employed. However, the disadvantage is that the PDE model must be known.

\section{Robust Sampling}
\label{sec:optimal-sampling}
To select the integration samples robustly, we aim to balance the computation speed and accuracy. CROM employs $\sampleSetCardinality$ spatial samples, and the computation speed roughly scales linearly with the number of samples. To maximize computation speed, we aspire to select the fewest number of samples possible.

Given a target accuracy, our greedy algorithm selects the fewest number of integration samples from the discrete samples $\discretePosSet$ of the full-order PDE solution in order to achieve the target accuracy.

\begin{algorithm}[H]
    \SetAlgoLined
    \SetKwProg{Fn}{Function}{:}{}
    $\sampleSet = \{\xb^{k}\}$ // randomly initialize integration sample set, $k\in\{1,\ldots,\nSpatial\}$  \\
    
    \While {true} {
        $res\_vec = Calculate\_Residual(\sampleSet)$ \\
        \uIf{$Metric(res\_vec) <$ target\_accuracy}{
            \KwRet $\sampleSet$
        }
        \Else{
            $indices$ = $Max\_Indices$($res\_vec$, $\numtry$)
            // find $\numtry$ indices with the largest individual residuals\\
            $res\_list = \mathbf{0}\in\RR{\numtry}$ \\
            \For{$i$ in $indices$} { 
                $res\_list[i] = Metric(Calculate\_Residual(\sampleSet\cup\{\xb^{i}\}))$ \\
            }
            $j = Min\_Indices(res\_list, 1)$ // find the index with the smallest global residual\\
            $\sampleSet = \sampleSet \cup \{\xb^{j}\}$
        }
    }

    \Fn{$Calculate\_Residual(\sampleSet)$}{
        $res\_vec = \mathbf{0}\in\RR{\nSpatial}$ \\
        \For{$\params\in\paramDomainTrain$}{
            Obtain the final latent space vector $\latentSpaceVecArg{\discreteEnd; \params}$ with the integration samples ($\sampleSet$)\\
            \For{$\spatialIndexSet$}{
              $res\_vec[i] += \|\lowDimensionalManifoldNNArgs{\discretePos}{\latentSpaceVecArg{\discreteEnd; \params}} - \fullOrderModelArgs{\discretePos}{\discreteEnd}\|$
            }
        }
        \KwRet $res\_vec$
    }

     \Fn{$Metric(res\_vec)$}{
        \KwRet $mean(res\_vec) + max(res\_vec)$ 
    }
    \caption{Robust sampling}
    \label{alg:sampling}
\end{algorithm}
In every iteration, the greedy algorithm adds one spatial sample to the sample set $\sampleSet$ and lowers the error of latent space dynamics. Specifically, the algorithm loops over $\numtry$ spatial samples with the largest individual residuals. We test adding each sample to the integration sample set and compute a new global residual. From these $\numtry$ samples, we select the one leading to the smallest new global residual and add it to the actual sample set. The algorithm repeats until the target accuracy is met. In practice, we find $\numtry=10$ gives sufficient results. \Cref{alg:sampling} describes the details. Compared to the naive uniform sampling approach, our robust sampling scheme yields significantly more accurate results (see \Cref{img:heat_all}). We also highlight that \Cref{alg:sampling} happens before any simulation occurs (latent space dynamics). It is strictly a pre-computation.
\section{Network Inversion}
\label{sec:net-inversion-details}
In order to generate latent space dynamics, we need to invert the neural network employed for the manifold-parameterization function. We do so by solving the minimization problem:
\begin{align}
    \min_{\discreteLatentSpaceVecPlus\in\RR{\nred}}\sum_{\posSample\in\sampleSet} \|\lowDimensionalManifoldNNArgs{\posSample}{\discreteLatentSpaceVecPlus} - \fullOrderModelArgs{\posSample}{\discreteTimePlus}\|_2^2.
\end{align}
Since this is a nonlinear least-squares problem, we solve it with the classic iterative Gauss-Newton algorithm \citep{nocedal2006numerical}. Unlike Newton's method, Gauss-Newton does not require the Hessian of the objective function. Consequently, in addition to the neural network itself, we only need to evaluate the gradient of the neural network with respect to the latent space vector $\lowDimensionalManifoldNNwrtLatentFlat(\posSample,\discreteLatentSpaceVecPlus)$. We also equip our Gauss-Newton implementation with a standard backtracking line-search scheme to improve the convergence rate.

\subsection{Initial Guess for the Nonlinear Solver}
\label{sec:initial_guess}
An initial guess close to the minimum is necessary for the convergence of Gauss-Newton. The previous time step latent space vector $\discreteLatentSpaceVec$ serves as an ideal initial guess since $\fullOrderModel$ does not vary much over a time step. However, in the extreme cases where $\fullOrderModel$ deviates significantly over one time step, we find it helpful to compute the initial guess through an encoder network. For convenience, we employ the encoder network from training, which requires computing the dynamics for $\nSpatial$ samples. Future work should consider building this encoder network strictly for the degrees of freedom in $\sampleSet$. We can further facilitate nonlinear solver convergence by \rev{encouraging smoothness} of the latent space, such as employing a Jacobian penalty term \citep{chen2023model} / local isometry term \citep{du2021learning} and enforcing Lipschitz constraints \citep{liu2022learning}. \rev{These smoothness regularizations would be particularly important for challenging PDE cases involving bifurcations and buckling \citep{brush1975buckling}.}

\subsection{Linearization}
We can further obtain computation-save by bypassing the iterative nonlinear solver and linearizing the least-squares problem.

We define $\deltaFullOrder$ as,
\begin{align}
    \deltaFullOrderArgs{\posSample}{\discreteTimePlus} &= \fullOrderModelArgs{\posSample}{\discreteTimePlus} - \fullOrderModelArgs{\posSample}{\discreteTime}\\
    &= \fullOrderModelArgs{\posSample}{\discreteTimePlus} - \lowDimensionalManifoldNNArgs{\posSample}{\discreteLatentSpaceVec}.
\end{align}

Employing Taylor expansion of $\lowDimensionalManifoldNN$ at $\discreteLatentSpaceVec$, we have $\lowDimensionalManifoldNNArgs{\posSample}{\discreteLatentSpaceVecPlus} \approx \lowDimensionalManifoldNNArgs{\posSample}{\discreteLatentSpaceVec} + \lowDimensionalManifoldNNwrtLatentFlatArgs{\posSample}{\discreteLatentSpaceVec}\deltaDiscreteLatentSpaceVecPlus$, where $\discreteLatentSpaceVecPlus = \discreteLatentSpaceVec + \Delta\discreteLatentSpaceVecPlus$.
Therefore, the original objective function can be approximated as,
\begin{align}
    \lowDimensionalManifoldNNArgs{\posSample}{\discreteLatentSpaceVecPlus} - \fullOrderModelArgs{\posSample}{\discreteTimePlus} &=
    \lowDimensionalManifoldNNArgs{\posSample}{\discreteLatentSpaceVecPlus} - \lowDimensionalManifoldNNArgs{\posSample}{\discreteLatentSpaceVec} - \deltaFullOrderArgs {\posSample}{\discreteTimePlus}\\
    &\approx \lowDimensionalManifoldNNwrtLatentArgs{\posSample}{\discreteLatentSpaceVec}\deltaDiscreteLatentSpaceVecPlus - \deltaFullOrderArgs{\posSample}{\discreteTimePlus}.
\end{align}

Therefore, we can obtain the latent space dynamics by solving the \emph{linear} least-squares problem,

\begin{align}
    \min_{\deltaDiscreteLatentSpaceVecPlus\in\RR{\nred}}\sum_{\posSample\in\sampleSet} \|\lowDimensionalManifoldNNwrtLatentArgs{\posSample}{\discreteLatentSpaceVec})\deltaDiscreteLatentSpaceVecPlus - \deltaFullOrderArgs{\posSample}{\discreteTimePlus}\|_2^2.
\end{align}

This is the normal equation and can be solved in the closed form:
\begin{align}
    \deltaDiscreteLatentSpaceVecPlus = (\Jb^T\Jb)^{-1}\Jb^T\bb,
\end{align}
where $\Jb$ is the $(\sampleSetCardinality\cdot\dimensionOut)$ by $\nred$ Jacobian matrix that contains $\lowDimensionalManifoldNNwrtLatentFlatArgs{\posSample}{\discreteLatentSpaceVec}, \forall\posSample\in\sampleSet$; $\bb$ is the $(\sampleSetCardinality\cdot\dimensionOut)$ by $1$ residual vector containing $\deltaFullOrderArgs{\posSample}{\discreteTimePlus}, \forall\posSample\in\sampleSet$.

Since our work features both small $\sampleSetCardinality$ and small $\nred$, the closed-form network inversion costs just a few small dense matrix multiplications and thereby introduces minimal overhead. In fact, the solver itself is extremely efficient, and the majority of the computation cost is forming the Jacobian matrix $\Jb$ itself, whose optimization will be discussed in \Cref{sec:nn-efficient-gradient}. Additional performance details are listed in \Cref{sec:performance_eval}.

Although our network inversion step follows the standard optimal-projection-based ROM literature \citep{carlberg2013gnat}, future work may consider directly learning an inverse during training time. The encoder introduced in the manifold construction process may be a step towards learning an inverse. However, the encoder in its current form does not have optimal accuracy and is bounded to a particular discretization. Future work should consider lifting these limitations.
\section{Network Details}
\label{sec:network_details}
Consistent with the implicit neural representation literature, we parameterize the manifold-parameterization function manifold with an MLP network $\lowDimensionalManifoldNN$. The input dimension of the network is $\dimensionIn + \nred$ while the output dimension of the network is $\dimensionOut$, where $\dimensionIn$ is the dimension of the input spatial vector while $\dimensionOut$ is the output dimension of the vector field of interest. Our MLP network contains 5 hidden layers, each of which has a width of $(\scaleMlp\cdot\dimensionOut)$, where $\scaleMlp$ is the hyperparameter that defines the learning capacity of the network. Essentially, our network has two tunable hyperparameters, $\nred$ and $\scaleMlp$. A detailed study on these hyperparameters will be discussed in \Cref{sec:hyper-study}.

Since we require the network to be continuously differentiable with respect to both the spatial coordinates $\xb$ and the latent space vector $\latentSpaceVec$, we adopt continuously differentiable activation functions. In practice, either ELU \citep{clevert2015fast} or SIREN \citep{sitzmann2020implicit} serves this purpose.

For the encoder network $\encoder$, the input is a 1D vector of length $\nSpatial$ with $\dimensionOut$ channels. The output is a vector of dimension $\nred$. Using MLPs would lead to enormous network size when $\nSpatial$ is large. We therefore design an encoder network first to apply multiple 1D convolution layers of kernel size 6, stride size 4, and output channel size $\dimensionOut$ until the output 1D vector's length is the closest to $32/\dimensionOut$ but no smaller. Afterward, the encoder network reshapes the vector into 1 channel and applies an MLP layer to reduce the dimension of the vector to 32. The last MLP layer then transforms the previous 32-dimensional vector into dimension $\nred$. Discretization-specific encoder networks can also be used, e.g., PointNet for point clouds and convolutional neural network for grid data.

\subsection{Training Details}
We use the Adam optimizer \citep{kingma2014adam} for stochastic gradient descent. We use the Xavier initialization for ELU layers and the default initialization ($\omega_0=30$) for SIREN layers. Unless otherwise noted, we train with a base learning rate of $lr=1e-4$ and adopt a learning rate decay strategy ($10\cdot lr\rightarrow5\cdot lr\rightarrow2\cdot lr\rightarrow1\cdot lr\rightarrow0.5\cdot lr\rightarrow0.2\cdot lr$). For each aforementioned learning rate, we train for $30,000$ epochs. We adopt a batch size $16$ (i.e., $16$ simulation snapshots) for the encoder and therefore a batch size of $16\cdot\nSpatial$ for the implicit-neural-representation-based manifold-parameterization function. We implement the entire training pipeline in PyTorch Lightning\citep{falcon2019pytorch} which facilitates distributed training across multiple GPUs. The training vector fields are standardized to have zero mean and unit variance. The training spatial coordinates are also standardized for the ELU implementation, but they are preprocessed to be between $[-1,1]$ for the SIREN implementation.
\section{Network Gradients}
\label{sec:net_grad}
Gradients of the implicit neural representation ($\gradbspatial\lowDimensionalManifoldNN$ and $\lowDimensionalManifoldNNwrtLatentFlat$) are crucial for our reduced-order pipeline. We use them to compute the spatial and temporal gradients of the vector fields, $\gradb\fullOrderModelArgs{\pos}{\continuousTime}=\gradbspatial\lowDimensionalManifoldNNArgs{\pos}{\latentSpaceVec}$ and $\dot{\fullOrderModel}(\pos,\continuousTime)=\lowDimensionalManifoldNNwrtLatentFlatArgs{\pos}{\latentSpaceVec}\latentSpaceVecDot$. Furthermore, the gradient with respect to the latent space $\lowDimensionalManifoldNNwrtLatentFlat$ is also a key ingredient for the network inversion (\Cref{sec:net-inversion-details}).

\subsection{Network Gradients via Direct Differentiation}
One way to compute the gradients is through directly differentiating the continuously differentiable network.

\subsubsection{Efficient Implementation}
\label{sec:nn-efficient-gradient}
\begin{algorithm}[H]
    \SetAlgoLined
    \KwData{$\xb$}
    \KwResult{$\pdflat{\lowDimensionalManifoldNN}{\xb}(\xb), \lowDimensionalManifoldNN(\xb)$}
    Let $\lowDimensionalManifoldNN=\gb_n \circ \cdots \circ \gb_1$ be a $n$-layer MLP network of interest.\\
    Initialization: $\pdflat{\yb}{\xb} = \identity$, $\yb=\xb$\\
    \For{$i=1,\ldots,n$}{
        $\pdflat{\yb}{\xb} = \pdflat{\gb_i}{\xb}(\yb)\pdflat{\yb}{\xb}$\\
        $\yb=\gb_i(\yb)$
    }
    \KwRet $\pdflat{\yb}{\xb}, \yb$
    \caption{Network gradient}
    \label{alg:grad}
\end{algorithm}

Unfortunately, computing these differentiations via auto-diff (computational graph tracking) is too slow for the high-performance application explored in this work. To ensure maximum efficiency, we implement the gradients analytically through the chain rules of each network layer, similar to the grad net approach by \citep{lindell2021autoint}. Since we only use fully-connected layers, the implementation is straightforward. In addition, along the way of computing the gradient via chain rules, we also obtain the function value of the neural network itself. Therefore, we obtain both the gradient and the function value in a single forward pass. \Cref{alg:grad} describes the details.

\subsubsection{Failure Modes}
While direct differentiation is generally accurate, we observe failure modes when the training samples are sparse. Consider the case where $\lowDimensionalManifoldNNArgs{\pos}{\latentSpaceVec}=\xb$, e.g., the undeformed, reference configuration of elastodynamics. In this case, the ground truth spatial gradient is $\gradbspatial\lowDimensionalManifoldNN=\identity$.

\begin{figure}
    \centering
    \includegraphics[width=\textwidth]{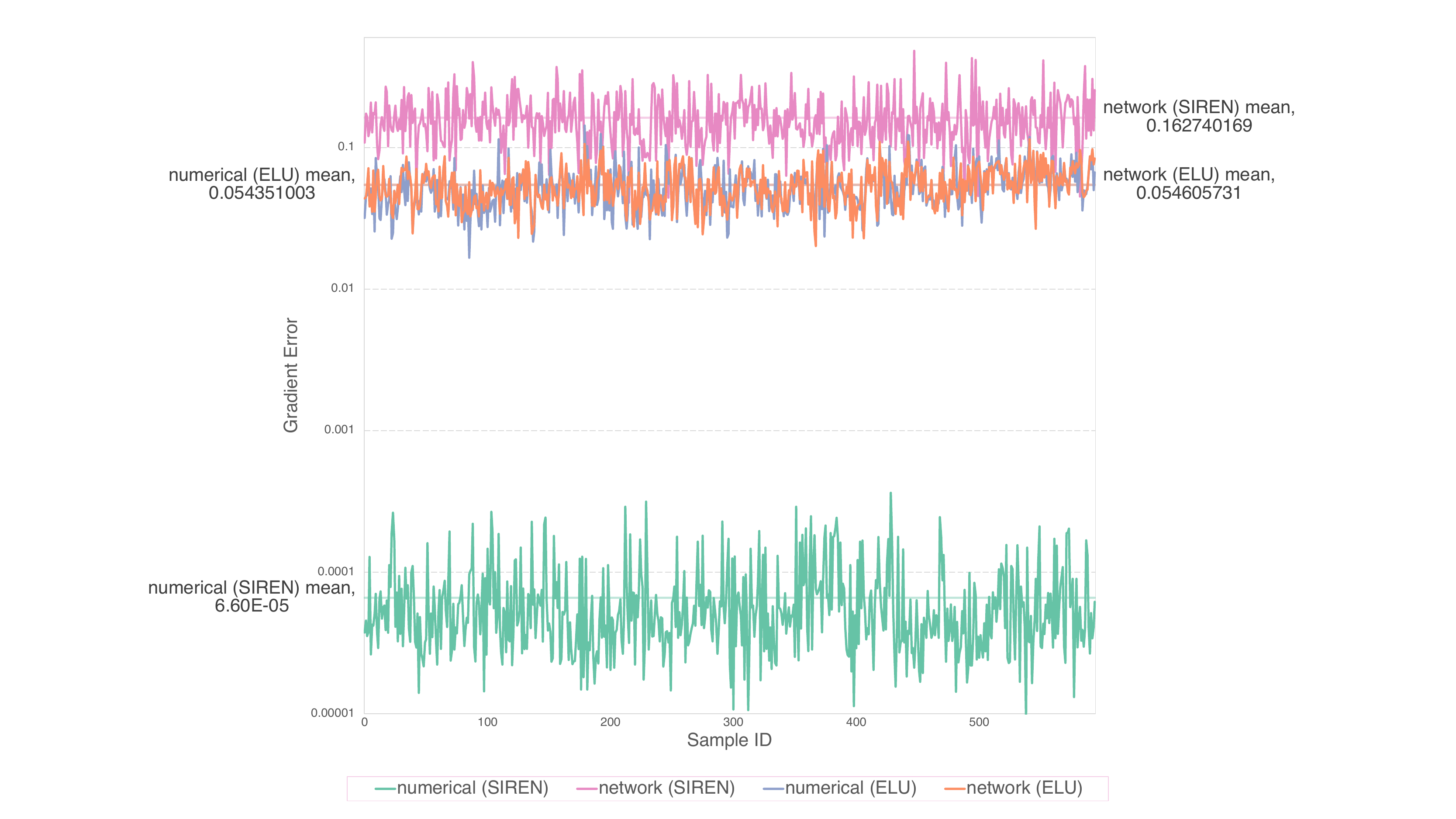}
    \caption{Gradient errors evaluated at the centers of tetrahedra. Numerical differentiation (specifically, finite element with linear basis) significantly improves the gradient accuracy of SIREN.}
    \label{img:failure-gradient}
\end{figure}

We train the neural network with spatial samples discussed in \Cref{sec:gravity-impact}. After training, we evaluate the spatial gradients at the center of each tetrahedron and compute its error (from the identity matrix) in the L2 norm. As shown in \Cref{img:failure-gradient}, errors are observed in both ELU and SIREN networks while SIREN's error is significantly larger, c.f.,\citep{yang2021geometry}. This discrepancy can be understood as the high-frequency prior by SIREN being more suitable for approximating high-frequency functions. By contrast, the low-frequency prior by ELU is better at approximating low-frequency functions \citep{hertz2021sape}.  

We find training with more samples and/or gradient supervision \citep{chen2023model} reduces the errors above at the cost of additional training resources but does not resolve the issue completely. Future work may consider more advanced gradient regularization methods \citep{liu2022learning}.

\subsection{Network Gradients via Numerical Differentiation}
\label{sec:nn-numerical-gradient}
Another approach to computing neural network gradients is numerical approximation, such as the finite difference method and the finite element method. \Cref{img:failure-gradient} demonstrates the drastically improved gradient accuracy with the finite element method (linear basis function) on networks trained with SIREN. We find such a hybrid approach to be an ideal middle ground, i.e., we represent the function itself with the discretization-independent neural network and compute the gradient via discretization-dependent numerical differentiation.

With numerical gradients, users still benefit from our continuous representation that allows them to use \emph{any} discretization for gradient computation (e.g., ground impact elasticity deformation with remeshing). Prior ROM approaches require these numerical gradients to be computed by a fixed mesh.
\section{Hyperparameters}
\label{sec:hyper-study}
During the offline training stage, we can modify two network architecture hyperparameters: the size of the latent space ($\nred$) and the width of the MLP ($\scaleMlp$). After training and during the online deployment stage of the model, the key hyperparameter is the number of integration samples ($\sampleSetCardinality$). In this section, we run a sensitivity analysis on these hyperparameters.

\begin{figure}
    \centering
    \includegraphics[width=\textwidth]{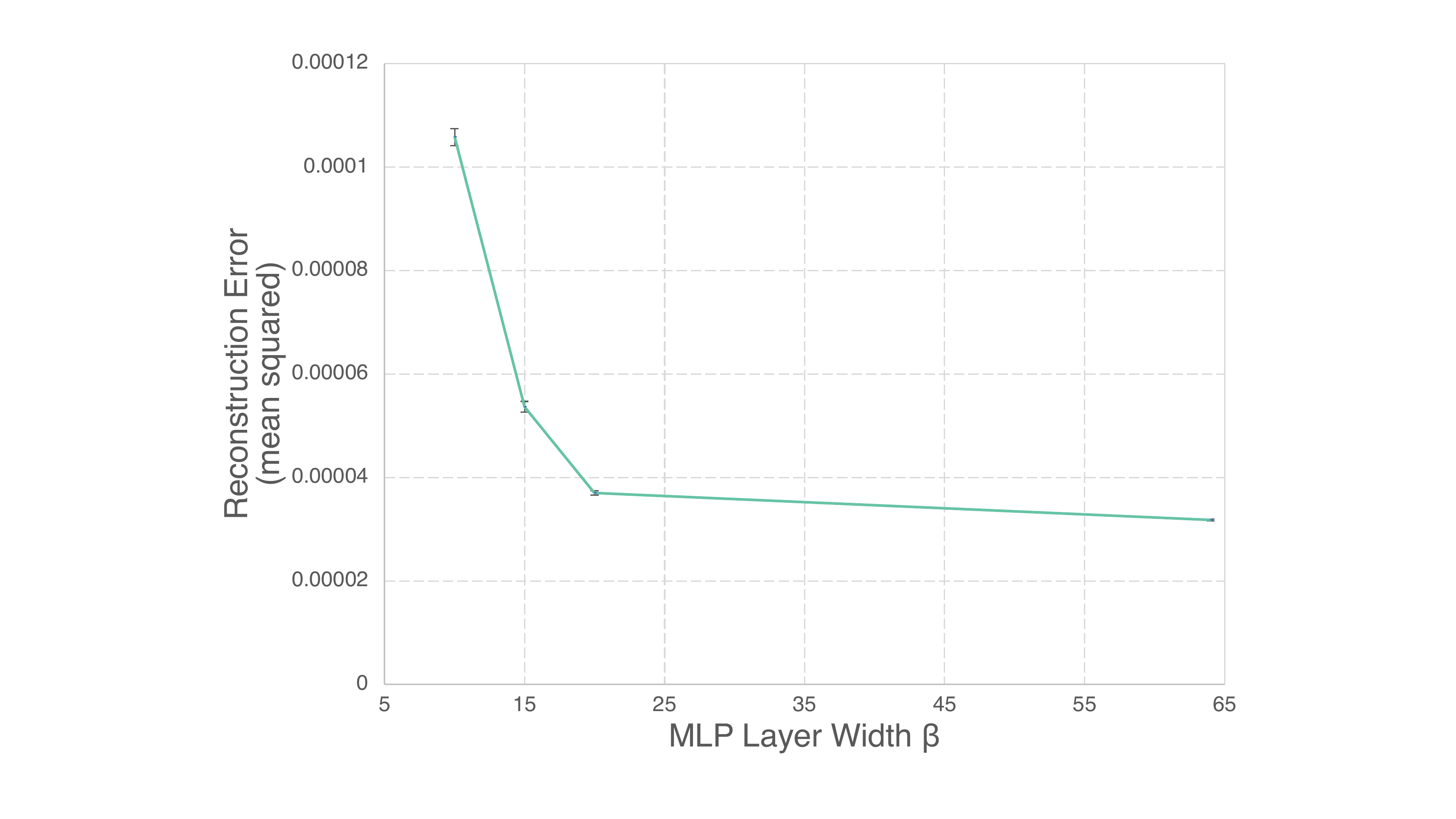}
    \caption{Reconstruction accuracy vs. the MLP layer width ($\scaleMlp$). For each setup, we repeat the training for $8$ times. The mean values and the error bars are shown. Increasing MLP layer width leads to higher accuracy.}
    \label{img:scalemlp-accuracy}
\end{figure}
\Cref{img:scalemlp-accuracy} shows that increasing the width of each MLP layer improves the reconstruction accuracy. Unless otherwise noticed, all reconstruction errors reported are mean squared errors (MSE). While \Cref{img:scalemlp-accuracy} demonstrates an empirical convergence, future work should consider a theoretical analysis \citep{kunisch2002galerkin}.

\begin{figure}
    \centering
    \includegraphics[width=\textwidth]{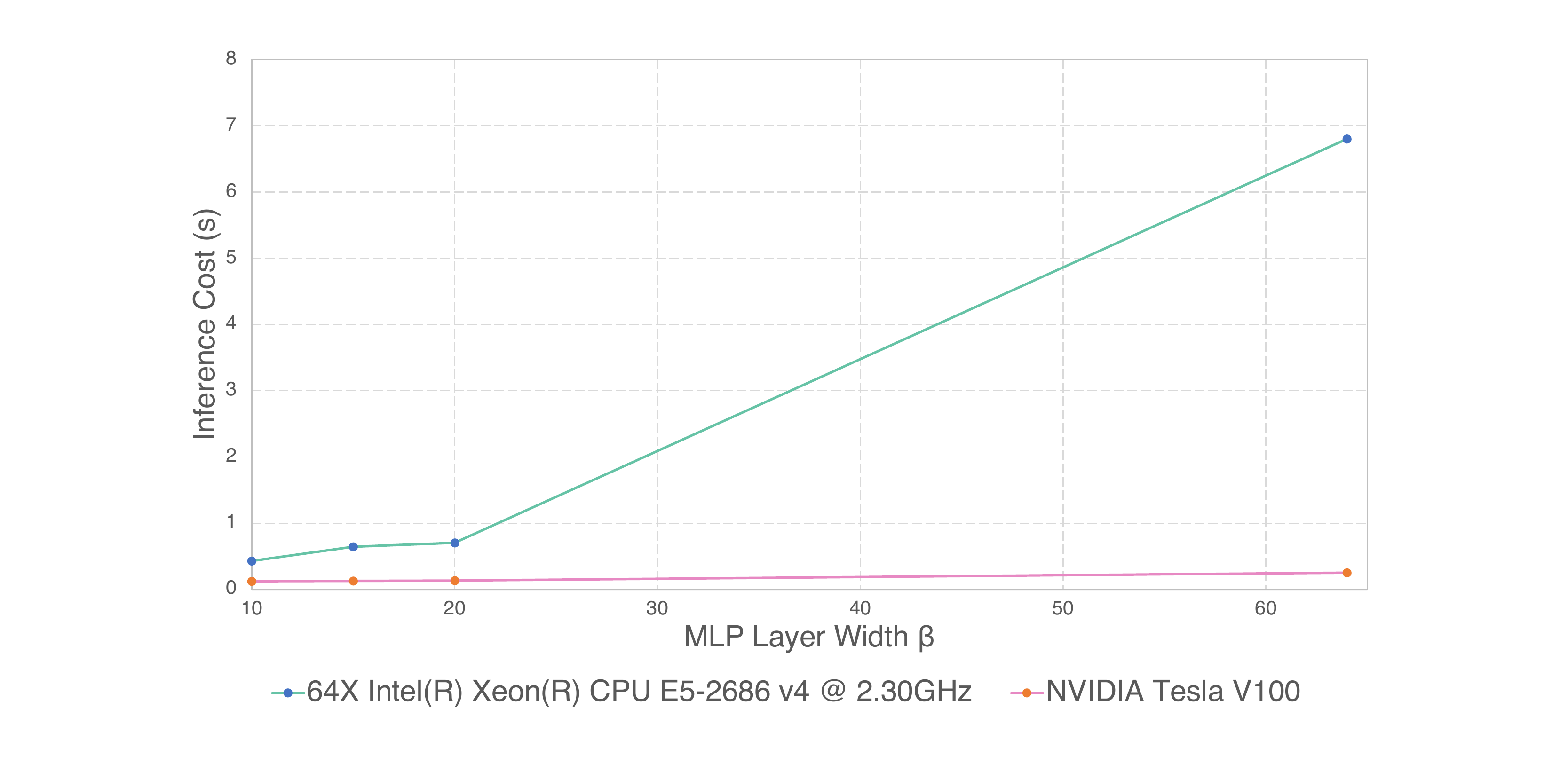}
    \caption{Network inference cost (CPU and GPU) vs. the MLP layer width ($\scaleMlp$). Inference cost (over 100 runs) increases as the network size increases. CPU suffers from a more severe performance drop than GPU.}
    \label{img:scalemlp-speed}
\end{figure}
However, large layer widths also entail higher computational cost (see \Cref{img:scalemlp-speed}). Compared to the highly parallelized GPUs, CPUs suffer from a more severe performance drop due to the larger network size.

Unlike traditional PCA-based model reduction approaches (e.g., POD), increasing the dimension of the latent space vector does not yield higher accuracy for our approach (see \Cref{img:mem_and_accuracy}). In prior ROM approaches, augmenting the dimension of the latent space vector directly increases the number of entries in the manifold-parameterization function matrix, which, in turn, improves the approximation capacity of the manifold-parameterization function. However, as shown in \Cref{img:scalemlp-accuracy}, in the implicit neural representation framework, we can control the learning capacity of the network architecture by explicitly modifying the width of the MLP layer. Therefore, the importance of the latent space dimension diminishes. This experimental observation is consistent with prior work’s theoretical analysis (Remark 2.1 by \citet{lee2020model}) which says the lower bound of the latent space dimension is the dimension of the problem parameters (e.g., material properties) plus one. Consequently, one should consider increasing the learning capacity of the manifold-parameterization function instead of increasing the latent space dimension.

\begin{figure}
    \centering
    \includegraphics[width=\textwidth]{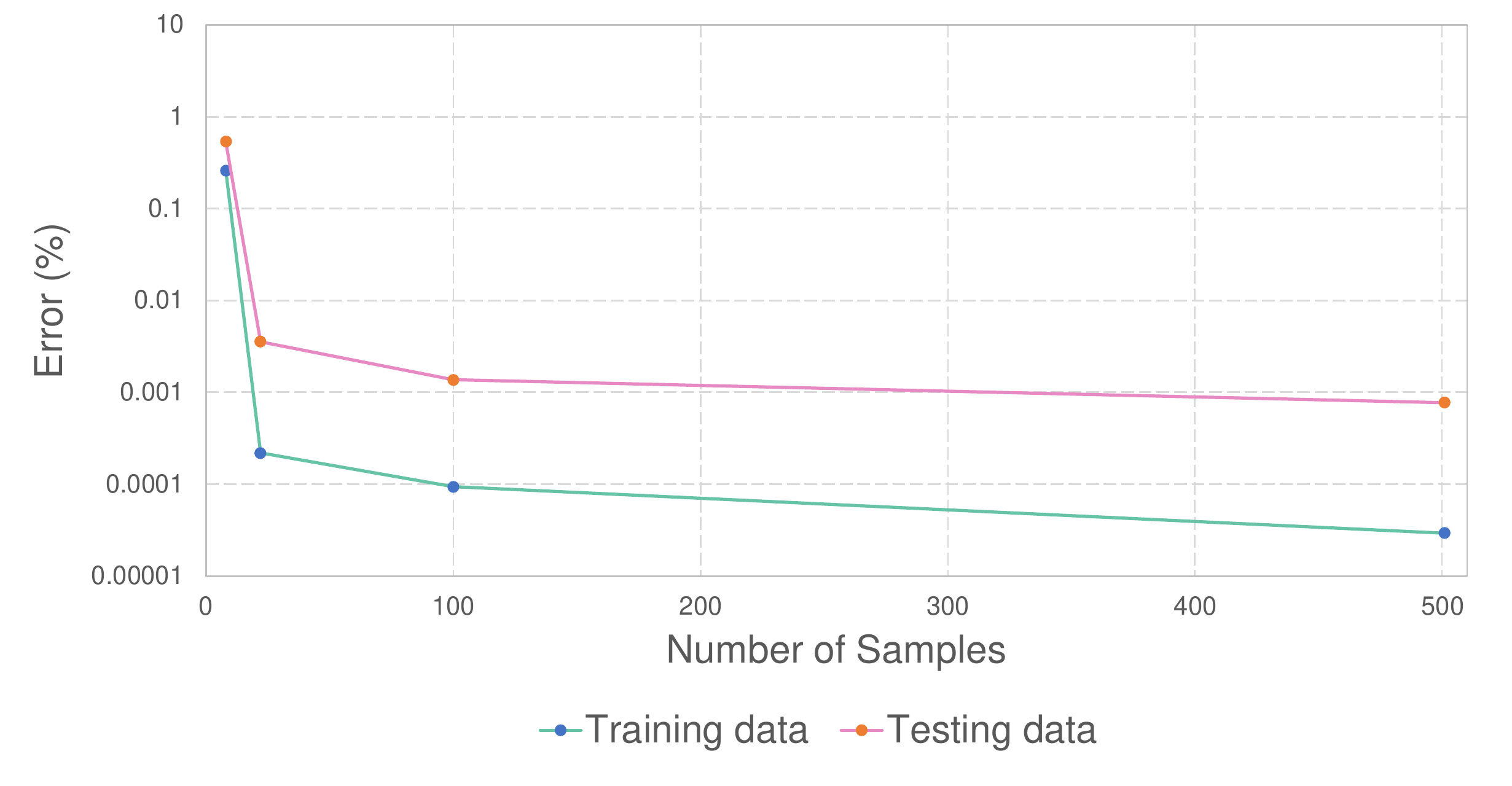}
    \caption{PDE solution accuracy vs. the number of samples. Increasing integration sample counts yields higher accuracies on both training and testing data.}
    \label{img:optimal-sampling-plot}
\end{figure}
During the online stage of the reduced-order model, we select integration samples via the greedy sampling scheme (\Cref{sec:optimal-sampling}). Increasing the number of integration samples leads to higher accuracy (\Cref{img:optimal-sampling-plot}).

\section{Thermodynamics}
\label{sec:thermo_details}
\subsection{Continuous PDE}
\begin{align}
  \quad\pd{u}{t}-\nu(x)\pdtwo{u}{x} = 0  
\end{align}
In thermodynamics, we study the 1D heat equation of the spatiotemporal dependent temperature $u$. We assume a zero-Dirichlet boundary condition, though other boundary conditions can also be incorporated.

\subsection{Full-order Model}
The full-order model discretizes the spatial vector field with a regular Eulerian grid using 501 equally spaced samples ($\nSpatial = 501$).

We then approximate the spatial gradient using the finite difference method, 
\begin{align}
  \pdtwo{u}{x}(x^i, \discreteTime) = \frac{u(x^i-\gridSpacing, \discreteTime) + u(x^i+\gridSpacing, \discreteTime) - 2u(x^i, \discreteTime)}{\gridSpacing^2},  
\end{align}
where $\gridSpacing$ is the grid spacing.

\rev{We use this to compute the next-time step velocity}, 
\begin{align}
\dot{u}^{i}_{n+1}=\nu(x^i)\pdtwo{u}{x}(x^i, \discreteTime).
\end{align}

We assume a first-order explicit time-stepping scheme,
\begin{align}
u^{i}_{n+1} =  u^{i}_{n} + \dt\dot{u}^{i}_{n+1}.
\end{align}

\subsection{Reduced-order Model}
\label{sec:thermmo:rom}
Instead of a finite-difference treatment of the spatial gradient, we compute the gradient via direct differentiation of the network. The latent space dimension is $\nred=16$ and the width of the MLP is $\scaleMlp = 128$.

\subsection{Training and Testing Data}
\label{sec:thermodynamics-training-data}
\begin{figure}
    \centering
    \includegraphics[width=\textwidth]{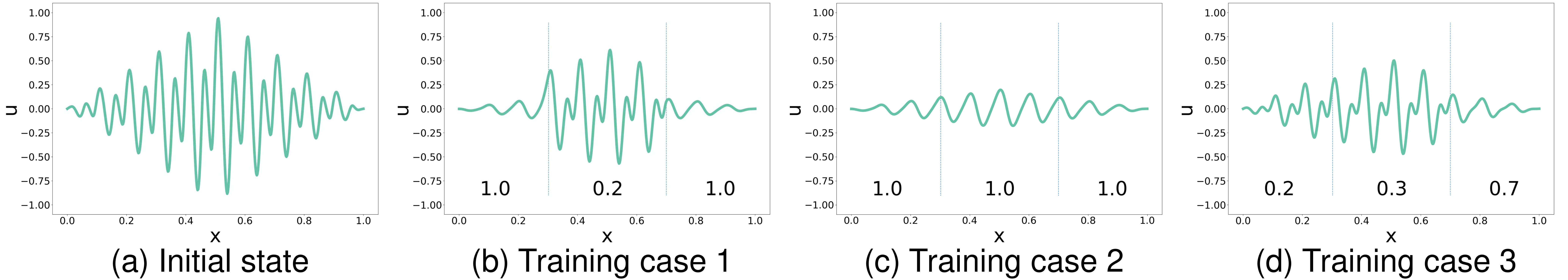}
    \caption{Thermodynamics: training data.}
    \label{img:thermo:training_data}
\end{figure}

We generate training data by setting $\nu$ to different piecewise constant functions of three regions, i.e., the parameter vector $\params\in \paramDomain=[0.2, 1.0]^{3} \subset \RR{3}$. \Cref{img:thermo:training_data} displays the initial and the final states of training data. Notice how different regions diffuse with different speeds due to the spatially varying $\nu$. In total, $8$ PDE temporal sequences (of 100 time steps) with different $\nu$'s are generated for training. We sample another $4$ $\nu$'s for testing purposes. All training and testing data adopt the same initial condition (\Cref{img:thermo:training_data}a).

\subsection{Results}
\begin{figure}
  \centering
  \includegraphics[width=\textwidth]{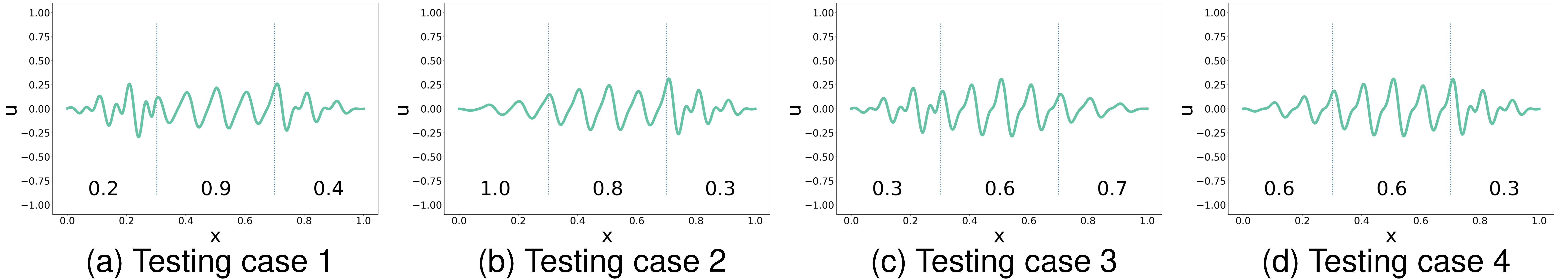}
  \caption{Thermodynamics: reduced-order simulation on the testing dataset.}
  \label{img:thermo:testing_result}
\end{figure}
\Cref{img:thermo:testing_result} demonstrates the performance of the reduced-order model on the testing dataset. With the robust sampling scheme, we can achieve $0.43\%$ error using just $22$ spatial samples (see \Cref{img:heat_all}).
\begin{figure}
  \centering
  \includegraphics[width=0.5\textwidth]{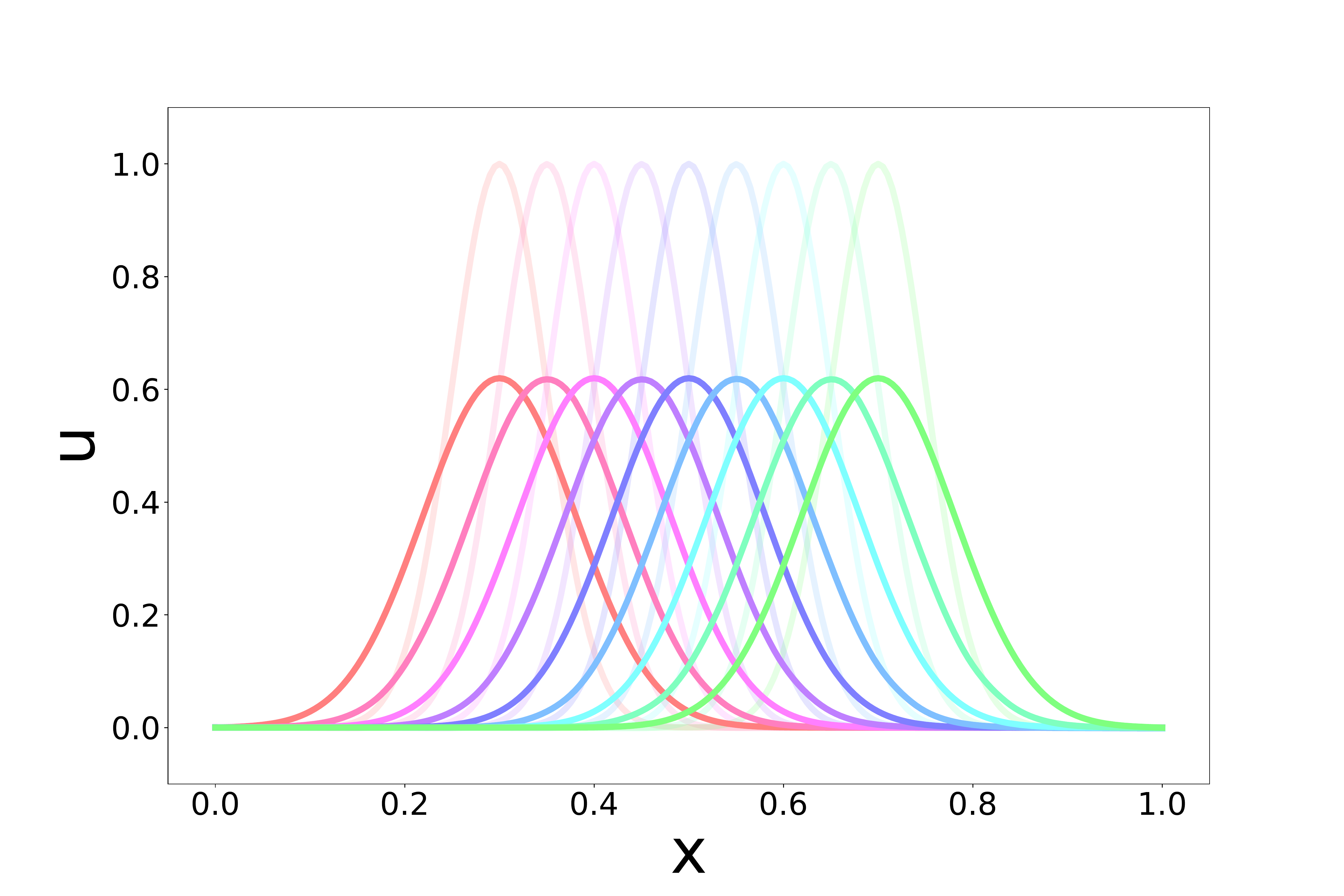}
  \caption{Our reduced-order diffusion solver with various initial conditions (transparent) and the final diffused states. Different colors correspond to different initial conditions.}
  \label{img:interpolate-initial}
\end{figure}
In addition, by incorporating different initial conditions (e.g., Gaussian profiles centered at different horizontal locations) in the training data, our approach can capture diffusion with a wide range of initial conditions (\Cref{img:interpolate-initial}).

To gauge our approach's ability to conserve physical laws, we measure the energy evolution over time in the region between $x_1=0.396$ and $x_2=0.526$ (\Cref{img:thermo:testing_result}). As shown by \citet{cannon1984one}, the escaped energy (EE) is measured by the flux difference ($\pdflat{u}{x}(x_1,t)-\pdflat{u}{x}(x_2,t)$) at the left and right boundaries integrated over time; the stored energy (SE) is measured by the heat energy density, proportional to the temperature ($u$), integrated over space; the total energy (TE) is the sum of the prior two energies, i.e., TE=EE+SE. \Cref{img:heat_all} demonstrates that the reduced simulation preserves the total energy over time.
\section{Image Processing}
\label{sec:img-processing}

\subsection{Continuous PDE}
\begin{align}
  \quad\pdflat{u}{t}-\nu(\xb)\gradb^2u = 0
\end{align}

In image processing, we study the Perona–Malik diffusion equation of the gray-scale image $u$ \citep{perona1990scale}.

\subsection{Full-order Model}
The full-order model discretizes the spatial 2D pixel field with a regular Eulerian grid using $256 \times 256$ equally spaced samples ($\nSpatial = \imagePixelNumber$).

We then approximate the spatial gradient using the finite difference method, 
\begin{align}
  \pdtwo{u}{x}(x^i, y^i, \discreteTime) = \frac{u(x^i-\gridSpacing, y^i, \discreteTime) + u(x^i+\gridSpacing, y^i, \discreteTime) - 2u(x^i, y^i, \discreteTime)}{\gridSpacing^2},
\end{align}

\begin{align}
  \pdtwo{u}{y}(x^i, y^i, \discreteTime) = \frac{u(x^i, y^i-\gridSpacingY, \discreteTime) + u(x^i, y^i+\gridSpacingY, \discreteTime) - 2u(x^i, y^i, \discreteTime)}{\gridSpacingY^2}.
\end{align}

\rev{We use these to compute the next-time step velocity}, 
\begin{align}
\dot{u}^{i}_{n+1}=\nu(x^i, y^i)( \pdtwo{u}{x}(x^i, y^i, \discreteTime) + \pdtwo{u}{y}(x^i, y^i, \discreteTime)).
\end{align}

We assume a first-order explicit time-stepping scheme,
\begin{align}
u^{i}_{n+1} =  u^{i}_{n} + \dt\dot{u}^{i}_{n+1},
\end{align}
where $\gridSpacing = \gridSpacingY$ is the grid spacing.

\subsection{Reduced-order Model}
The reduced-order model follows the same treatments as \Cref{sec:thermmo:rom}.

\subsection{Training and Testing Data}
\begin{figure}
  \centering
  \includegraphics[width=\textwidth]{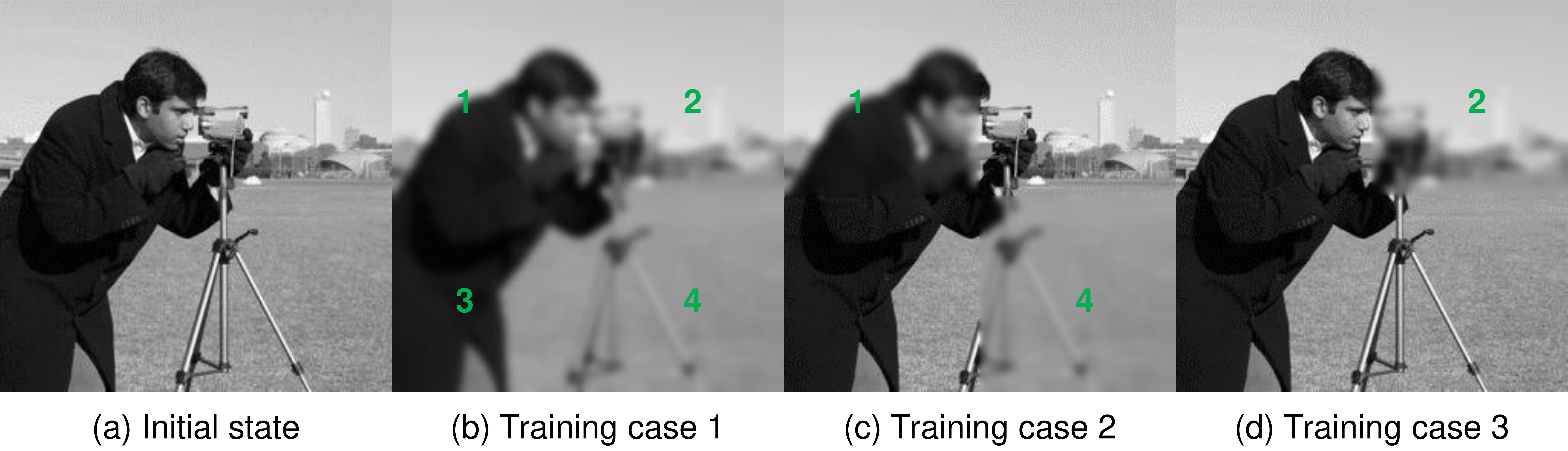}
  \caption{Image Processing:training dataset. The blurred region is numbered.}
  \label{img:imgprocess:training_result}
\end{figure}

Similar to \Cref{sec:thermodynamics-training-data}, we generate training data by setting $\nu$ to different piecewise constant functions of four equally-spaced regions, i.e., the parameter vector $\params \in \paramDomain=\{0, 0.2\}^{4}\subset\RR{4}$. $\paramDomainTrain$ contains $11$ parameter vectors (\Cref{img:imgprocess:training_result}), where $0$, $2$ or $3$ regions have zero diffusion coefficients, i.e., no blurring. $\paramDomainTest$ contains $4$ parameter vectors, where only $1$ region is unblurred. For each training and testing parameter vector, we run a simulation of $20$ time steps. All training and testing data adopt the same initial condition (\Cref{img:imgprocess:training_result}a).

\begin{figure}
  \centering
  \includegraphics[width=\textwidth]{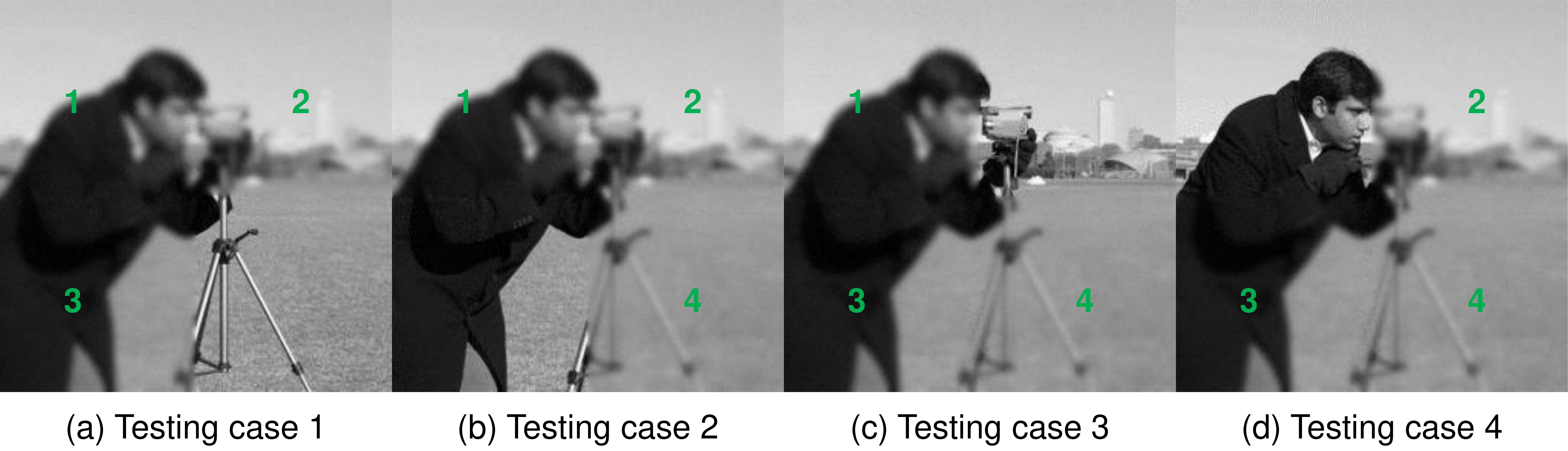}
  \caption{Image Processing: our reduced-order simulation successfully blurs the target regions in the testing dataset. The blurred region is numbered.}
  \label{img:imgprocess:testing_result}
\end{figure}

\subsection{Results}
\begin{figure}
  \centering
  \includegraphics[width=\textwidth]{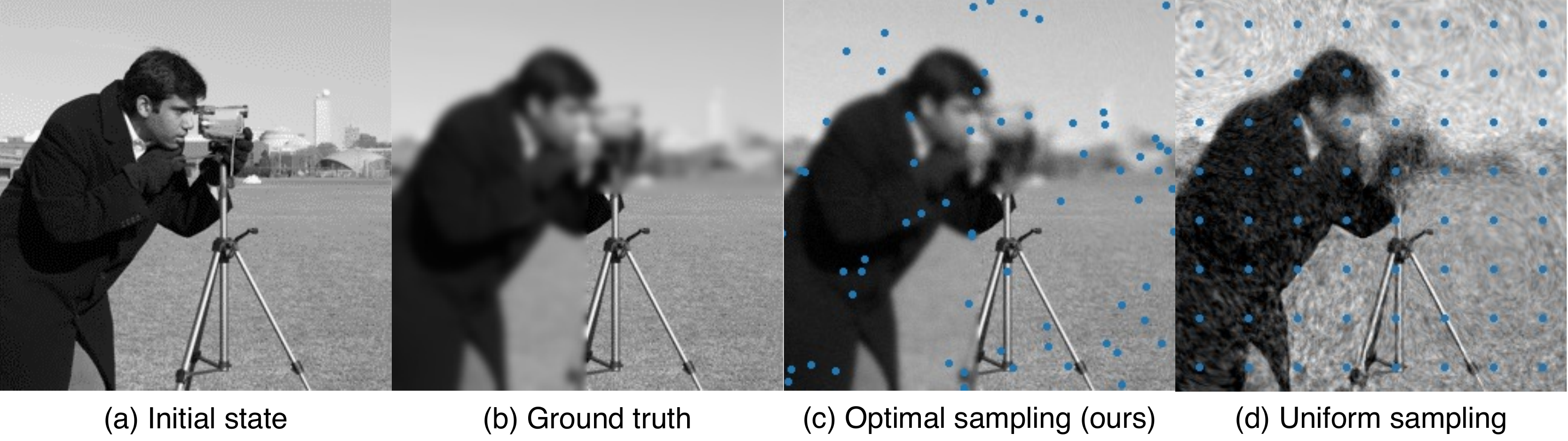}
  \caption{Blurring of portions of the image (20 time steps). (b) Ground truth solution uses all $\nSpatial=\imagePixelNumber$ pixels. (c) Our approach uses very few integration samples ($\sampleSetCardinality=63$, blue circles) and obtains a similar result as the ground truth (PSNR $36.1$). (d) Naive uniform sampling ($\sampleSetCardinality=64$, blue circles) leads to a poor agreement with the ground truth (PSNR $23.4$).}
  \label{img:image-smoothing}
\end{figure}

\Cref{img:imgprocess:testing_result} demonstrates the reduced simulation's performance on the testing dataset. Furthermore, with the robust sampling scheme, we can achieve $36.1$ PSNR using just $63$ spatial samples (see \Cref{img:image-smoothing}). The dimension reduction is $99.90\%$. 

\section{Advection Equation}
\label{sec:advection_details}
Nonlinear ROM techniques have been shown to significantly outperform linear ROM techniques on problem with slowly decaying Kolmogorov n-widths \citep{peherstorfer2022breaking,lee2020model}. In this section, we test CROM, also a nonlinear ROM method, on a transport-dominated problem, given by the advection equation.

\subsection{Continuous PDE}
\begin{align}
    \pd{\advectQuantity}{t} + (\advectSpeed\cdot\grad)\advectQuantity=0,
\end{align}
where $\advectQuantity$ is the advected quantity and $\advectSpeed$ is the advection velocity.

\subsection{Full-order Model}
The full-order model discretizes the 1D spatial field with a regular grid using $100$ equally spaced samples. The PDE is temporally discretized via forward Euler.

\subsection{Reduced-order Model}
We vary the latent space dimension $\nred$ and use a fixed width of the MLP $\scaleMlp = 20$.

\subsection{Training and Testing Data}
In this example, we consider a reproductive case where the training and the testing data are the same. In particular, the initial condition is given by a Gaussian profile and is then advected under a constant velocity.

\subsection{Results}
\begin{figure}
    \centering
    \includegraphics[width=0.8\textwidth]{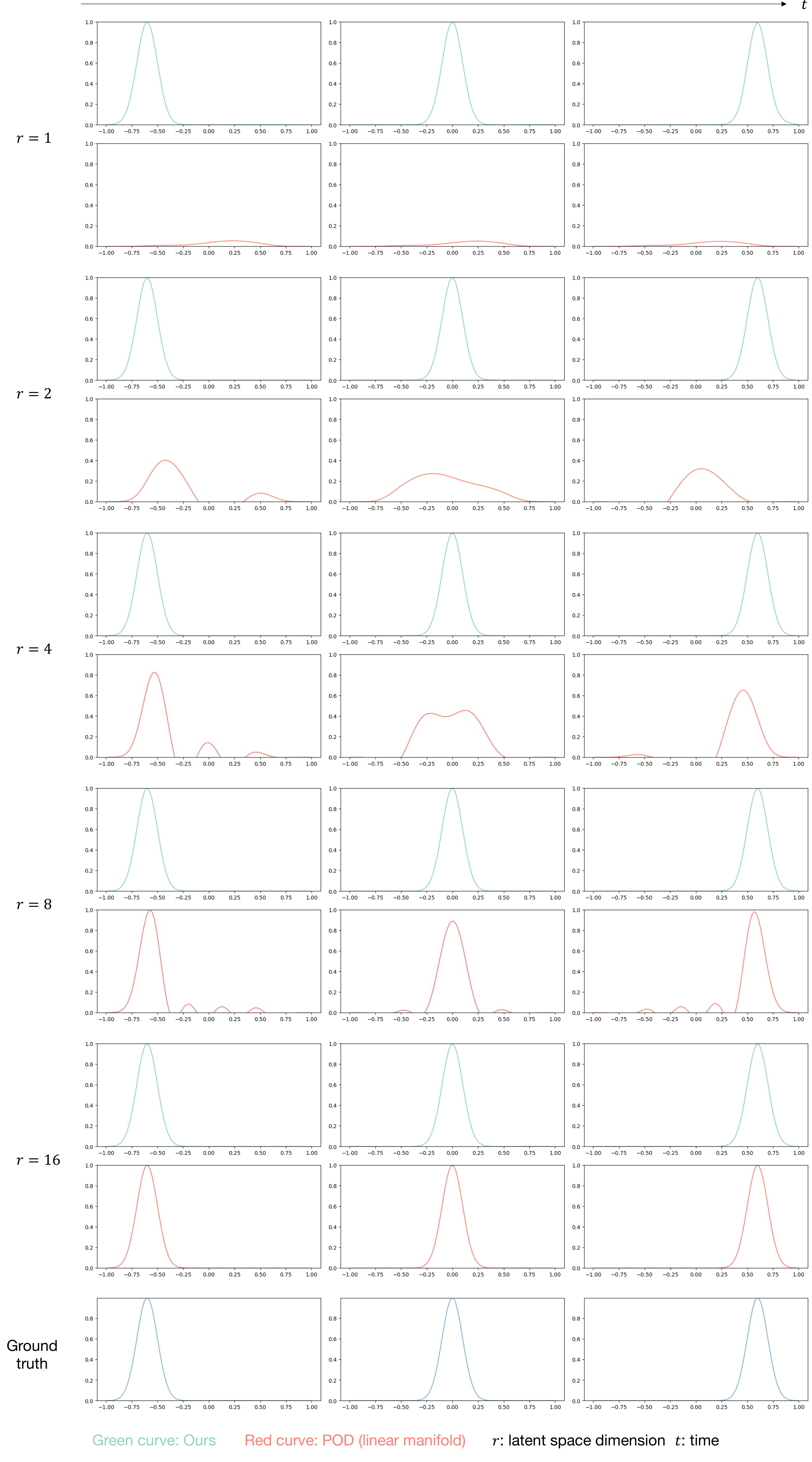}
    \caption{Advection. Our CROM employs a nonlinear manifold-parameterization function and accurately captures the advection phenomenon with just a one-dimensional latent space. By contrast, linear manifolds, such as POD, require a significantly larger latent space dimension. Lower dimensional latent spaces on linear manifolds yield significant artifacts due to the inability to model slowly decaying Kolmogorov n-widths problems.}
    \label{img:advection}
\end{figure}

Since this example is reproductive and the only parameter of the system is time $t$, the intrinsic solution-manifold dimension \citep{lee2020model} is $1$. \Cref{img:advection} demonstrates that CROM indeed accurately captures the transport behavior of the Gaussian profile even at the the intrinsic solution-manifold dimension ($\nred=1$). By contrast, POD suffers from serious artifacts with lower dimensional latent space and requires a significantly higher dimensional latent space ($\nred=16$) in order to model the advection behavior. Furthermore, \Cref{img:advection_burgers} shows that our approach is also more accurate than the convolutional autoencoder approach by \citet{lee2020model} when $\nred=1$. 

\section{Burgers' Equation}
\label{sec:burgers_details}
In this section, we continue to investigate problems with slowly decaying Kolmogorov n-widths. In particular, we obtain the setup by \citet{lee2020model} (see Section 7.1 of their paper) and compare with their (discretization-dependent) convolutional autoencoder manifold. 

\subsection{Continuous PDE}
\begin{align}
    \pd{\burgerQuantity}{t} + \pd{0.5\burgerQuantity^2}{x}=0.02e^{\burgerD x},
\end{align}
where $\burgerQuantity$ is the variable of interest. The initial condition is as follows.
\begin{align}
    \burgerQuantity(0,0)=4.25\\
    \burgerQuantity(x,0)\quad\forall x\in(0,100]
\end{align}
We further assume a zero Neumann boundary condition.

\subsection{Full-order Model}
The full-order model discretizes the 1D spatial field with a regular grid using $256$ equally spaced samples. The PDE is temporally discretized via forward Euler.

\subsection{Reduced-order Model}
We vary the latent space dimension $\nred$ and use a fixed width of the MLP $\scaleMlp = 64$.

\subsection{Training and Testing Data}
We generate training and testing data by varying the parameter $\burgerD$. We set the training parameter to $\burgerDtrain =\{0.015 +(0.015/7)j\}_{j=0,1,...7}$, resulting in $8$ training cases. For online testing, we set the parameter to $\burgerDtest = 0.021$.

\subsection{Results}
Since we can uniquely parameterize the family of training and testing data with two variables $t$ and $\burgerD$, the intrinsic solution-manifold dimension \citep{lee2020model} is $\inDimension=2$, which is the lower bound for the latent space dimension $\nred$.

\Cref{img:advection_burgers}b reports the result for $\nred=\inDimension=2$. While both POD and convolutional autoencoder (CAE) approach from \citet{lee2020model} display clear artifacts, CROM agrees well with the ground truth. Such an agreement at the latent space dimension lower bound (i.e., $\nred=\inDimension$) shows that CROM is very close to achieving the optimal performance of \emph{any} nonlinear trial manifold.

\begin{figure}
    \centering
    \includegraphics[width=0.8\textwidth]{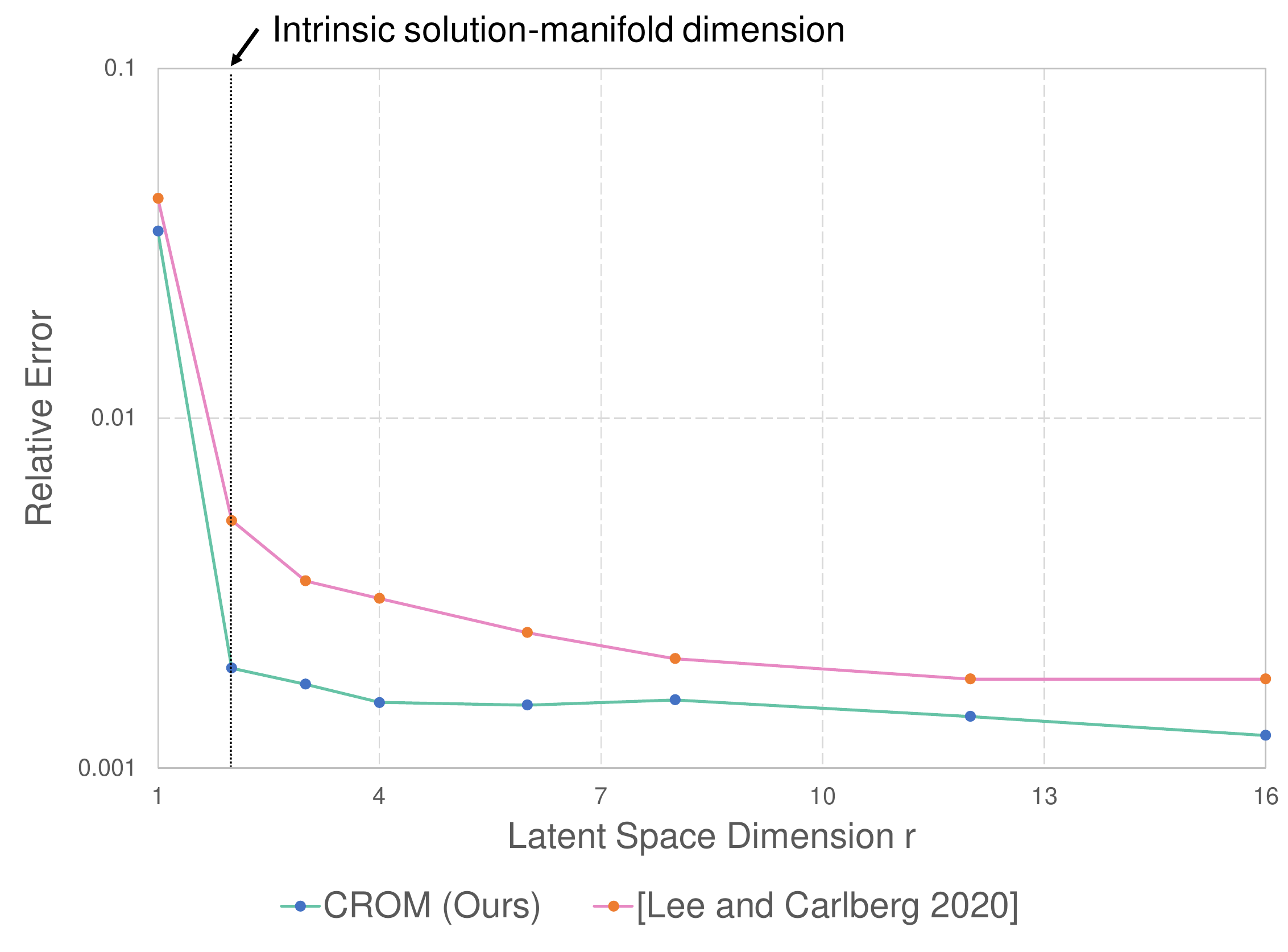}
    \caption{Reconstruction errors. Our approach yields higher accuracies than the convolutional autoencoder approach by \citet{lee2020model} across different latent space dimensions. Our method's accuracy advantage is particularly noticeable at the intrinsic solution-manifold dimension (see the vertical dashed line).}
    \label{img:relative_error}
\end{figure}

Furthermore, \Cref{img:relative_error} demonstrates the accuracy advantage of our approach over the CAE approach across different latent space dimensions.

\section{Fluid Dynamics}
\label{sec:fluid_details}
\subsection{Continuous PDE}
\begin{align}
    \quad
\pdflat{\fluidVelo}{t} + (\fluidVelo\cdot\gradb)\fluidVelo = -\gradb \pressure + \nu\gradb^2\temperatureVec + \externalForce, \, \gradb \cdot \fluidVelo = 0
\end{align}
We study the incompressible Navier-Stokes equations, where $\fluidVelo$ is the velocity field, $\pressure$ is the pressure field, and $\nu$ is the viscosity. We assume there is no external force, i.e., $\externalForce=\zerob$. We assume a no-penetration boundary condition for the velocity and a zero-Dirichlet boundary condition for the pressure.

\subsection{Full-order Model}
The full-order model assumes an operator-splitting scheme following the classic Chorin's projection method \citep{chorin1968numerical}. In particular, we follow the implementation by \citet{stam1999stable}. We summarize the major ingredients and refer to his paper for details.

We sequentially apply 3 linear operators to the velocity field: diffusion, advection, and projection. From the previous time step velocity field $\fluidVelo_{n}$, we apply the diffusion operator according to the viscosity and obtain the diffused velocity $\fluidVelo_{n}^{\text{visc}}$. Next, we employ semi-Lagrangian to obtain the advected velocity field $\fluidVelo_{n}^{\text{adv}}$. For the last projection step, we first compute the pressure $\pressure_{n+1}$ by solving the poisson equation, $\grad^2\pressure_{n+1} = \gradb \cdot \fluidVelo_{n}^{\text{adv}}$. Afterwards, we apply the pressure gradient to obtain the divergence-free velocity $\fluidVelo_{n+1}$. All operations are done on a 2D $16$ by $16$ Eulerian grid.

\subsection{Reduced-order Model}
We compute latent space dynamics ($\discreteLatentSpaceVecPlus$) of the velocity field by projecting $\fluidVelo_{n+1}$ onto the manifold-parameterization function $\lowDimensionalManifoldNN$. For network hyperparameters, we adopt a latent space dimension $\nred=6$ and a MLP width $\scaleMlp = 10$.

\subsection{Training and Testing Data}
We generate training data by varying the spatially-constant viscosity: $\params=\nu \in \paramDomain = [0, \viscMax] \subset \RR{}$. We sample 3 different viscosities from $\paramDomain$. For each viscosity $\nu$, we run a simulation of $50$ time steps. In total, $150$ time steps are used for training. For testing, we first sample 3 viscosity values from $\paramDomain=[0, \viscMax]$. Furthermore, we also sample 3 values from the extrapolated region $\paramDomain^{\text{extrapolate}}=[\viscMax, \viscMaxExtrapolate]$. Notice how the extrapolated region has a larger range ($0.08$) than the interpolated region ($0.02$). All training and testing data adopt the same initial condition (see \Cref{img:fluid:testing_result}).

\subsection{Results}
\begin{figure}
    \centering
    \includegraphics[width=\textwidth]{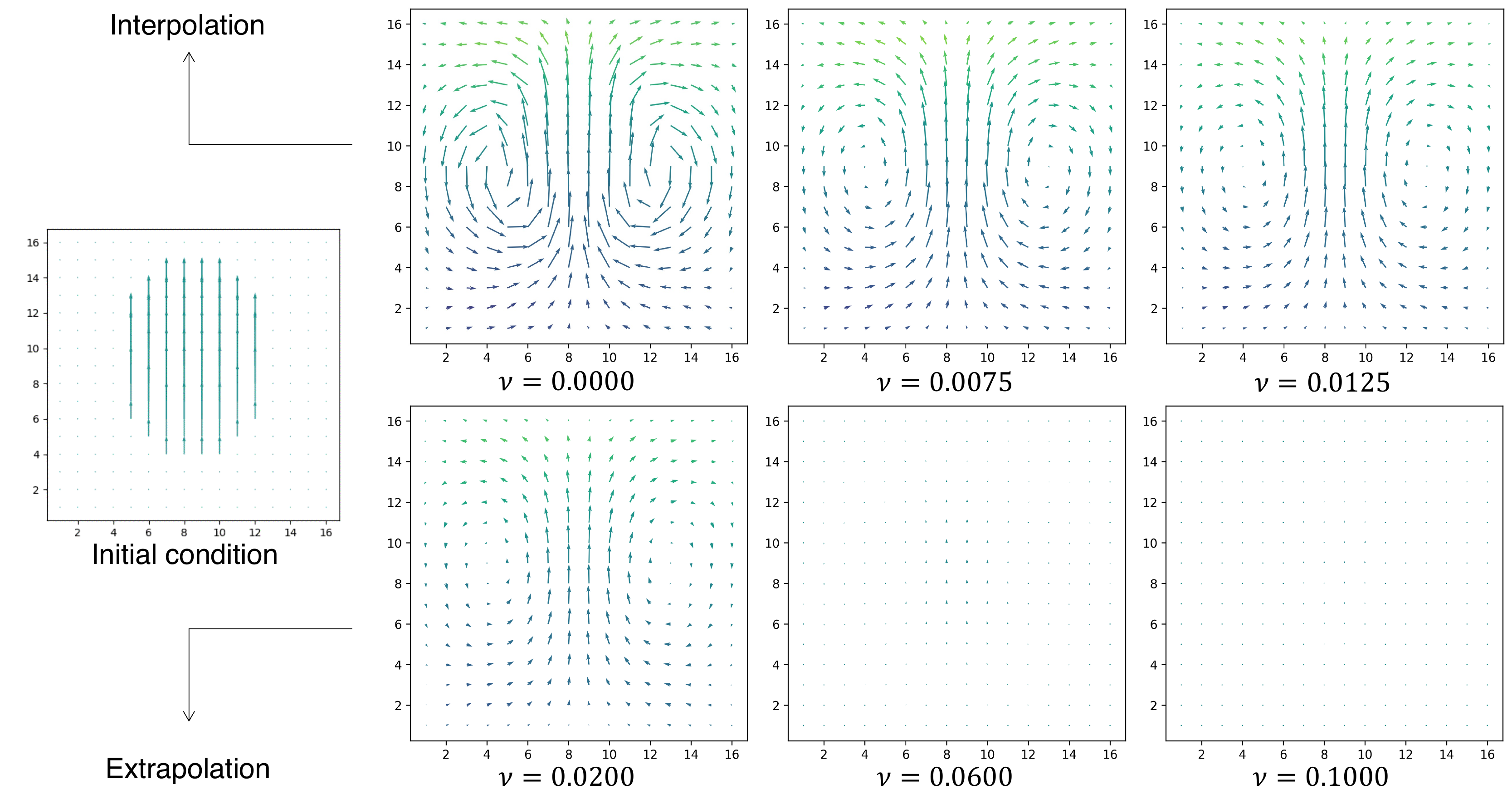}
    \caption{Fluid dynamics: our method captures the velocity fields of a wide range of viscous fluids. Large viscosity severely dissipates the velocity. All frames shown are captured from time step $25$. The first row consists of the interpolated result while the second row consists of the extrapolated result well outside the training range.}
    \label{img:fluid:testing_result}
  \end{figure}
  \Cref{img:fluid:testing_result} demonstrates our method's ability to qualitatively capture fluid dynamics with different viscosities on both interpolated data as well as extrapolated data. Surrogate models that predict velocity fields directly from viscosities (e.g., using a neural network) typically cannot handle extrapolation. By contrast, our approach handles extrapolated scenarios because we only approximate the spatial representation (kinematics) with a neural network, not the PDE itself. Therefore, as long as the extrapolated cases involve vector fields that can be represented by the low-dimensional implicit neural representation, our reduced-order solver is able to evolve the latent space vector to capture them according to the PDE. However, if the extrapolated quantity lies outside of the space spanned by the manifold-parameterization function, our approach would not be able to capture the dynamics. CROM, like most ROM approaches, does not handle arbitrary extrapolation scenarios. See more discussion on generalizability at the end of the main text.

  \rev{To further test the proposed method, we set up a Karman vortex street experiment. The Karman vortex street is particularly challenging as the flow bifurcates from a steady state to the periodic vortex shedding regime. Mathematically speaking, the fixed point solution undergoes a Hopf bifurcation and becomes a limit cycle \citep{huerre1990local}. To capture such an effect, the latent space dynamics needs to be robust under extreme nonlinearity. Many neural-network-based methods, such as physics informed neural network \citep{raissi2019physics}, are well-known to fail in this case \citep{chuang2022experience}.}
  
  \rev{We follow the geometry and material parameters outlined by \citet{nabizadeh2022covector}. No hyper-reduction is considered. \Cref{img:fluid:karman} shows that our latent space dynamics method faithfully represents the ground truth flow: from the initial steady state to the bifurcated vortex shedding regime. The ground truth solver utilizes a $100$ by $200$ grid, with a total $40,000$ degrees of freedom. By contrast, our method captures the same level of details while employing a latent space dimension of only $\nred=4$. Due to CROM's kinematics-approximation-only nature, our approach also generalize beyond the training temporal range.}
  
  Future work should test CROM on more challenging boundary conditions \citep{copeland2022reduced} \rev{and chaotic systems involving non-Markovian effects \citep{pan2018data}}.
\section{Solid Mechanics}
\label{sec:solid_details}
\subsection{Continuous PDE}
We solve the elastodynamics equation arisen from solid mechanics for deformable soft body modeling:
\begin{align}
    \rho_0 \ddot{\deformationMap} = \gradb \cdot \PKStress(\gradb \deformationMap)  + \rho_0\bodyForce,
\end{align}
where $\deformationMap(\Xb,t)$ maps the undeformed (reference) position ($\Xb$) of an arbitrary material point from the reference configuration $\spatialDomain$ to its deformed (current) position ($\xb$) at time $\continuousTime$. Proper Dirichlet and Neumann boundary conditions are applied.

We assume the first Piola–Kirchhoff stress $\PKStress$ is strictly a function of the deformation gradient $\deformationGrad = \gradb \deformationMap$. In particular, we adopt the corotated linear elasticity constitutive law,
\begin{align}
    \PKStress(\deformationGrad) = 2\secondlame(\deformationGrad-\Rb) + \firstlame\trace^2(\Rb^T\deformationGrad-\identity)\Rb,
\end{align}
where $\Rb$ is the rotation tensor from the polar decomposition of the deformation gradient $\deformationGrad=\Rb\Sb$, $\firstlame$ and $\secondlame$ are the first lame parameter (Pa) and the second lame parameter / shear modulus (Pa). The proposed framework also works with other elasticity constitutive laws and can be easily extended to capture hysteresis and inelasticity.

\subsection{Full-order Model}
We adopt the tetrahedral-mesh-based linear finite element method (FEM) for the full-order model. We closely follow the course by \citet{sifakis2012fem}. Here we review only the salient features and refer to their work for theoretical and practical details.

We spatially discretize the domain of interest using linear tetrahedra. For each tetrahedron, we have four undeformed vertex positions ($\Xb_1, \cdots, \Xb_4$) and four deformed vertex positions ($\xb_1, \cdots, \xb_4$). For each vertex, we have $\xb_i = \deformationGrad\Xb_i + \bb$, where $\bb$ is the translation.

Consequently, we can obtain the deformation gradient $\deformationGrad$ of each tetrahedron by solving $\Db_s = \deformationGrad\Db_m$, where $\Db_s=\begin{bmatrix} \xb_1-\xb_4 & \xb_2-\xb_4 & \xb_3-\xb_4 \end{bmatrix}$ and $\Db_m=\begin{bmatrix} \Xb_1-\Xb_4 & \Xb_2-\Xb_4 & \Xb_3-\Xb_4 \end{bmatrix}$.

Furthermore, the internal elastic forces on the vertices ($\fb_1, \cdots, \fb_4$) can be computed though $\begin{bmatrix} \fb_1 & \fb_2 & \fb_3 \end{bmatrix} = -\frac{1}{6}|\det\Db_m|\PKStress(\deformationGrad){\Db_m}^{-T}$ and $\fb_4 = -\fb_1-\fb_2-\fb_3$.

\subsection{Reduced-order Model}
Since the initial condition is known for a given problem, we opt to construct a manifold-parameterization function for the displacement field $\ub(\Xb,t) = \deformationMap - \Xb$. The deformation map can then be computed via $\deformationMap = \ub+\Xb$. The hyperparameters of the manifold-parameterization function are $\nred=2$ and $\beta=20$.

Unlike previously discussed PDEs, the elastodynamics equation is second-order in time. Consequently, in addition to the function value of the manifold, we also need to infer the temporal derivative during latent space dynamics. We do so by employing the tangent space of the manifold,
\begin{align}
    \dot{\deformationMap}(\pos,\discreteTime)=\lowDimensionalManifoldNNwrtLatent\discreteLatentSpaceVecDot.
\end{align},
\rev{where $\discreteLatentSpaceVecDot=(\discreteLatentSpaceVec- \discreteLatentSpaceVecMinus)/\dt$ and $\discreteLatentSpaceVecDotZero=\zerob$.}

To compute the acceleration (and increment the velocity) of a vertex belonging to the integration samples set $\sampleSet$, we need to evaluate the internal force at this particular vertex. We can calculate the vertex force by accumulating the internal forces of tetrahedra incident on this particular vertex. Tetrahedron forces are computed using the formula described in the previous section. To attain optimal gradient accuracy (\Cref{sec:nn-numerical-gradient}), we opt to use finite element's linear basis function to compute the deformation gradient $\deformationGrad$ of each tetrahedron.

To compute these internal forces and deformation gradients, we also need to obtain the position information (via the manifold-parameterization function) of all vertices in the one-ring neighborhoods of the integration samples. We refer to the set of these neighboring vertices as $\sampleSetNeighbor$. Therefore, the total number of spatial samples in the reduced-order model now becomes $\sampleSetCardinality + \sampleSetNeighborCardinality$. Since one-ring neighbors only entail a small subset of the original degrees of freedom, we still offer significant spatial sample reduction from the full order model ($\nSpatial$).

\subsection{Training and Testing Data} 
\subsubsection{Gravity-induced Impact}
\label{sec:gravity-impact}
An initially-static rectangular-shaped deformable object ($\nSpatial=\remeshHiVertexNumber$ vertices and $\remeshHiTetNumber$ tetrahedra) accelerates due to downward gravity. After impacting the collision objects underneath, the material undergoes intense deformation.

We generate training and testing data by sampling the shear modulus $\params=\mu \in \paramDomain=[60,000, 70,000] \subset \RR{}$. The first lame parameter $\firstlame$ is fixed to be zero, in which case the corotated linear elasticity constitutive law simplifies to the As-Rigid-As-Possible energy \citep{sorkine2007rigid}. Specifically, we uniformly generate $4$ samples of the shear modulus for training data. We simulate $200$ time steps for each shear modulus. Therefore, a total of $800$ time steps are used for training. We then randomly generate $3$ samples of the shear modulus from $\paramDomain$. We also simulate $200$ time steps for these shear moduli and use them for testing. We use the same initial condition for training and testing.

\subsubsection{Torsion and Tension}
\label{sec:roten}
We study another common solid mechanics application where the material undergoes torsion and tension. The left and the right boundaries kinematically move at constant translational and angular velocities of the same magnitude but in opposite directions.

We solve the parameterized family of problem where $\params=(\secondlame, \firstlame)\in \paramDomain=[6,200, 20,000]\times[10,000,65,000] \subset \RR{2}$. We obtain $9$ training samples via uniform full-factorial sampling of the parameter space. We simulate $100$ time steps for each training sample. We then use the Latin hypercube method to randomly generate another $4$ testing samples. We also simulate $100$ time steps for these testing samples. Training and testing data adopt the same initial condition.

\subsection{Results}
\Cref{img:falling_remeshing} demonstrates a testing case ($\mu=62500$, $2.68\%$ error from the full-order ground truth) for the experiment from \Cref{sec:gravity-impact}. Since our method has a discretization-agnostic architecture, we can adapt the simulation resolution to improve efficiency. The adaptive simulation adopts a contact-based oracle that switches from the low-resolution mesh to the high-resolution mesh once a contact event is detected. Before the start of the simulation, we generated both the low-resolution and the high-resolution meshes using TetWild \citep{hu2018tetrahedral} with small and large ideal edge lengths.

\begin{figure}
    \centering
    \includegraphics[width=\textwidth]{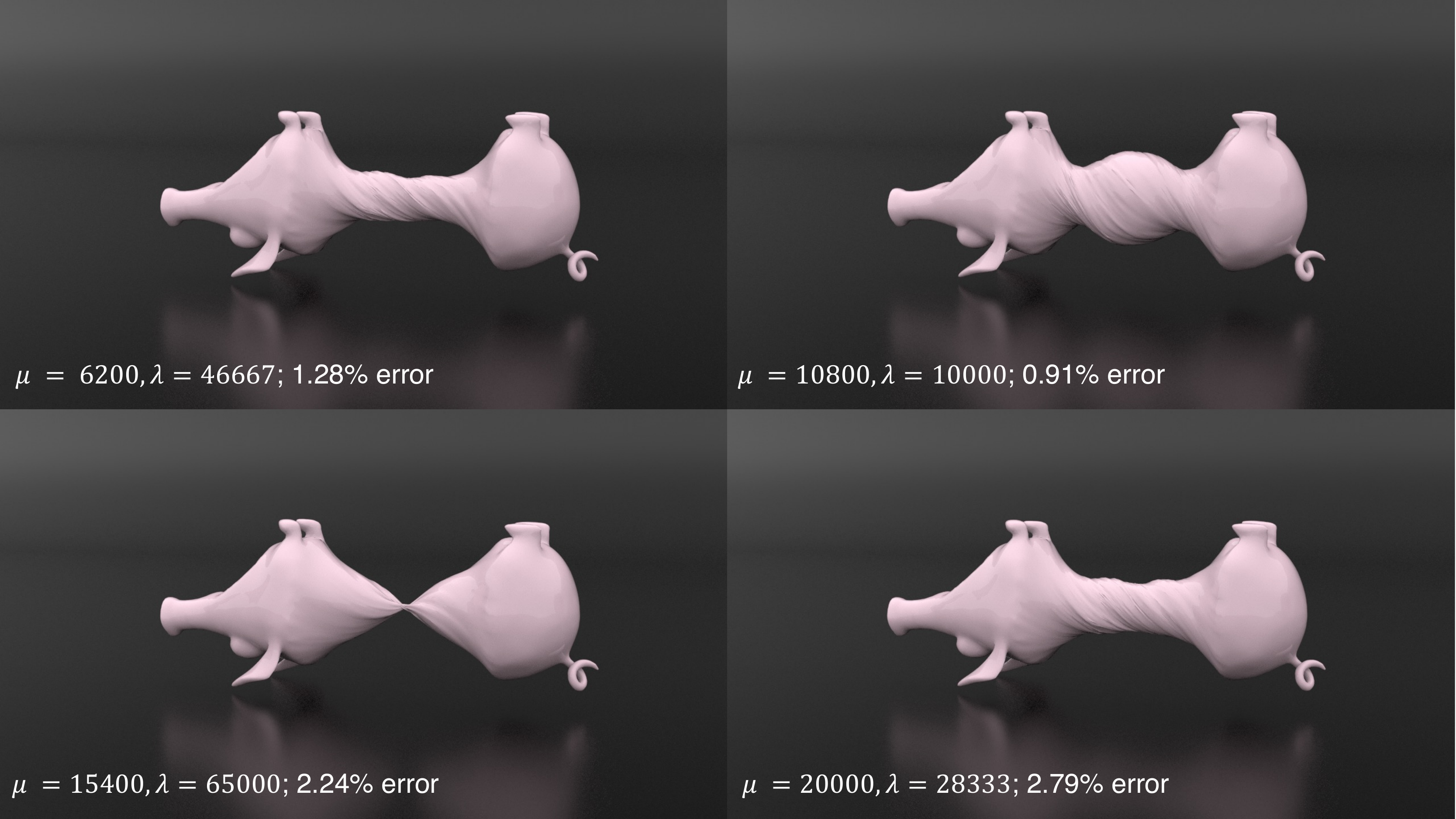}
    \caption{Our approach captures a wide range of elastic moduli with modest errors compared to the full-order ground truth.}
    \label{img:solid_mechanics:piggy:different_moduli}
\end{figure}
After training on the experiment from \Cref{sec:roten}, our method captures a wide range of shearing and volume-preserving behaviors (see \Cref{img:solid_mechanics:piggy:different_moduli}).

\begin{figure}
    \centering
    \includegraphics[width=\textwidth]{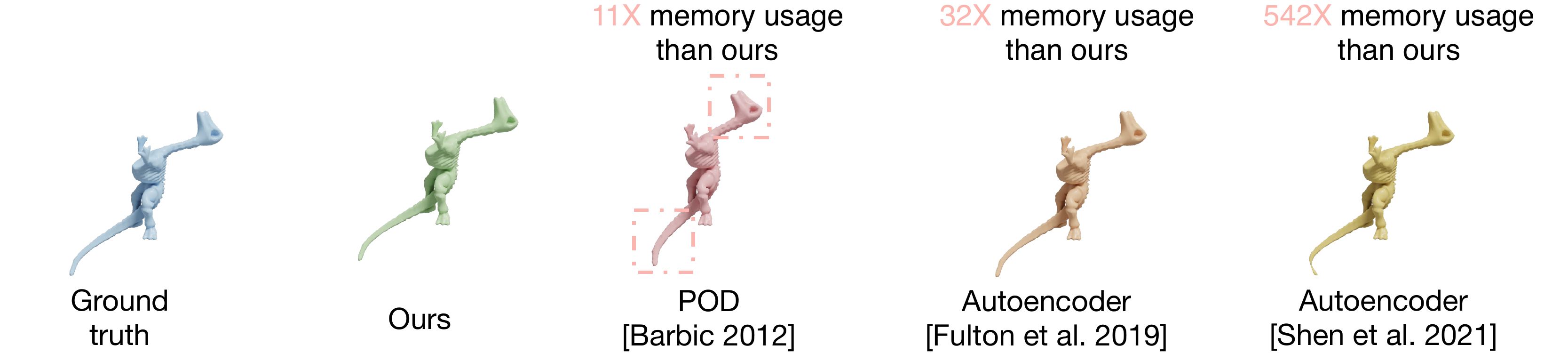}
    \vspace{-6mm} %
    \caption{CROM agrees well with the ground truth simulation while consuming far less memory than prior discretization-dependent ROM approaches, including POD \citep{barbivc2012fem} and neural-network-based autoencoder approaches \citep{fulton2019latent,shen2021high}. The experimental setup is taken from \citet{shen2021high}.}
    \label{img:dino_and_dragon}
  \end{figure}

Furthermore, to showcase CROM's ability to handle large deformations, we reproduce the dinosaur example by \citet{shen2021high}. \Cref{img:dino_and_dragon} shows that CROM accurately captures the large deformations of the dinosaur. By contrast, under the same latent space dimension, POD has a limited expressivity and suffers from visual artifacts: the head and the tail's motions are constrained. This observation is consistent with the results reported by \citet{shen2021high} where nonlinear approaches outperform linear approaches. In terms of memory consumptions, CROM is orders-of-magnitude more efficient than these prior discretization-dependent ROM approaches \citep{barbivc2012fem,fulton2019latent,shen2021high}.

\section{Comparison with Prior Approaches}
\label{sec:vs-prior}
We compare our approach with prior reduced-order approaches that generate fast PDE solutions via latent space dynamics. In total, we compare with four baselines: (1) the classic widely-adopted proper orthogonal decomposition (POD) method \citep{berkooz1993proper,holmes2012turbulence}. Our POD implementation follows the PCA approach by \citet{barbivc2005real,barbivc2012fem}; (2) the neural-network-based autoencoder approach by \citet{fulton2019latent}; (3) the neural-network-based autoencoder approach by \citet{shen2021high}; (4) the convolutional autoencoder (CAE) approach by \citet{lee2020model}. All these four baselines are \emph{discretization-dependent}.

SIREN, PINN, or GNN are related to our work but are very different in nature. While their goal is to directly solve PDEs with neural network architectures, our goal is to leverage neural networks for dimension reduction. Therefore, we do not compare with them and focus on comparison with other dimension reduction methods.

POD approximates the discretized vector field with a linear basis $\podbasis$,
\begin{align}
    \encInputArg{t}=\encInputDetailArg{t} = \podbasis\latentSpaceVecArg{t},
\end{align}
where $\podbasis$ is a matrix of size $\nSpatial\dimensionOut$ by $\nred$. $\podbasis$ is most easily constructed via a singular value decomposition of the discretized training data \citep{sifakis2012fem}.

Autoencoders have the same input and output as POD. However, the linear basis is replaced with nonlinear neural networks,
\begin{align}
    \encInputArg{t}=\encInputDetailArg{t} = \decoderArg{\latentSpaceVecArg{t}},
\end{align}
where $\decoder$ is the decoder portion of the autoencoder structure. \citet{fulton2019latent,shen2021high,lee2020model} all follow this formulation but differ in their particular architecture choices.

We implement both POD and the autoencoder approaches in the same PyTorch framework as CROM. These approaches share exactly the same PDE time-stepping algorithm (step 2) during latent space dynamics. For POD and the autoencoder approach, the network inference step (step 1) and the network inversion step (step 3) employ the linear basis $\podbasis$ and the nonlinear decoder function $\decoder$, respectively.

\subsection{Speed}
We measure the wall-clock computation time during the falling stage (before contact) of the gravity-induced impact experiment (see \Cref{sec:gravity-impact} and \Cref{img:falling_remeshing}). In particular, since all model reduction methods (including our approach, POD, and the autoencoder approaches) share the same PDE time-stepping algorithm (step 2), we measure its wall-clock computation cost to ensure the generalizability of the comparison.

\begin{figure}
    \centering
    \includegraphics[width=\textwidth]{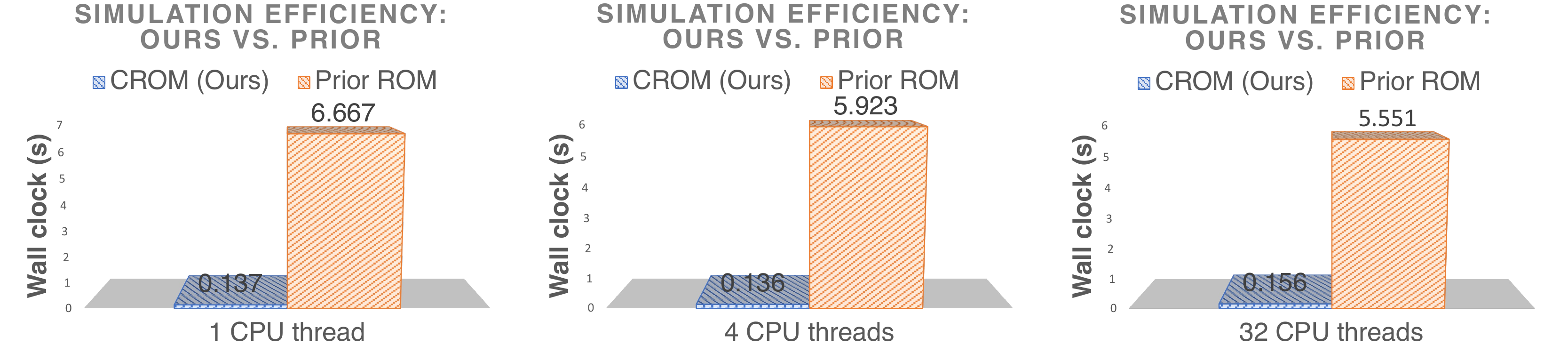}
    \caption{Wall clock time comparison with prior ROM approach. Our approach obtains considerable speedups across different computational resources.}
    \label{img:vs_prior:speed}
\end{figure}

Since POD and the autoencoder approach are baked into a particular discretization, it does not allow for adaptive discretization. Therefore, the high-resolution mesh ($\remeshHiVertexNumber$ vertices, $\remeshHiTetNumber$ tetrahedra) has to be used throughout the falling stage even though there is minimal deformation. Consequently, its computation cost is significantly higher than our approach, which employs a low-resolution mesh ($\remeshLoVertexNumber$ vertices, $\remeshLoTetNumber$ tetrahedra) during the falling stage. \Cref{img:vs_prior:speed} shows that our approach is faster than the prior ROM approach with different CPU thread counts. Our method's advantage is most obvious when there is extremely limited computational resource (1 CPU thread). Since this speedup comes from the discretization independent nature of CROM, we expect it to generalize to other prior discretization-dependent ROM approaches as well. However, for application where remeshing is unnecessary, we do not expect this kind of speedup.

Hyper-reduction techniques \citep{an2008optimizing}, such as the integration samples introduced in our approach, can also be employed to speed up the high-resolution mesh solution by reducing spatial samples. However, this is orthogonal to our adaptive discretization contribution, and the hyper-reduction techniques can be employed to further accelerate the low-resolution mesh solution as well.

\subsection{Memory}
We compare the memory consumptions of our method, POD, and the autoencoder approaches with the same latent space dimension ($\nred$), both trained on data from the torsion and tension experiment (see \Cref{sec:roten}). For all implicit neural representation networks, we set the MLP width to be $\scaleMlp=20$. Independent of latent space dimension ($\nred$), our network has a near-constant memory consumption (see \Cref{img:mem_and_accuracy}). By contrast, the memory consumptions of POD and the autoencoder approaches scale linearly with $\nred$. More importantly, since the output dimension of POD and the autoencoder is $\nSpatial\dimensionOut$, their memory consumptions scale linearly with the number of discretized positions $\nSpatial$. By contrast, our approach's memory consumption is independent of the number of spatial samples. Consequently, in this large-scale example that features $\nSpatial=\piggyVertexNumber$ vertices, we observe more than ten-fold advantages with our method across all latent space dimensions (see \Cref{img:mem_and_accuracy}).

\begin{figure}
    \centering
    \includegraphics[width=\textwidth]{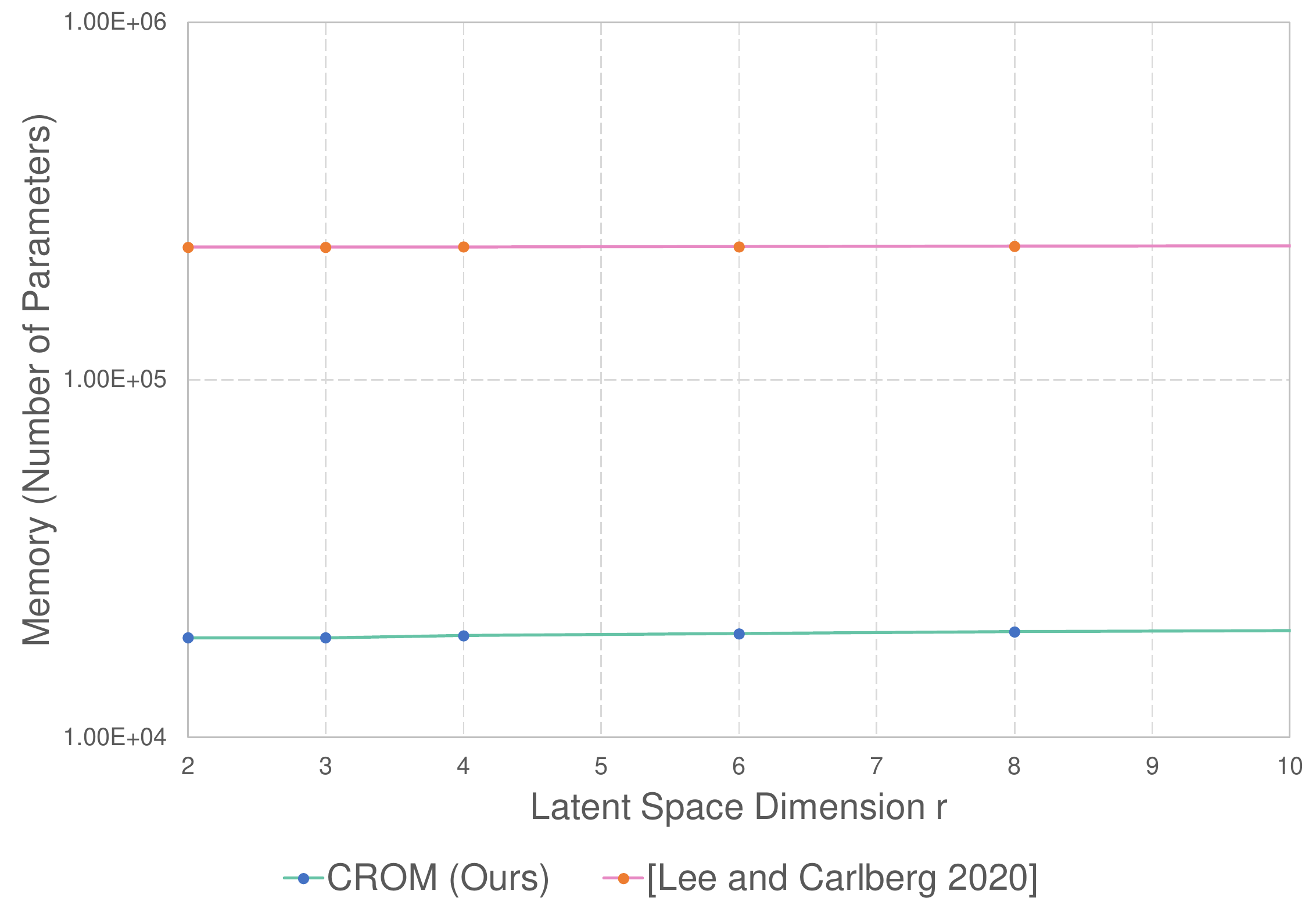}
    \caption{Memory comparison between our CROM and \citet{lee2020model}}
    \label{img:vs_prior_cnn:memory}
\end{figure}

In addition, we also compare our method with POD on the image processing experiment (see \Cref{sec:img-processing}). CROM uses an order-of-magnitude less memory than POD (see \Cref{img:vs_prior:imgprocess}c). In \Cref{img:vs_prior_cnn:memory}, we also demonstrate CROM's memory advantage over the convolutional autoencoder approach by \citet{lee2020model}.

Even though the overall memory consumption of our approach is significantly lower, our approach's per-sample evaluation cost is higher. For any given discretized spatial sample, POD only requires a vector-matrix multiplication of a $\dimensionOut$ by $\nred$ matrix. By contrast, our approach needs to go through the entire MLP that employs $\dimensionOut\scaleMlp$ by $\dimensionOut\scaleMlp$ matrices. The autoencoder approaches share similar computation overheads.

\subsection{Accuracy}
\Cref{img:mem_and_accuracy} demonstrates the accuracy of the networks studied in the previous memory section. Our method consistently offers orders-of-magnitude smaller training accuracies. This accuracy advantage is especially noticeable with small latent space dimensions (e.g., $\nred=2$). Further increasing the latent space dimension leads to higher accuracies with POD and the autoencoder approaches but not our approach. As discussed in \Cref{sec:hyper-study}, our approach's accuracy depends more saliently on the MLP width as opposed to the latent space dimension. Furthermore, unlike these discretization-dependent approaches, our continuous manifold-parameterization function also facilitates accurate handling of adaptive spatiotemporal data \citep{pan2022neural}.

Furthermore, \Cref{img:vs_prior:imgprocess} shows that our method is more accurate than POD on the image processing experiment, both visually and quantitatively. \Cref{img:relative_error} displays our accuracy advantage over CAE on Burgers' equations.

\subsection{\rev{Training time}}
\begin{figure}
    \centering
    \includegraphics[width=\textwidth]{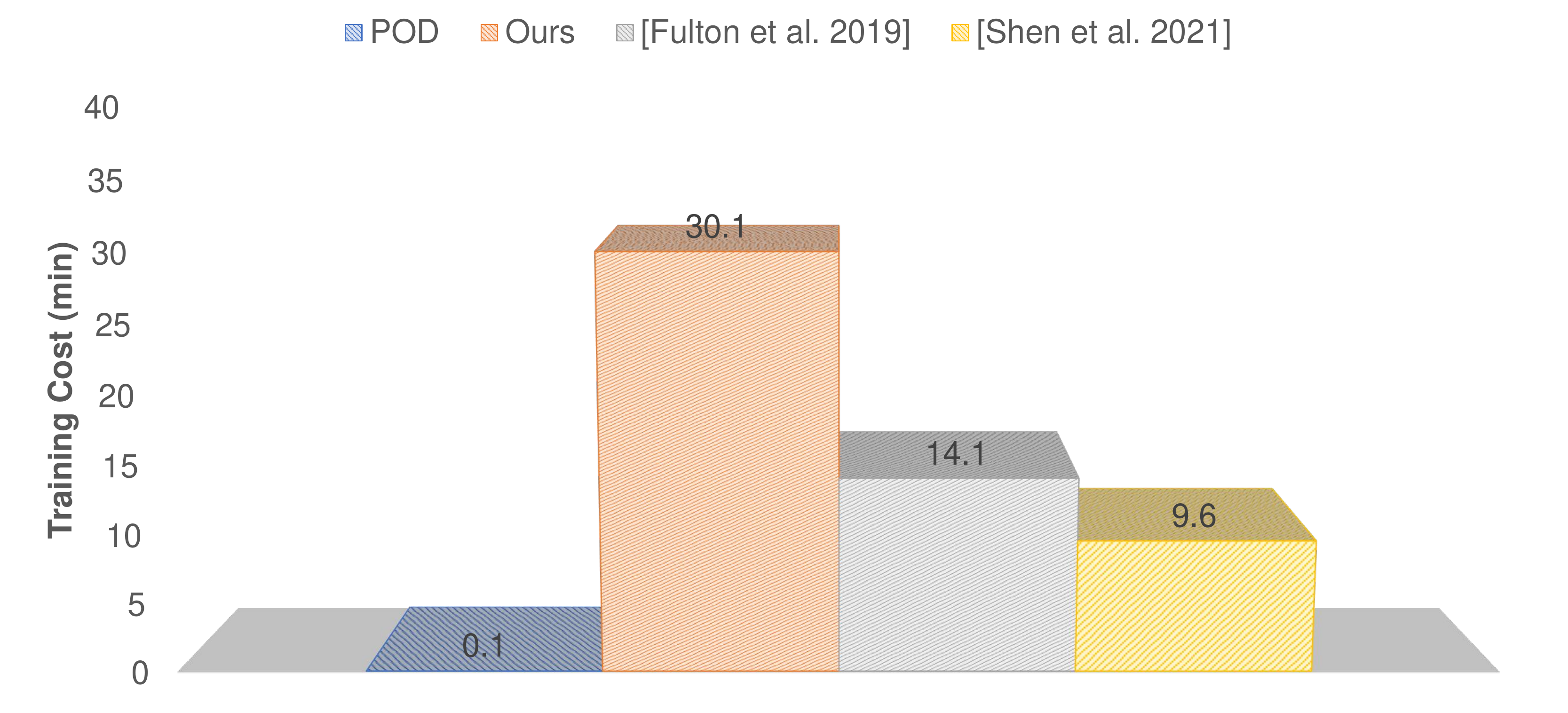}
    \caption{\rev{Training time for the elasticity example presented in \Cref{img:dino_and_dragon}. POD \citep{barbivc2012fem} uses significantly less time than neural-network-based methods, including \citep{fulton2019latent,shen2021high} and ours.}}
    \label{img:vs_prior:inference_speed}
\end{figure}
\rev{\Cref{img:vs_prior:inference_speed} reports the offline training time for different ROM approaches. All neural-network-based methods, including ours, take longer time to train than the linear POD approach. We emphasize that the timing data is purely informative and calculated post-hod. The main target of our work is not to optimize training time but to achieve efficient latent space dynamics. Nevertheless, specifically for our implicit neural representation approach, future work may consider expediting training time via more advanced data structures and optimization \citep{muller2022instant,liu2020neural,martel2021acorn,takikawa2021neural}.}

\section{Comparison with the Full-order Model}
\label{sec:performance_eval}
To gauge the practical performance of our approach over the (unreduced) full-order model, we measure the sheer wall-clock performances of our reduced-order model and the full-order model.

To facilitate fair comparison, we implement, optimize, and (fully) parallelize the full-order PDE solution and the reduced-order approach within the PyTorch framework \citep{paszke2019pytorch} without any other dependency. The full-order model and the reduced model share the same (parallelized) PDE solver code (step 2), and the only difference between them is the neural network evaluation in the reduced model. To encourage fair comparison, for operations involving neural networks (step 1 and step 3), we do not use customized SIMD vectorizations, CUDA kernels, or highly optimized inference libraries such as TensorRT \citep{vanholder2016efficient}, though these optimizations should lead to a further performance gain of our algorithm. In addition, having both models implemented in the same framework allows us to compare them under the same computing environments, avoiding biased comparison where the full-order model runs on a significantly limited computing resource (e.g., 8-core CPU) while the reduced-order model runs on a high-performance computing device (e.g., 5120-core GPU).

\begin{figure}
    \centering
    \includegraphics[width=\textwidth]{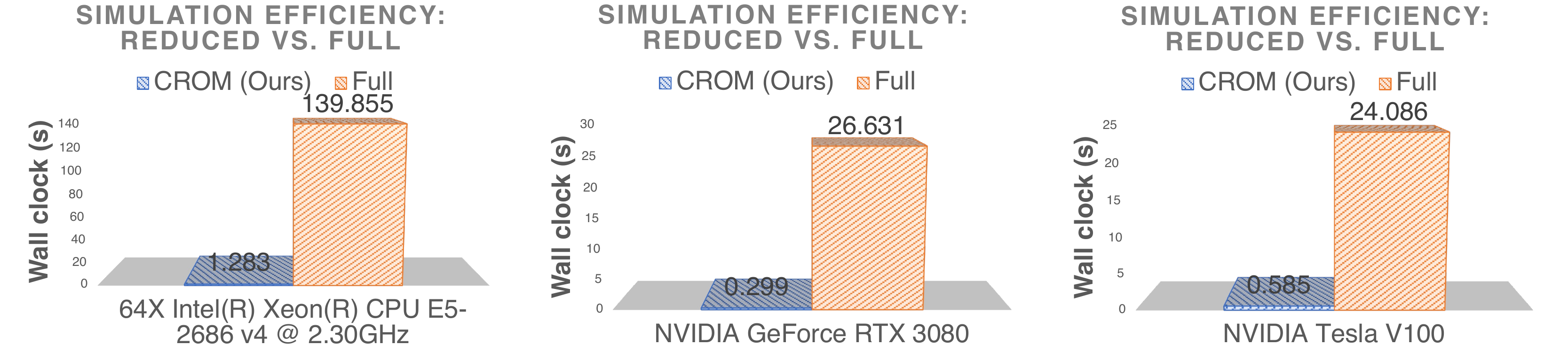}
    \caption{Wall clock time comparison with the full-order method. Our approach obtains significant speedups across different computing platforms.}
    \label{img:vs_full:speed}
\end{figure}

\Cref{img:vs_full:speed} reports the wall clock time of our reduced-order method and the full-order method across different computing platforms. While our approach is significantly faster on all computing platforms, its strongest speedup is obtained on the CPUs. This is unsurprising since the computationally expensive full-order model benefits more from high-end data-center GPUs (e.g., V100) than the reduced-order model. In addition, because the neural network in the reduced-order model only employs a few $60$ by $60$ matrices ($\dimensionOut=3,\scaleMlp=20$), inference on CPU is also efficient. This opens doors for employing our models on limited computing platforms such as mobile and VR devices.

\begin{figure}
        \centering
        \includegraphics[width=\textwidth]{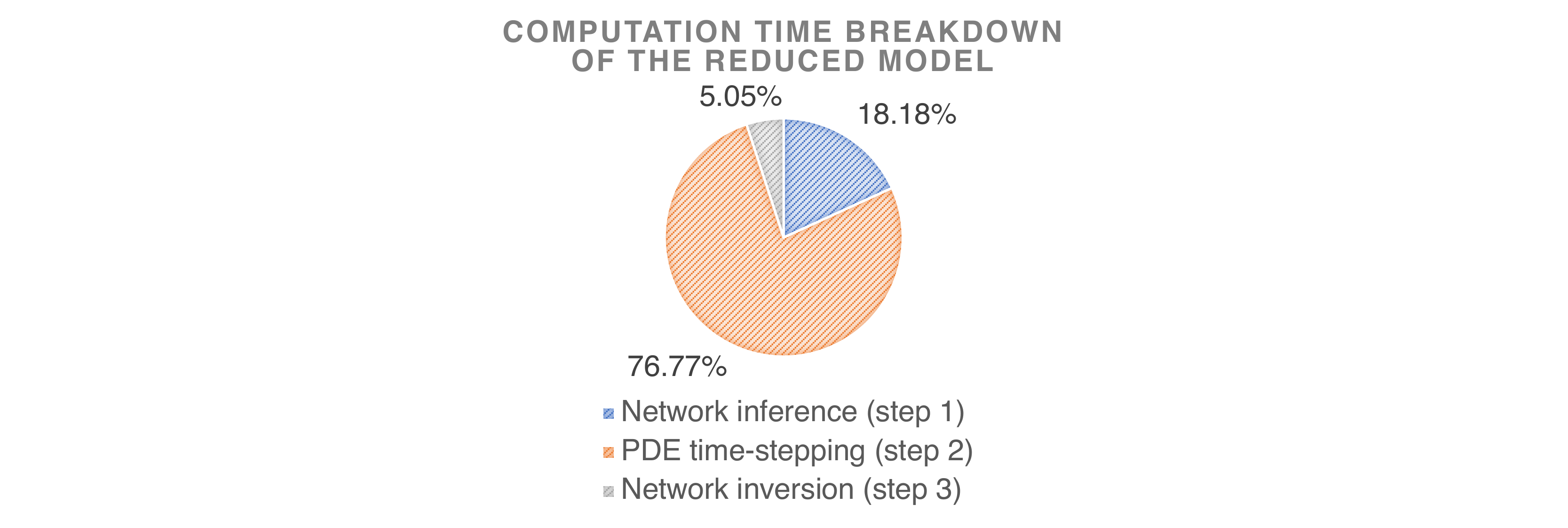}
        \caption{CROM timing breakdown. Neural network operations introduce minimal overhead. The computation bottleneck remains to be PDE time-stepping.}
        \label{img:vs_full:breakdown}
\end{figure}
\Cref{img:vs_full:breakdown} further breaks down the computation time spent on each component of the reduced-order model. While the overhead of the neural network operations (step 1 and step 3) is non-negligible, the bottleneck of the algorithm remains to be PDE time-stepping (step 2). In particular, the network inversion cost is very low, thanks to the linearized inversion solver discussed in \Cref{sec:net-inversion-details}.

\begin{figure}
    \centering
    \includegraphics[width=\textwidth]{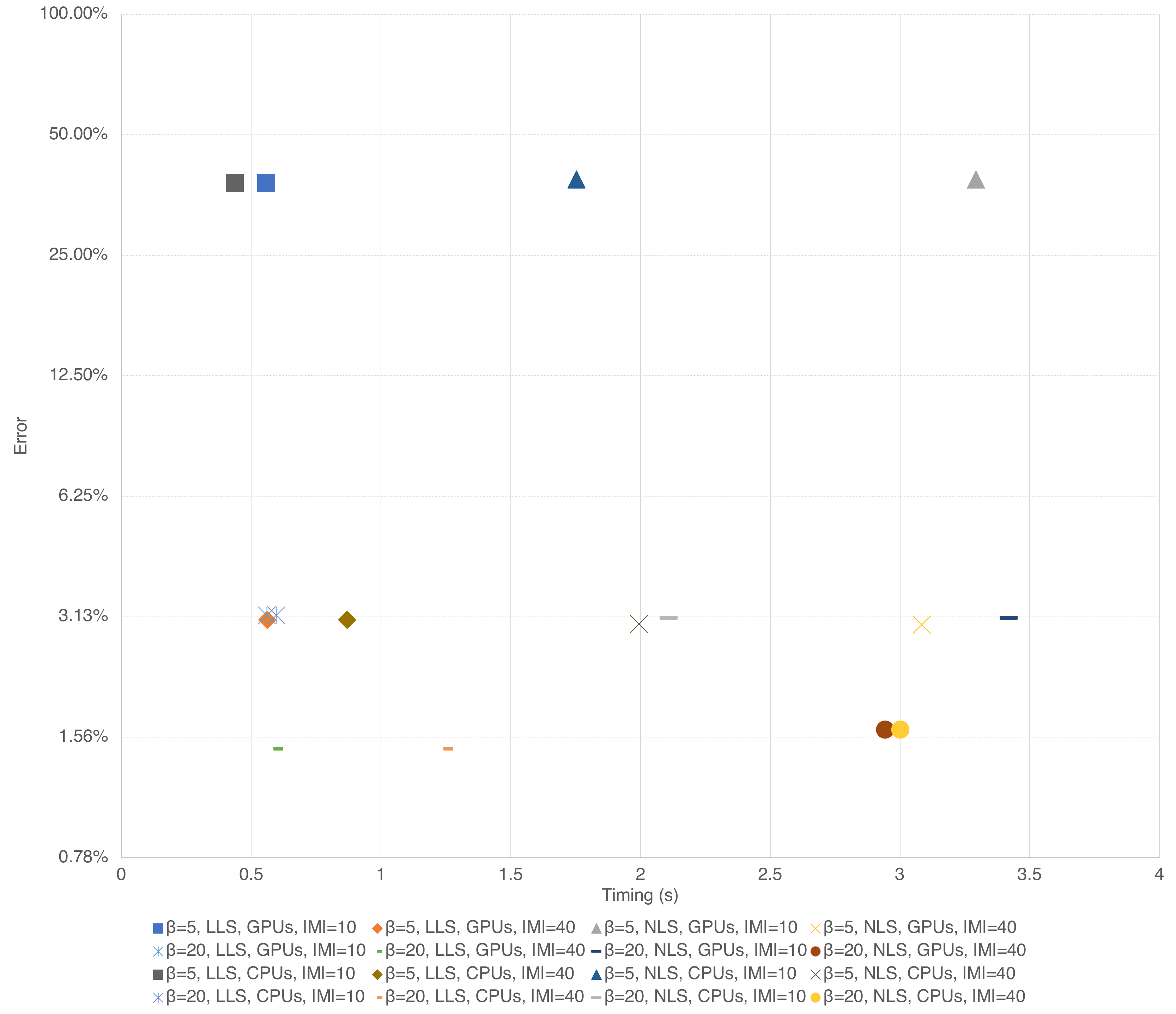}
    \caption{Accuracy vs. Speed: our approach under different setups. In comparison, the ground truth full-order takes $139.855$s on CPUs and $24.086$s on GPUs (see \Cref{img:vs_full:speed}). Our approach obtains significant speedups with less than $5\%$ error under various settings.}
    \label{img:vs_full:pareto}
\end{figure}

\Cref{img:vs_full:pareto} reports the accuracy and the speed of our approach using different MLP layer widths ($\scaleMlp$), neural network inversion methods (nonlinear least squares (NLS) vs. linear least squares (LLS)), computing platforms (CPU vs. GPU), and integration sample sizes ($\sampleSetCardinality$). In general, we observe similar accuracies between NLS and LLS (e.g., [$\scaleMlp=20$, LLS, GPUs, $\sampleSetCardinality=40$] vs. [$\scaleMlp=20$, NLS, GPUs, $\sampleSetCardinality=40$]), while LLS is significantly faster. Increasing sample size lead to higher accuracies but also require longer computation time (e.g., [$\scaleMlp=5$, NLS, CPUs, $\sampleSetCardinality=10$] vs. [$\scaleMlp=5$, NLS, CPUs, $\sampleSetCardinality=40$]). Increasing the MLP layer width yields higher accuracies but also increases the computation time. In particular, the computation time is increased more on CPUs (e.g., [$\scaleMlp=5$, LLS, CPUs, $\sampleSetCardinality=40$] vs. [$\scaleMlp=20$, LLS, CPUs, $\sampleSetCardinality=40$]) than on GPUs.

\section{Latent-space trajectory}
\label{sec:latent_space_traj}
A key element of CROM is latent space dynamics, in which the latent space vector evolves nonlinearly over time under explicit PDE constraints.

\begin{figure}
  \centering
  \includegraphics[width=\textwidth]{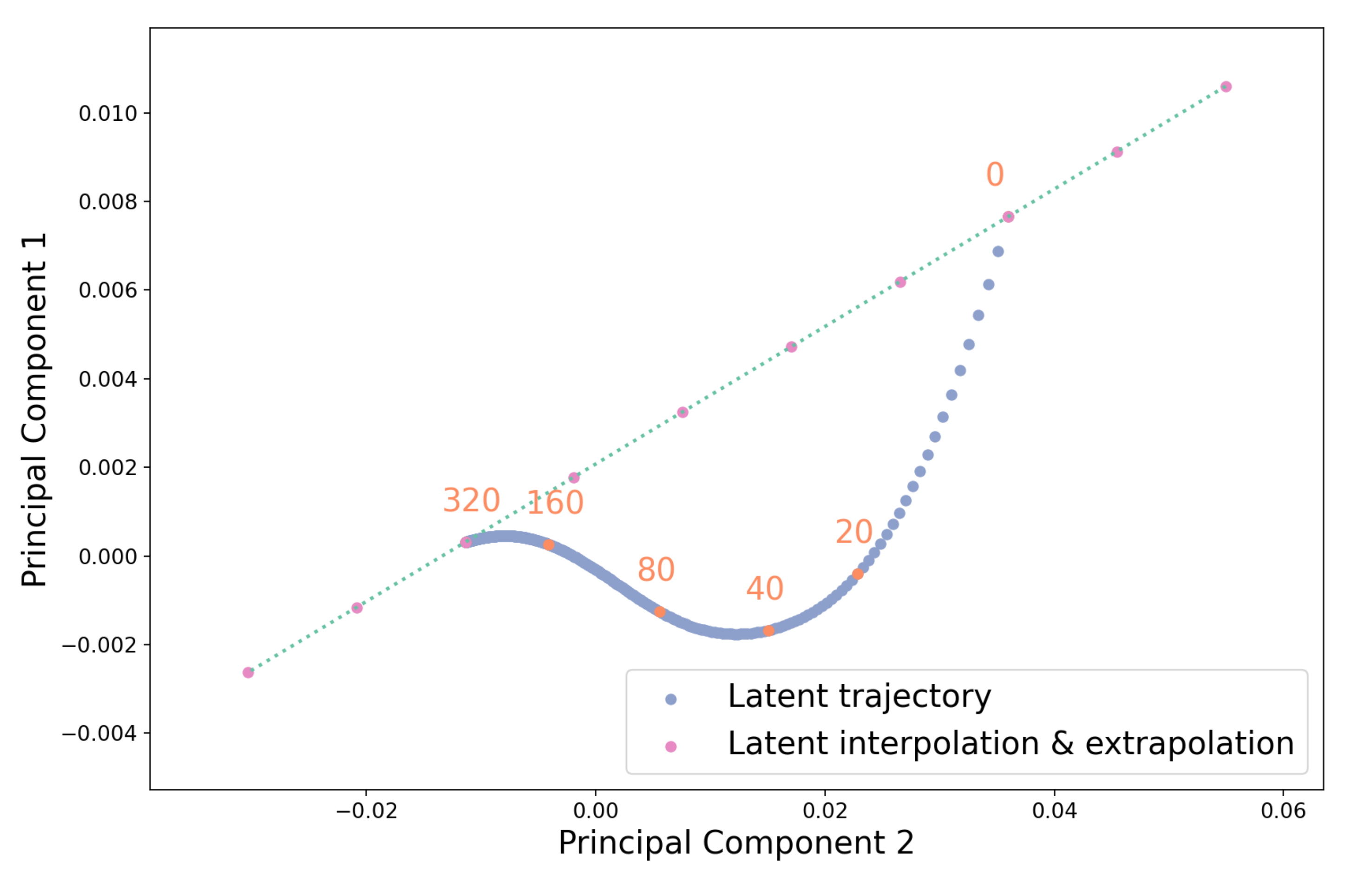}
  \caption{Latent-space trajectory. We visualize the first two principal components of the latent space vector. The red numbers displayed on top of the trajectory are time step numbers. The simulation starts from $0$ and ends at $320$. The corresponding images of these time steps are shown in \Cref{img:trajectory:simulation}. Overall, we observe a smooth nonlinear trajectory over time. In addition, we also linearly interpolate and extrapolate inside the latent space. The corresponding images are displayed in \Cref{img:trajectory:interpolatio}.}
  \label{img:trajectory:latent}
\end{figure}

\begin{figure}
  \centering
  \includegraphics[width=\textwidth]{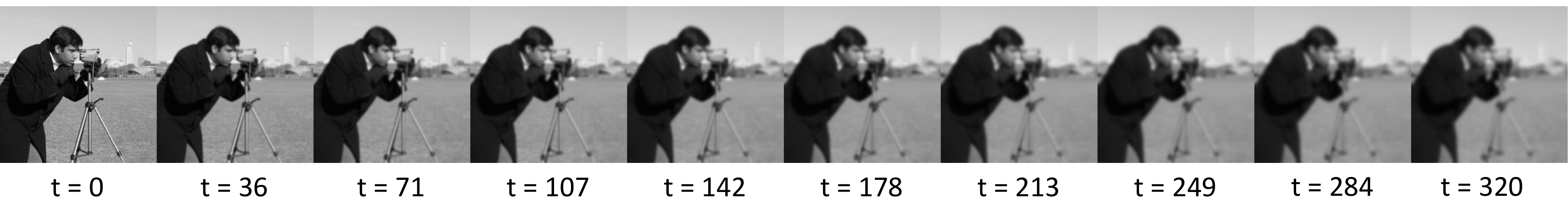}
  \caption{Reduce-order simulation of image smoothing at at different time steps $t$. The corresponding latent space trajectory is visualized in \Cref{img:trajectory:latent}.}
  \label{img:trajectory:simulation}
\end{figure}

\Cref{img:trajectory:latent} demonstrates a \emph{nonlinear} latent space trajectory from the image smoothing experiment. Guided by the PDE, the latent space vector traverses through the latent space and obtains the desirable smoothing effects in the image (see \Cref{img:trajectory:simulation}).

\begin{figure}
  \centering
  \includegraphics[width=\textwidth]{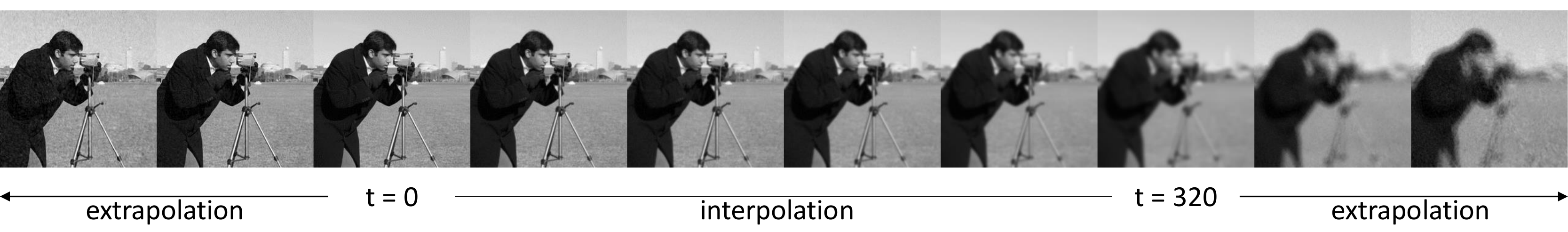}
  \caption{Linear interpolation and extrapolation of the latent space vector. We observe smooth transitions. The corresponding latent space vector values are shown in \Cref{img:trajectory:latent}.}
  \label{img:trajectory:interpolatio}
\end{figure}

After the simulation is finished, as a postprocessing experiment, we \emph{linearly} interpolate and extrapolate inside the latent space (see \Cref{img:trajectory:latent}). The interpolation end points are the latent vector for the unmodified image and the latent vector for the final image determined by the latent space dynamics. \Cref{img:trajectory:interpolatio} demonstrates the corresponding smooth transitions in the image space.

\section{Stability analysis}
\label{sec:stability}
\begin{figure}
  \centering
  \includegraphics[width=\textwidth]{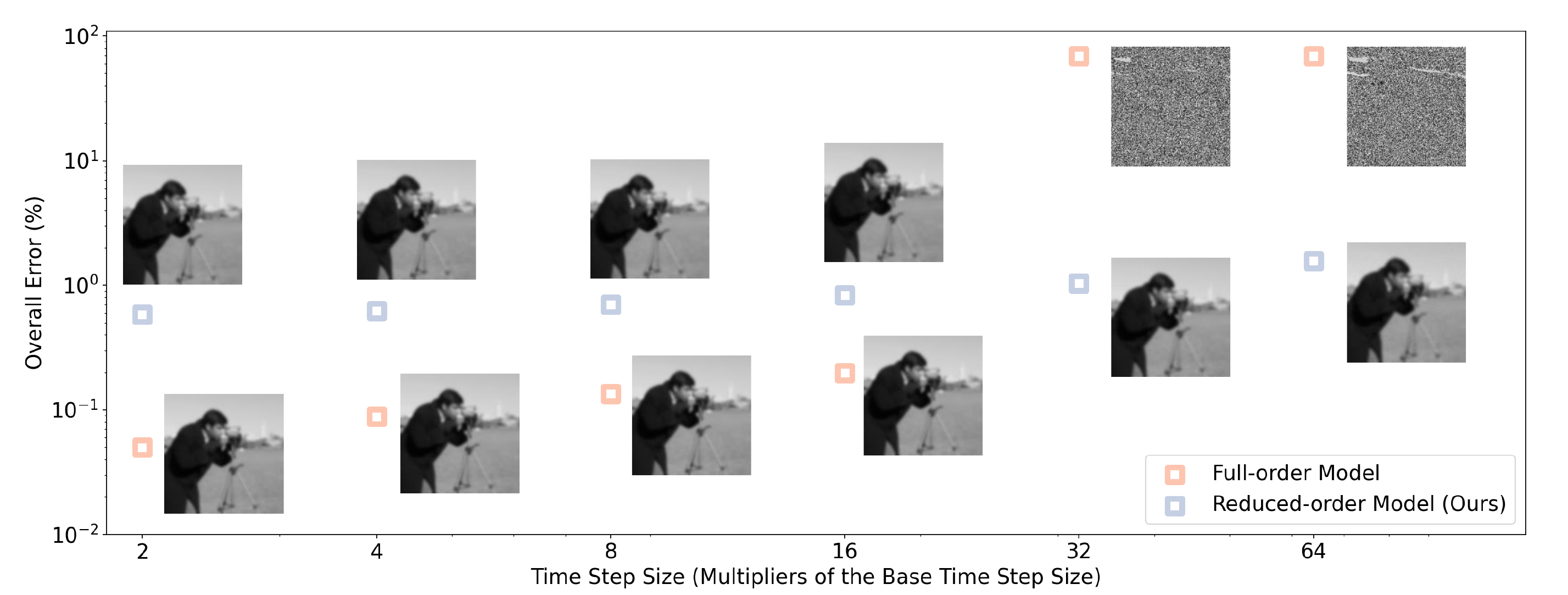}
  \caption{Stability analysis. We systematically increase the time step size ($\dt$). Our method remains stable in a wide range of time step size and maintains an error close to $1\%$. By contrast, the full-order model (that was used for training data generation) goes unstable as the time step increases and suffers from major visual artifacts.}
  \label{img:stability:stable}
\end{figure}
\Cref{img:stability:stable} systematically studies the stability of our method. CROM exhibits much larger stable time step sizes than its unreduced counterpart. This is consistent with the theoretical analysis reported in the literature \citep{bach2018stability}.
\section{Latent-space Dynamics Pseudocode}
\label{sec:pseudocode}

\begin{algorithm}[H]
    \SetAlgoLined
    \KwData{Latent space vector $\discreteLatentSpaceVec$}
    \KwResult{Next-time-step latent space vector $\discreteLatentSpaceVecPlus$}
    Network inference: gather full-space information \rev{(\Cref{alg:inference})}.

    PDE time-stepping with boundary conditions: update full-space information via differential equations, e.g., the elastodynamics equation \rev{(more details available from \Cref{sec:thermo_details} to \Cref{sec:solid_details})}.

    Network inversion: optimally project the updated full-space information onto the low-dimensional embedding \rev{(\Cref{alg:inversion})}.
    \caption{Latent-space Dynamics}
    \label{alg:latent-space-dynamics}
\end{algorithm}

\begin{algorithm}[H]
    \SetAlgoLined
    \KwData{Latent space vector $\discreteLatentSpaceVec$}
    \KwResult{Full space information $\fullOrderModel_{n}, \gradb\fullOrderModel_{n}, \dot{\fullOrderModel}_{n}$}
    \For{$\posSample\in\sampleSet$}{
        Compute the vector field value: $\fullOrderModel(\posSample,\discreteTime)=\lowDimensionalManifoldNNArgs{\posSample}{\discreteLatentSpaceVec}$.\\
        Compute the temporal gradient either by differentiating the network $\dot{\fullOrderModel}(\posSample,\discreteTime)=\lowDimensionalManifoldNNwrtLatentFlat\discreteLatentSpaceVecDot$, where $\discreteLatentSpaceVecDot=(\discreteLatentSpaceVec- \discreteLatentSpaceVecMinus)/\dt$, or by numerical approximation (\Cref{sec:net_grad}).\\
        Compute the spatial gradient either by differentiating the network,  $\gradb\fullOrderModel(\posSample,\discreteTime)=\gradbspatialSample\lowDimensionalManifoldNN$, or by numerical approximation (\Cref{sec:net_grad}).
    }
    \caption{\rev{Network inference: gather full-space information}}
    \label{alg:inference}
\end{algorithm}

\begin{algorithm}[H]
    \SetAlgoLined
    \KwData{Updated full space information $\fullOrderModel_{n+1}$}
    \KwResult{Updated latent space vector $\discreteLatentSpaceVecPlus$}
    Find the latent space vector $\discreteLatentSpaceVecPlus$ that best matches the evolved configuration ${\fullOrderModel}_{n+1}$ by solving the least-squares:
    \begin{align*}
        \min_{\discreteLatentSpaceVecPlus\in\RR{\nred}}\sum_{\posSample\in\sampleSet} \|\lowDimensionalManifoldNNArgs{\posSample}{\discreteLatentSpaceVecPlus} - \fullOrderModelArgs{\posSample}{\discreteTimePlus}\|_2^2 \ .
    \end{align*}
    \caption{\rev{Network inversion: optimally project the updated full-space information onto the low-dimensional embedding}}
    \label{alg:inversion}
\end{algorithm}
\section{Reduced-order Model's statistics}
\label{sec:model_statistics}
\begin{table}[t]
    \caption{Dimension reduction statistics of our reduced-order model}
    \centering
    \begin{tabular}{cccccc} \toprule
    PDE & Full-order degrees  & Latent space & Dimension & Spatial & Spatial sample  \\ 
    & of freedom & dimension & reduction & sample count &  reduction\\ \midrule
    1D Thermo & 501 & 16 & 31$\times$ & 22 & 23$\times$    \\ 
    \midrule
    2D Image & 65,536 & 16 & 4,096$\times$ & 63 & 1,040$\times$    \\ \midrule
    1D Advection & 100 & 1 & 100$\times$ & n/a & n/a    \\ \midrule
    1D Burger & 256 & 2 & 128$\times$ & n/a & n/a    \\ \midrule
    2D fluid & 512 & 6 & 86$\times$ & n/a & n/a    \\ \midrule
    3D solid impact & 6,195  & 2 & 3,098$\times$ & n/a & n/a    \\ \midrule
    3D solid piggy & 199,824  & 2 & 99,912$\times$ & 40 & 1,665$\times$     \\ \midrule
    3D solid dinosaur & 11,565  & 1 & 11,565$\times$ & n/a & n/a     \\ 
    \bottomrule
    \end{tabular}
    \label{tab:dimension-reduction}
    \end{table}
CROM obtains considerable dimension reductions across different PDEs (\Cref{tab:dimension-reduction}).

\end{document}